\documentclass[review]{elsarticle}

\usepackage{lineno,hyperref}
\usepackage{graphicx}

\usepackage{color}
\usepackage{comment}
\usepackage{amsmath,amssymb,amsthm}
\usepackage{lineno,hyperref}
\usepackage[caption=false]{subfig}
\usepackage{epstopdf}
\usepackage{layouts}
\usepackage{hyperref}
\usepackage{tabularx}
\modulolinenumbers[5]

\makeatletter
\def\input@path{{./}}
\makeatother

\journal{ISPRS Journal of Photogrammetry and Remote Sensing}

\begin{document}
	
%Mathematik Kuerzel

% Pfeile
\newcommand{\ra}{\rightarrow}
\newcommand{\Ra}{\Rightarrow}

%atan2 (mit 1 Argument; sin- und cos-Anteil mit , oder ; getrennt)
\newcommand{\atan}[1]{\mbox{atan2} \, ({#1})}

% sans serif N, Norm
\newcommand  {\N}   {{\sf N}}

% sans serif P, Projektivit"t
\renewcommand{\P}   {{\sf P}}

%#1 = sans serif, Menge
\newcommand  {\M}[1]{{\cal {#1}}}

% Abstand der Komponenten in der Vektorschreibweise (vector-distance)
\newcommand{\vd}{\:}

%Vektor-dot
\newcommand{\vdot}{{\bf\cdot}}

%#1 = dick, Vektor, Matrix
\renewcommand{\d}[1]{\mbox{\boldmath$#1$}}
\newcommand{\di}[1]{\mbox{{\begin{footnotesize}\boldmath$#1$\end{footnotesize}}}}
\newcommand{\dii}[1]{\mbox{{\begin{scriptsize}\boldmath$#1$\end{scriptsize}}}}
% underline
\newcommand{\un}[1]{\underline{#1}}

%#1 = stochastisch, Skalar
\newcommand{\s}[1]{{\un{#1}}}

%#1 = dick, stochastisch Vektor, Matrix
\newcommand{\sd}[1]{{\un{\mbox{\boldmath{$#1$}}}}}

% dick, Vektor, Matrix%
\newcommand{\dd}[1]{{\mbox{\boldmath$#1$\unboldmath}}}

% Ausgabe von Schaetzgroessen%
\newcommand{\est}[1]{\widehat{#1}}

% geschaetzte Groessen %
\newcommand{\estd}[1]{\hat{\mbox{\boldmath$#1$}}}

%#1 = System, #2 = Skalar
%\newcommand{\cs}[2]{\overset {#1}{#2}}
\newcommand{\cs}[2]{\; ^{#1}\!\,{#2}}

%#1 = System, #2 = Vektor/Matr.
%\newcommand{\cv}[2]{\overset {#1}{\mbox{\boldmath$#2$}}}
\newcommand{\cv}[2]{\;^{#1}\!\,{\mbox{\boldmath$#2$}}}

%#1 = System, #2 = Quaternion/4 x 4 Matrix
%\newcommand{\cq}[2]{\overset {#1}{\bf #2}}
\newcommand{\cq}[2]{\;^{#1}\!\,{\bf #2}}

% transpose
\newcommand{\tra}[0]{{\sf T}}
\newcommand{\trans}[0]{^{\sf T}}
\newcommand{\transs}[0]{{'}^{\sf T}}
\newcommand{\transss}[0]{{''}^{\sf T}}
\newcommand{\transsss}[0]{{'''}^{\sf T}}
\newcommand{\transi}[0]{^{-\sf T}}

\newcommand{\transn}[0]{^{(0)\sf T}}
\newcommand{\transsn}[0]{{'}^{(0)\sf T}}
\newcommand{\transssn}[0]{{''}^{(0)\sf T}}

\newcommand{\sn}[0]{{'}^{(0)}}
\newcommand{\ssn}[0]{{''}^{(0)}}

\newcommand{\Amatrix}[0]{{\,\sf
  I\!\!\protect\rule[1.45ex]{0.8em}{0.8pt}\!\!\sf I \,}}

\newcommand{\Aqmatrix}[0]{\overline{\Amatrix}}
\newcommand{\Bqmatrix}[0]{\overline{\Bmatrix}}

%dualizing operator
\newcommand{\dual}[1]{{\overline{#1}}}

% transpose ohne ^ 
\newcommand{\tr}[0]{\sf T}

%
%homogener 4-Vektor, Quaternion, zugeh"riges(!)%
%4x4 Matrix-Symbol,fett-gerade%
\newcommand{\q}[1]{{{\bf #1}}}

\newcommand{\supremum}[2]{{{\underset{#1}{\mathtt{sup}}\left[ #2 \right]}}}

% function ..
\newcommand{\func}[2]{{{\mathtt {#1}\left[ #2 \right]}}}

% function, no argument
\newcommand{\funca}[1]{{{\mathtt {#1}}}}

% image ..
\newcommand{\img}[2]{{{\mathcal {#1}\left[ #2 \right]}}}

% image, no argument
\newcommand{\imga}[1]{{{\mathcal {#1}}}}

\newcommand{\sq}[1]{{{\s{\bf #1}}}}

% Index f"ur homogenes Element: Aufrecht
\newcommand{\qi}[1]{{\mbox{\tiny #1}}}

% Matrix
%%% \newcommand{\m}[1]{{\mbox{{\fontencoding{T1}\sffamily\slshape{#1\/}}}}}
% \newcommand{\m}[1]{\d #1}
 \newcommand{\m}[1]{{\mbox{{\fontencoding{T1}\sffamily{\itshape #1}}}}}
% \newcommand{\mq}[1]{\d #1}

% homogene Matrix
% \newcommand{\mq}[1]{{\mbox{{\fontencoding{T1}\sffamily{#1}}}}}

%\newcommand{\m}[1]{{\mbox{{\fontencoding{T1}\sffamily{#1}}}}}
\newcommand{\mq}[1]{{\q #1}}

\newcommand{\qindex}[1]{{{\bf #1}}}

% vec-Operator
\renewcommand{\vec}[0]{\mbox{vec}}

% lb-Operator
\newcommand{\lb}[0]{\mbox{lb}}

% FFT pair
\newcommand {\ftr}       {\circ\hspace{-3.5pt}-\hspace{-5pt}\bullet}

% trace-Operator
\newcommand{\trace}[0]{\mbox{tr}}

% rank-Operator
\newcommand{\rank}[0]{\mbox{rank} }

% Diag-Operator
\newcommand{\Diag}[0]{\mbox{Diag}}

% Allgemeine Matrix: #1: Anzahl der Spalten, #2: String für Elemente
% incl. //
\newcommand{\matr}[2]{%
\left( \begin{array}{*#1{c}}    #2   \end{array} \right) %
  }

% Allgemeine Determinante: #1: Anzahl der Spalten, #2: String für Elemente
% incl. //
\newcommand{\dete}[2]{%
\left| \begin{array}{*#1{c}}    #2   \end{array} \right| %
  }

%4 Vektor%
\newcommand{\vvector}[4]
  {    \left(
          \begin{array}{c}
            {#1} \\ {#2} \\{#3} \\ {#4}
          \end{array}
       \right) }

%5 Vektor%
\newcommand{\fvector}[5]
  {    \left(
          \begin{array}{c}
            {#1} \\ {#2} \\{#3} \\ {#4}\\ {#5}
          \end{array}
       \right) }

%%6 Vektor%
\newcommand{\svector}[6]
  {    \left(
          \begin{array}{c}
            {#1} \\ {#2} \\{#3} \\ {#4}\\ {#5}\\ {#6}
          \end{array}
       \right) }

%3 Vektor%
\newcommand{\dvector}[3]
  {    \left(
          \begin{array}{c}
            {#1} \\ {#2} \\{#3}
        \end{array}
       \right) }

%2 Vektor%
\newcommand{\zvector}[2]
  {    \left(
          \begin{array}{c}
            {#1} \\ {#2}
        \end{array}
       \right) }

%2 x 3 Matrix%
\newcommand{\zdmatrix}[6]
  {    \left(
          \begin{array}{ccc}
            {#1} & {#2} & {#3} \\
            {#4} & {#5} & {#6} 
          \end{array}
       \right) }

%2 x 4 Matrix%
\newcommand{\zvmatrix}[8]
  {    \left(
          \begin{array}{cccc}
            {#1} & {#2} & {#3} & {#4}\\
            {#5} & {#6} & {#7} & {#8}
          \end{array}
       \right) }

%3 x 2 Matrix%
\newcommand{\dzmatrix}[6]
  {    \left(
          \begin{array}{ccc}
            {#1} & {#2} \\
            {#3} & {#4} \\
            {#5} & {#6}
          \end{array}
       \right) }
%

%3 x 3 Matrix%
\newcommand{\dmatrix}[9]
  {    \left(
          \begin{array}{ccc}
            {#1} & {#2} & {#3} \\
            {#4} & {#5} & {#6} \\
            {#7} & {#8} & {#9}
          \end{array}
       \right) }

%2 x 2 Matrix%
\newcommand{\zmatrix}[4]
  {    \left(
          \begin{array}{cc}
            {#1} & {#2} \\
            {#3} & {#4}
          \end{array}
       \right) }

%Rotationmatrix R_i(a)%
\newcommand{\rot}[2]{\d{R}_{#1}(#2)}

%Rotationmatrix R^T_i(a)%
\newcommand{\rott}[2]{\d{R}\trans_{#1}(#2)}

%Determinante 2. Ordnung%
\newcommand{\zdet}[4]
  {    \left|
          \begin{array}{cc}
            {#1} & {#2} \\
            {#3} & {#4}
          \end{array}
       \right| }

%Determinante 3. Ordnung%
\newcommand{\ddet}[9]
  {    \left|
          \begin{array}{ccc}
            {#1} & {#2} & {#3}\\
            {#4} & {#5} & {#6}\\
            {#7} & {#8} & {#9}
          \end{array}
       \right| }

% Differentialgeometrische Bezeichnungen

% Gaussfunktion mit sigma
\newcommand{\Gs}{{G_{\sigma}}}

% nabla #f,
\newcommand{\nab}[1]{\nabla \! #1}

% average gradient of function #f, with sigma
\newcommand{\nabs}[1]{\nabla_{\!\!\sigma} #1}

% squared gradient of function #f
\newcommand{\sg}[1]{\dd {\Gamma} \! #1}

% average squared gradient of function #f
\newcommand{\asg}[1]{\overline {\dd {\Gamma} \! #1}}

% average squared gradient of function #f, with sigma
\newcommand{\assg}[1]{\overline {\dd {\Gamma}_{\!\!\sigma} #1}}

% Curvature of boundary, derived from assg, with sigma
\newcommand{\curv}[1]{\kappa_\sigma #1}

% Mathematische Blockbuchstaben
\newcommand{\E}{{\rm e}}

\newcommand{\mB}{\rm I\!B}
\newcommand{\mC}{\mathchoice {\setbox0=\hbox{$\displaystyle\rm
C$}\hbox{\hbox to0pt{\kern0.4\wd0\vrule height0.9\ht0\hss}\box0}}
{\setbox0=\hbox{$\textstyle\rm C$}\hbox{\hbox
to0pt{\kern0.4\wd0\vrule height0.9\ht0\hss}\box0}}
{\setbox0=\hbox{$\scriptstyle\rm C$}\hbox{\hbox
to0pt{\kern0.4\wd0\vrule height0.9\ht0\hss}\box0}}
{\setbox0=\hbox{$\scriptscriptstyle\rm C$}\hbox{\hbox
to0pt{\kern0.4\wd0\vrule height0.9\ht0\hss}\box0}}}
\newcommand{\mD}{\rm I\!D}
\newcommand{\mE}{\rm I\!E}
\newcommand{\mF}{\rm I\!F}
\newcommand{\mG}{\mathchoice {\setbox0=\hbox{$\displaystyle\rm
G$}\hbox{\hbox to0pt{\kern0.4\wd0\vrule height0.9\ht0\hss}\box0}}
{\setbox0=\hbox{$\textstyle\rm G$}\hbox{\hbox
to0pt{\kern0.4\wd0\vrule height0.9\ht0\hss}\box0}}
{\setbox0=\hbox{$\scriptstyle\rm G$}\hbox{\hbox
to0pt{\kern0.4\wd0\vrule height0.9\ht0\hss}\box0}}
{\setbox0=\hbox{$\scriptscriptstyle\rm G$}\hbox{\hbox
to0pt{\kern0.4\wd0\vrule height0.9\ht0\hss}\box0}}}
\newcommand{\mH}{\rm I\!H}
\newcommand{\mI}{\rm I\!I}
\newcommand{\mJ}{\mathchoice {\setbox0=\hbox{$\displaystyle\rm
J$}\hbox{\hbox to0pt{\kern0.4\wd0\vrule height0.9\ht0\hss}\box0}}
{\setbox0=\hbox{$\textstyle\rm J$}\hbox{\hbox
to0pt{\kern0.4\wd0\vrule height0.9\ht0\hss}\box0}}
{\setbox0=\hbox{$\scriptstyle\rm J$}\hbox{\hbox
to0pt{\kern0.4\wd0\vrule height0.9\ht0\hss}\box0}}
{\setbox0=\hbox{$\scriptscriptstyle\rm J$}\hbox{\hbox
to0pt{\kern0.4\wd0\vrule height0.9\ht0\hss}\box0}}}
\newcommand{\mk}{\rm I\!k}
\newcommand{\mK}{\rm I\!K}
\newcommand{\mL}{\rm I\!L}
\newcommand{\mM}{\rm I\!M}
\newcommand{\mN}{\rm I\!N}
\newcommand{\mO}{\mathchoice {\setbox0=\hbox{$\displaystyle\rm
O$}\hbox{\hbox to0pt{\kern0.4\wd0\vrule height0.9\ht0\hss}\box0}}
{\setbox0=\hbox{$\textstyle\rm O$}\hbox{\hbox
to0pt{\kern0.4\wd0\vrule height0.9\ht0\hss}\box0}}
{\setbox0=\hbox{$\scriptstyle\rm O$}\hbox{\hbox
to0pt{\kern0.4\wd0\vrule height0.9\ht0\hss}\box0}}
{\setbox0=\hbox{$\scriptscriptstyle\rm O$}\hbox{\hbox
to0pt{\kern0.4\wd0\vrule height0.9\ht0\hss}\box0}}}
\newcommand{\mP}{\rm I\!P}
\newcommand{\mQ}{\mathchoice {\setbox0=\hbox{$\displaystyle\rm
Q$}\hbox{\raise 0.15\ht0\hbox to0pt{\kern0.4\wd0\vrule
height0.8\ht0\hss}\box0}}{\setbox0=\hbox{$\textstyle\rm Q$}\hbox{\raise
0.15\ht0\hbox to0pt{\kern0.4\wd0\vrule height0.8\ht0\hss}\box0}}
{\setbox0=\hbox{$\scriptstyle\rm Q$}\hbox{\raise 0.15\ht0\hbox
to0pt{\kern0.4\wd0\vrule height0.7\ht0\hss}\box0}}{\setbox0=
\hbox{$\scriptscriptstyle\rm Q$}\hbox{\raise 0.15\ht0\hbox
to0pt{\kern0.4\wd0\vrule height0.7\ht0\hss}\box0}}}
\newcommand{\mR}{\rm I\!R}
\newcommand{\mS}{\mathchoice
{\setbox0=\hbox{$\displaystyle     \rm S$}\hbox{\raise0.5\ht0\hbox
to0pt{\kern0.35\wd0\vrule height0.45\ht0\hss}\hbox
to0pt{\kern0.55\wd0\vrule height0.5\ht0\hss}\box0}}
{\setbox0=\hbox{$\textstyle        \rm S$}\hbox{\raise0.5\ht0\hbox
to0pt{\kern0.35\wd0\vrule height0.45\ht0\hss}\hbox
to0pt{\kern0.55\wd0\vrule height0.5\ht0\hss}\box0}}
{\setbox0=\hbox{$\scriptstyle      \rm S$}\hbox{\raise0.5\ht0\hbox
to0pt{\kern0.35\wd0\vrule height0.45\ht0\hss}\raise0.05\ht0\hbox
to0pt{\kern0.5\wd0\vrule height0.45\ht0\hss}\box0}}
{\setbox0=\hbox{$\scriptscriptstyle\rm S$}\hbox{\raise0.5\ht0\hbox
to0pt{\kern0.4\wd0\vrule height0.45\ht0\hss}\raise0.05\ht0\hbox
to0pt{\kern0.55\wd0\vrule height0.45\ht0\hss}\box0}}}
\newcommand{\mT}{\mathchoice {\setbox0=\hbox{$\displaystyle\rm
T$}\hbox{\hbox to0pt{\kern0.3\wd0\vrule height0.9\ht0\hss}\box0}}
{\setbox0=\hbox{$\textstyle\rm T$}\hbox{\hbox
to0pt{\kern0.3\wd0\vrule height0.9\ht0\hss}\box0}}
{\setbox0=\hbox{$\scriptstyle\rm T$}\hbox{\hbox
to0pt{\kern0.3\wd0\vrule height0.9\ht0\hss}\box0}}
{\setbox0=\hbox{$\scriptscriptstyle\rm T$}\hbox{\hbox
to0pt{\kern0.3\wd0\vrule height0.9\ht0\hss}\box0}}}
\newcommand{\mU}{\mathchoice {\setbox0=\hbox{$\displaystyle\rm
U$}\hbox{\hbox to0pt{\kern0.4\wd0\vrule height0.9\ht0\hss}\box0}}
{\setbox0=\hbox{$\textstyle\rm U$}\hbox{\hbox
to0pt{\kern0.4\wd0\vrule height0.9\ht0\hss}\box0}}
{\setbox0=\hbox{$\scriptstyle\rm U$}\hbox{\hbox
to0pt{\kern0.4\wd0\vrule height0.9\ht0\hss}\box0}}
{\setbox0=\hbox{$\scriptscriptstyle\rm U$}\hbox{\hbox
to0pt{\kern0.4\wd0\vrule height0.9\ht0\hss}\box0}}}
\newcommand{\mZ}{{\sf Z\hspace{-1.8ex}Z}}

\newcommand{\mone}{\mathchoice {\rm 1\mskip-4mu l} {\rm 1\mskip-4mu l}
{\rm 1\mskip-4.5mu l} {\rm 1\mskip-5mu l}}
%

%%% Local Variables: 
%%% mode: latex
%%% TeX-master: t
%%% TeX-master: t
%%% End: 

\newcommand{\todo}[1]{{\textbf{\color{red}[TO-DO] #1}}}
\begin{frontmatter}

\title{Marine Bubble Flow Quantification Using Wide-Baseline Stereo Photogrammetry}

\author{Mengkun She, Tim Wei\ss, Yifan Song, Peter Urban,\\ Jens Greinert, Kevin K{\"o}ser}%\fnref{myfootnote}}
\address{GEOMAR Helmholtz Centre for Ocean Research Kiel, Wischhofstr. 1-3, 24148 Kiel}

% \address[mymainaddress]{ GEOMAR Helmholtz Centre for Ocean Research Kiel, Kiel, Germany\\
%	Tel.: ++49 431 600 2595\\}

\begin{abstract}
Reliable quantification of natural and anthropogenic gas release (e.g.\ CO$_2$, methane) from the seafloor into the water column, and potentially to the atmosphere, is a challenging task. 
While ship-based echo sounders such as single beam and multibeam systems allow detection of free gas, bubbles, in the water even from a great distance, exact quantification utilizing the hydroacoustic data requires additional parameters such as rise speed and bubble size distribution. 
Optical methods are complementary in the sense that they can provide high temporal and spatial resolution of single bubbles or bubble streams from close distance.
In this contribution we introduce a complete instrument and evaluation method for optical bubble stream characterization targeted at flows of up to 100ml/min and bubbles with a few millimeters radius.
The dedicated instrument employs a high-speed deep sea capable stereo camera system that can record terabytes of bubble imagery when deployed at a seep site for later automated analysis. Bubble characteristics can be obtained for short sequences, then relocating the instrument to other locations, or in autonomous mode of definable intervals up to several days, in order to capture bubble flow variations due to e.g. tide dependent pressure changes or reservoir depletion.
Beside reporting the steps to make bubble characterization robust and autonomous, we carefully evaluate the reachable accuracy to be in the range of 1-2\% of the bubble radius and propose a novel auto-calibration procedure that, due to the lack of point correspondences, uses only the silhouettes of bubbles. 
The system has been operated successfully in 1000m water depth at the Cascadia margin offshore Oregon to assess methane fluxes from various seep locations. Besides sample results we also report failure cases and lessons learnt during deployment and method development.
\end{abstract}


\end{frontmatter}

%\linenumbers

\section{Introduction}
\label{sec:introduction}
Greenhouse gases such as methane or CO$_2$ play a key role in climate change.
At the ocean floor they can escape from natural reservoirs\cite{suess14coldseeps,beaubien2013panarea}, from leaky or abandoned well\cite{VIELSTADTE2015848}  or carbon storage sites, and both gases participate in or result from microbial/biological metabolism as well as thermo-chemical reactions\cite{WHITICAR1999291} as part of geological processes.
Gas release at so called seep sites, particularly in concert with the formation and dissociation of gas hydrates, can also influence the mechanical stability of seafloor, e.g. at continental slopes. In addition, due to the gas release, seep sites can have an important impact on the local habitats and create an oasis-like ecosystem in the deep sea\cite{levin01influencesbiodiversity}.
For all these reasons, exact quantification and monitoring of gas release from the ocean floor, as well as understanding the controlling conditions, are important research questions.
\begin{figure}[!t]
	\begin{center}
		\includegraphics[height=45mm]{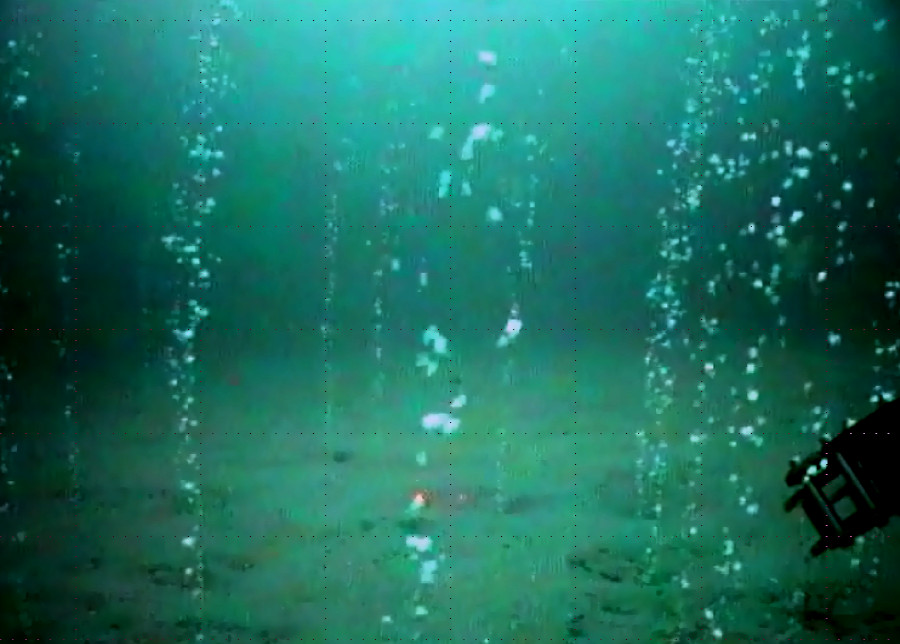}	 
		\includegraphics[height=45mm]{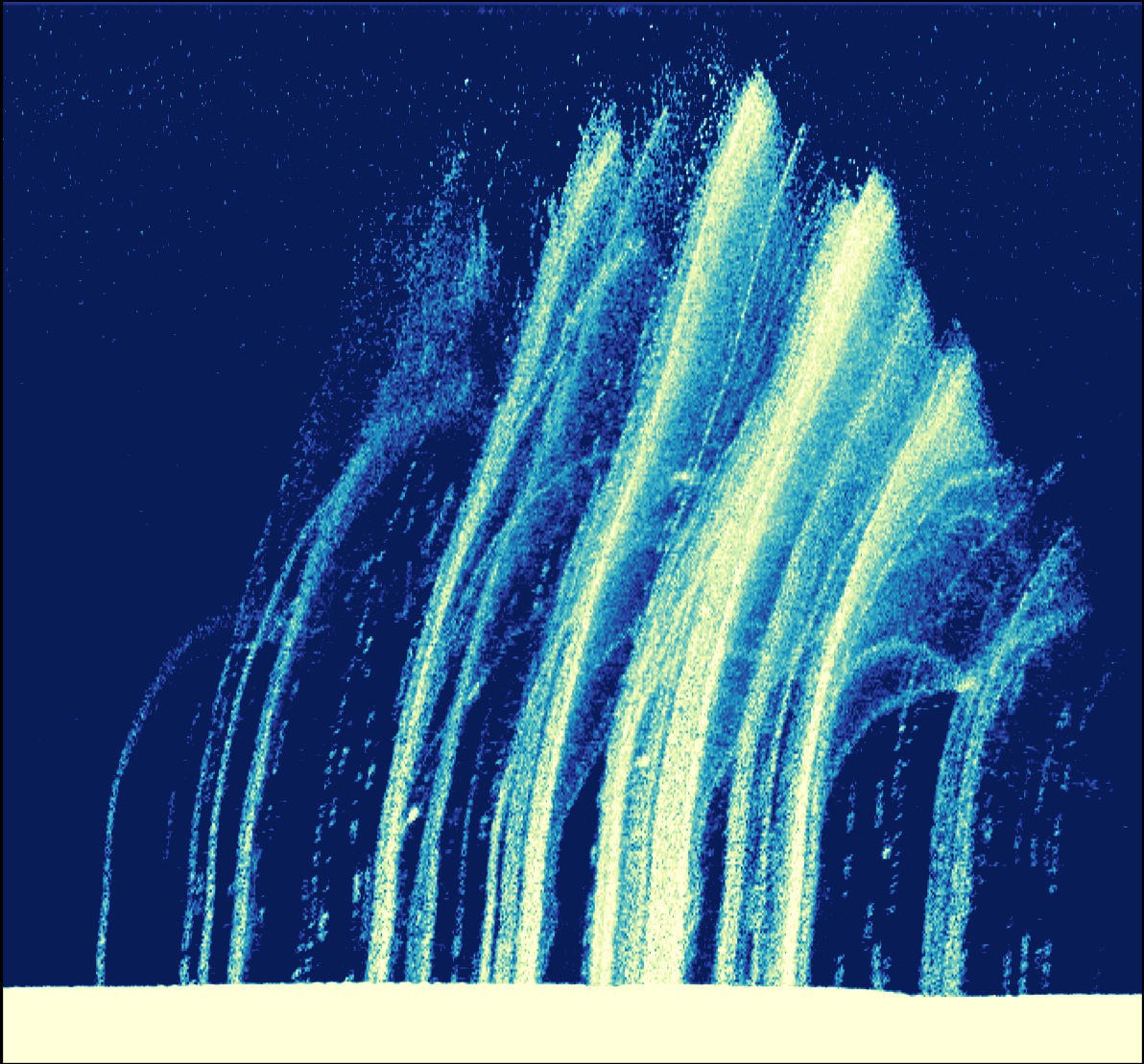}
	\end{center}
	\caption{Methane escaping from the seafloor in the North Sea and forming bubble streams (left: horizontally looking ROV camera. right: multibeam echo sounder). The visual sensor can be positioned on top of one of these streams to capture the stream parameters such as rising speed and the bubble size distribution. Such streams can be detected also in ship-based echo sounders\cite{urban2017processing} as can be seen in the right image, which is stacked from multiple subsequent pings of a multibeam echo sounder traversing over the bubble stream field. Jointly, optical and acoustical observations can be used for large scale quantification.  \label{fig:bubblestreams} }
\end{figure}

When released continuously from the seabed at relatively shallow depth ($<$100m), bubbles containing methane can rise towards the sea surface and potentially transport small amounts of this potent greenhouse gas into the atmosphere\cite{mcginnis06fateofmethanebubbles,Shakhova10methaneventingarctic,wang2020dynamics}, see Fig. \ref{fig:bubblestreams}.

\begin{figure}[!t]
	\begin{center}
		\includegraphics[height=4cm]{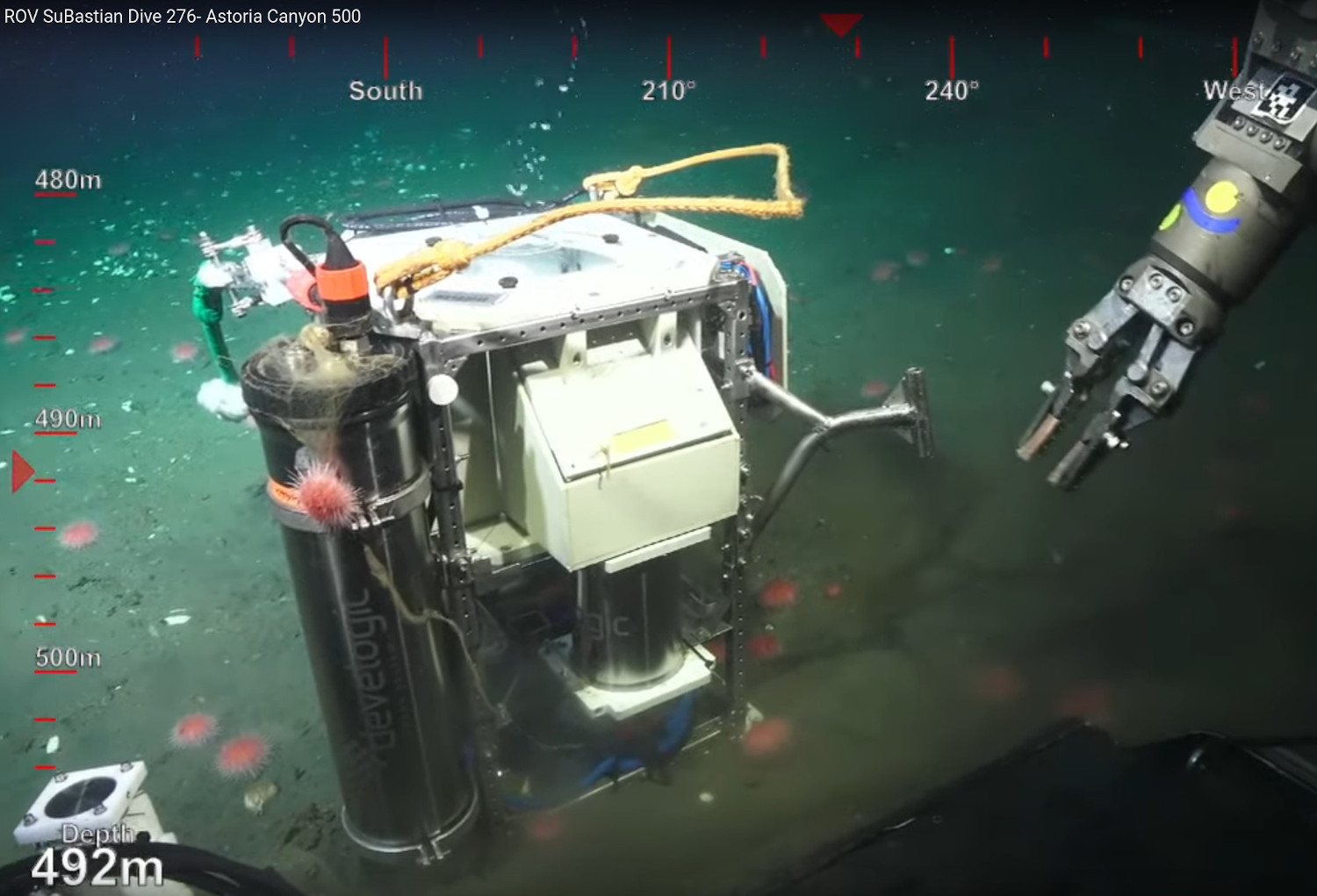}
		\includegraphics[height=4cm]{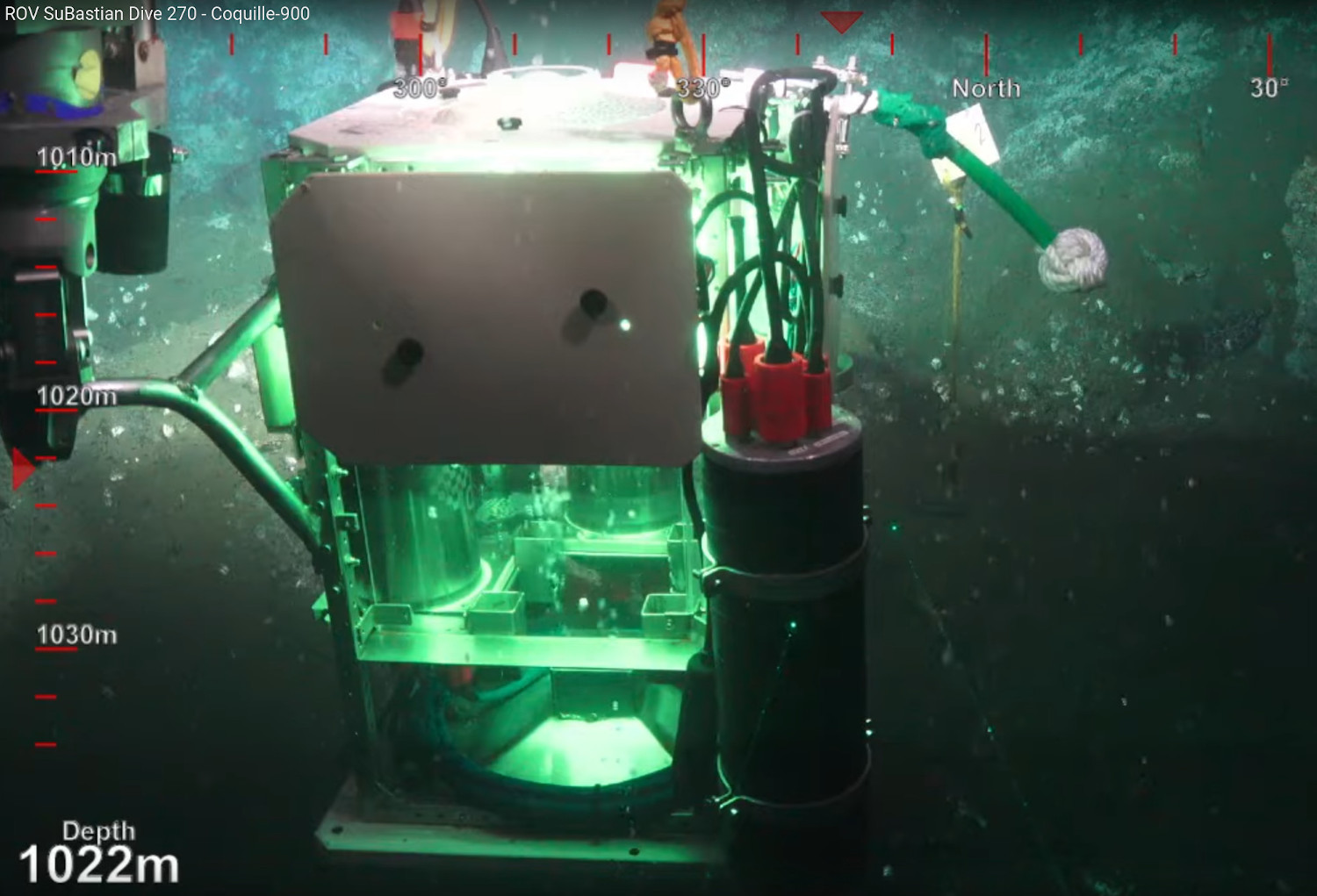}			
	\end{center}
	\caption{Instrument deployed by a remotely operated vehicle (ROV) in 500m and 1km water depth in the Pacific Ocean offshore Oregon during Falkor cruise FK190612 'Observing Seafloor Methane Seeps at the Edge of Hydrate Stability' (see also https://youtu.be/fxiD75qmQFU for a recording of the entire live stream captured by ROV SuBastian, Schmidt Ocean Institute). The instrument is positioned on the seafloor over a seep spot where methane bubbles escape;  bubbles rise through the instrument and are photographed by a wide baseline stereo camera setup at high frame rates for photogrammetric characterization of size, shape, rise speed and number.}
	\label{fig:falkor_bbx}
\end{figure}

Key quantification techniques for free gas release/bubbles include physical trapping of the gas with funnels, remote and in-situ acoustical methods and direct optical observations, all of which have advantages and disadvantages.
Measuring gas emissions with a funnel or catcher (e.g. \cite{Nikolovska08hydroacoustic}) over short time is a challenging and costly task as it usually requires remotely operated vehicles (ROVs) or divers to perform the mission.
Long-term automated monitoring of gas release over a large seep area using such a technique is therefore impractical. Since only the accumulated volume is measured, the funnel method does not provide detailed information on the individual bubbles; if the number of bubbles is counted, an average size could be calculated.
Passive acoustic methods \cite{leighton_quantification_2012,berges_passive_2015,li_passive_2020} essentially listen to the sounds created when bubbles detach from the seafloor and can quantify bubble sizes using multiple hydrophones at close range.
Active acoustic-based approaches can detect free gas underwater from a large distance and are currently probably the most efficient tools to find gas release locations in lakes and oceans\cite{ greinert2004hydroacoustic,greinert2008monitoring, merewether1985acoustically,von2010acoustic} and to map gas activity over large areas\cite{urban2017processing,HIGGS2019105985,colbo_review_2014,dupre_tectonic_2015,romer_assessing_2017}.
However, an exact quantification of the gas flow rate requires prior knowledge of some essential parameters such as the bubble size distributions \cite{veloso2014methane, veloso2015new} or bubble shape\cite{leblond_acoustic_2014}. Recently developed broadband methods\cite{weidner_wideband_2019,li_broadband_2020} aim at relaxing the need for external measurements of these parameters but the majority of 
%by using either specific split-beam echo sounders \cite{weidner_wideband_2019} or a combination of multiple echo sounders\cite{li_broadband_2020} in order to achieve broadband. 
quantitative inverse-hydroacoustic methods need good knowledge about the bubble size and rising speed distribution within the monitored volume of water.

Cameras on the other hand cannot be used to sense bubbles from a large distance and therefore require in-situ deployment; they cannot directly sense gas flow in large areas. However, photographing bubble streams with high speed cameras enables to derive many bubble characteristics and thus, in combination with active acoustics, facilitates the acoustic inversion to derive spatial estimates of the free gas flow in oceans and lakes.
In addition, visual information provides the only way for a better understanding of the bubble behavior, such as deformation and motion patterns through the water column\cite{jordt2015bubble,thomanek2010automated}.

With respect to optical bubble studies, early works focused on quantifying flow of bubbles in a laboratory setup\cite{bian2011reconstruction,bian20133d,zelenka2014gas,xue2013matching,jordt2015bubble}, where imaging and instrumentation conditions are much better controlled than in the real ocean.
 Fewer studies exist that actually design and deploy in-situ imaging systems for the ocean, including telecentric lenses for very small bubbles \cite{Leifer_03Bubbles}, a monocular camera \cite{thomanek2010automated} and a small baseline stereo system \cite{wang2015deep}. 
 All of these approaches required substantial manual interaction to obtain bubble information, and their focus was not on robust, automatic techniques that can work on gigabytes of image data.
Consequently, extracting the bubble information from the video sequences accurately and robustly remains a challenging task and a complete and robust pipeline is required to automatically analyze many thousands of images that are obtained under uncontrollable, and often sub-optimal, conditions.

In this paper, we build on our preliminary laboratory studies for bubble measure\-ments\cite{jordt2015bubble} and present the following advancements as novel contributions: We (i) present a deep-sea wide baseline bubble measurement system, including a robust and complete processing pipeline that is able to automatically characterize bubble streams using long-term video sequences. We (ii) introduce a new silhouette-based auto-calibration approach that can adjust the calibration without point correspondences, i.e. only from bubble observations. We show that the same technique can be used for accurate bubble ellipsoid estimation. We (iii) carefully evaluate the system performance using ground truth measurements and show results and challenges on real data acquired from deep sea missions down to 1000m water depth.

\section{Related Work}
\label{sec:related_work}
\paragraph*{3D Measurement of Bubbles}
Quantifying multi-phase flow parameters (as observed in interactions of liquids and gases) has been a topic of interest for natural and industrial applications, and Particle Tracking Velocimetry (PTV) is a common technique to tackle such problems\cite{maas1993particle}.
The gas/liquid flow is the basic scenario of two-phase flow, where bubbles can be identified in the image sequences and the bubble features can be extracted using 3D image processing techniques.
Therefore, in some laboratory-based works, special setups were built to photograph bubbly flow in a tube with high-speed cameras.
These approaches generally consist of three major steps which include bubble identification, bubble 3D size measurement and bubble tracking over time.
Zelenka et al. and Fu et al. \cite{zelenka2014gas,fu2016development} have focused on the bubble outline extraction, but both use a single camera.
Bian et al. \cite{bian2011reconstruction,bian20133d} assume that the bubble shape resembles two half-ellipsoids and thus proposed an approach to extract the characteristic parameters of a single rising bubble from a pair of stereo images.
In this respect a stereo camera system can significantly improve the 3D bubble size estimation but also poses additional challenges, for instance, finding the correspondence of the same bubble in the stereo image pair:
Bubbles are not easily distinguishable, and do not have rich texture information, consequently, traditional feature-based (e.g. SIFT\cite{Lowe-2004-Features}) or pixel-wise (e.g. SGM\cite{hirschmuller2007stereo}) matching approaches can hardly be applied in this scenario.
This is why laboratory approaches can often only reconstruct a single rising bubble.
To save the second camera, and to avoid synchronization while sacrificing image area, Xue et al. \cite{xue2012three,xue2013matching} have constructed a camera-mirror system, where the mirrors are used to generate a reflection of the bubble stream as seen from a different perspective.
Xue et al. also proposed an equal-height heuristic to disambiguate multiple match candidates across the stereo image pair, making strong assumptions on bubble position.
In a laboratory environment, many complex setups can be built to better reconstruct the shape of the bubble. For instance, Fu et al. \cite{fu20183d} have developed a space carving algorithm to reconstruct the free form surface of a single large rising bubble using multiple cameras.
A similar system is developed by Masuk et al. \cite{masuk2019robust} but with cameras and mirrors to create 4 virtual views for the space carving algorithm.
These systems and algorithms are generally attractive, but it can be difficult to transfer complicated or delicate setups into in-situ deep sea bubble stream characterization systems that can be used reliably in the ocean. Here, instruments have to be compact and robust to be able to be transported over large distances (e.g. containers) and deployed by heavy gear in harsh weather conditions. When a system is operated e.g. by an ROV at 1000m depth (100 bar water pressure), possibilities for online recalibration are very limited. Still, a few systems have been proposed to optically measure bubbles in the ocean, as discussed in the next section.

\paragraph*{Optical Bubble Measurement Systems for the Ocean}
In case no special measurement devices are available, it is common practise in oceanography to place reference objects into the bubble stream or close to the stream in order to obtain a rough estimate of the bubble sizes\cite{Nikolovska08hydroacoustic,SCHNEIDERVONDEIMLING2011867}.

Leifer et al.\cite{Leifer_03Bubbles} have presented a deep sea optical bubble meter to analyze sizes and motions of bubbles with a single camera, and a manual workflow to extract the bubble volume according to the ellipsoid projection assumption. Using a telecentric lens avoids scale ambiguity, but restricts the observation space to an extremely small volume.

Later, Sahling et al.\cite{sahling2009vodyanitskii} develop an optical device and apply it to measure the gas discharge of a bubble stream using a perspective lens from a camera mounted on an ROV.
Thomanek et al. \cite{thomanek2010automated} improved this system in terms of hardware design and proposed a more complete image processing workflow to extract the bubble sizes and rising speeds and turn them into gas flux estimation.
Both authors assume the distances of the bubbles to the camera to be constant, hence, the pixel-to-object scale can be calibrated using a ruler or reference.
Unfortunately, the bubbles in the ocean often rise in a helical way \cite{wu2002experimental} or the entire stream can bend with currents, consequently, the object distances can vary significantly.
Also, large bubbles are typically not spherical and the extent of the bubble in the camera's viewing direction cannot be observed then.
Therefore, monocular systems can suffer from relatively large uncertainty when measuring the size and shape of a bubble.
To address these issues Wang et al.\cite{wang2015deep} propose a short-baseline stereo camera system to obtain more 3D information, and later successfully deploy the system in an expedition\cite{wang2016observations}.

\paragraph*{Predecessor of this Work and Contributions}
In preparatory studies for this work, Jordt et al. \cite{jordt2015bubble} geometrically analyzed the uncertainty of the triangulation estimation in a short-baseline stereo setup and proposed a $90^\circ$ wide-baseline stereo camera setup\footnote{A wide baseline stereo setup describes a stereo camera system where the baseline (the distance of the cameras) is not significantly smaller as compared to the distance to the  objects. This is in contrast to small baseline scenarios where the cameras are often close to one another and observe a scene that is relatively far away.} and proved the bubble measurement feasibility by laboratory experiments. 
This configuration can observe both the frontal and the side view of the bubble to remedy shape and distance uncertainty.
The work also includes an extensive discussion of e.g image pre-processing, camera calibration considering flat port refraction \cite{agrawal2010analytical,JordtSedlazeck_2012RefCalibUnw}, bubble stereo matching and a 3D ellipsoid fitting approach for optical analysis of the bubble stream.
These proof of concept studies form the basis for our actual instrument that has been developed over the past years.
During the development, She et al. \cite{she2020considering} analyzed replacement of the flat port camera housings with dome ports, and discussed centering and calibration of the dome port cameras  \cite{she2019adjustment,she2021refractive}.
In this contribution we build on top of all of these studies and describe a practical deep ocean bubble measurement system, the Bubble Box (BBox) that, positioned on a bubble seep spot, captures bubbles using stereo photogrammetry. 
The Bubble Box system is robust against inevitable nuisances such as dirt, disturbed illumination from upstirred sediment, small offsets of the bubble stream, wobbly bubble ascent or temporal shift of the stream position. 
We provide automated methods for synchronization and robust background removal, as well as a completely new approach to instrument self-calibration without point correspondences: we use only the \emph{outlines} of bubbles for wide-baseline recalibration from in-situ data.
While previous systems have been qualitatively validated, we show a reconstruction accuracy in good conditions in the range of 1\% using ground-truthed observations.

\section{System Design}
\subsection{Hardware Overview}
\begin{figure}[!t]
	\begin{center}
		\subfloat[Schematic System Oerview]{
        	\includegraphics[width=0.6\textwidth]{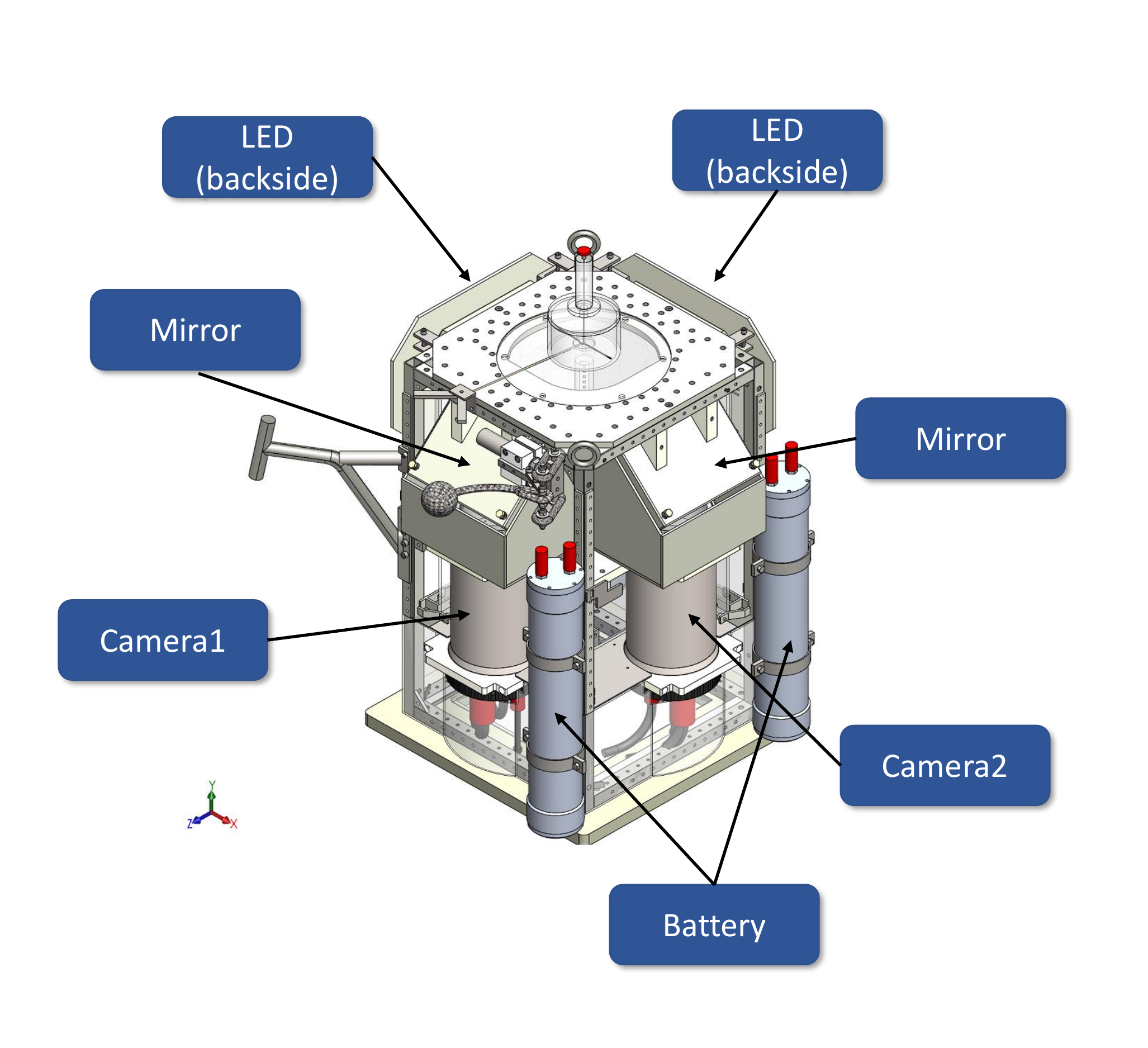}
	    }
		\subfloat[System in Test Tank] {
			\includegraphics[width=0.37\textwidth]{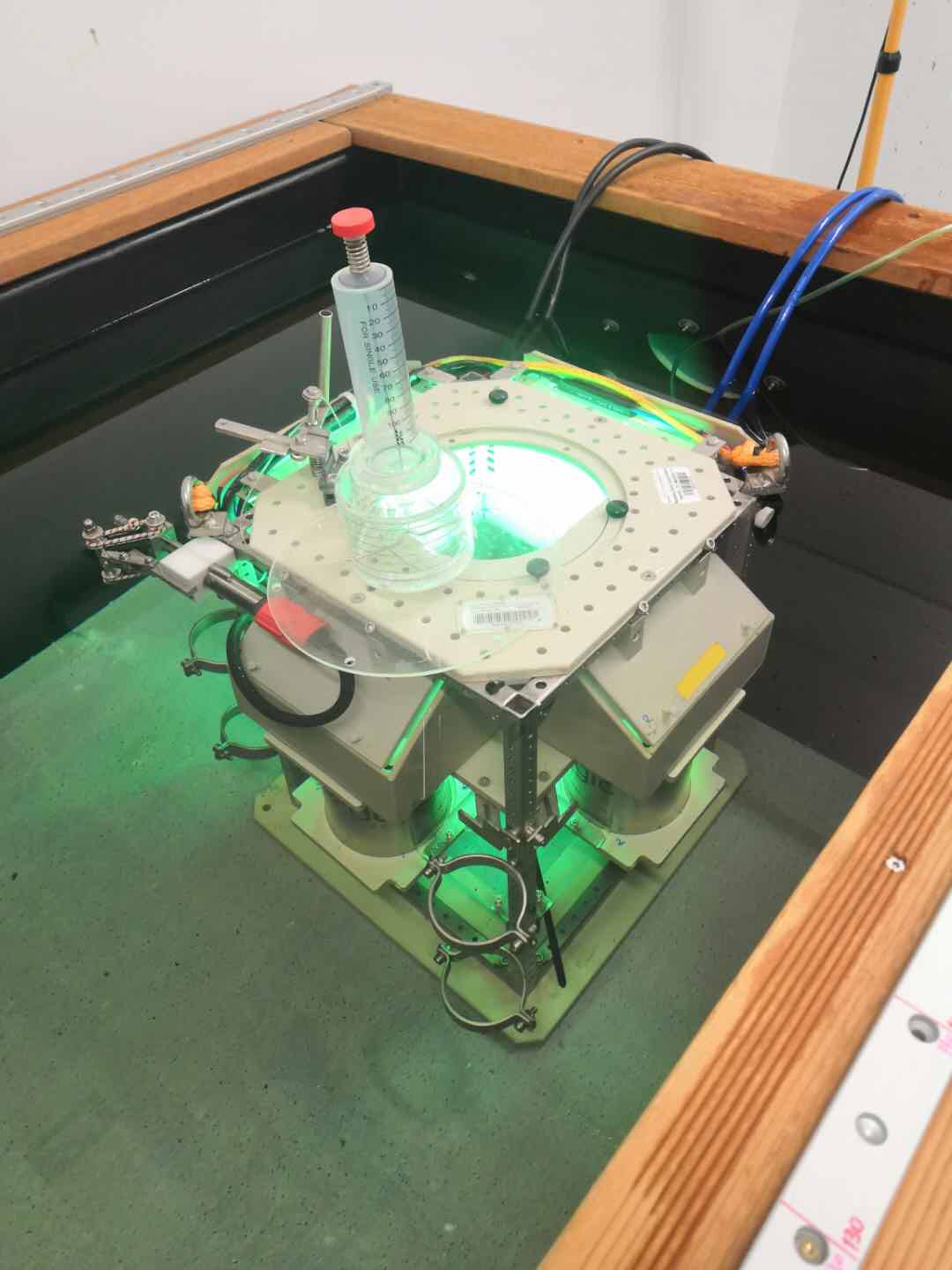}
		}
	\end{center}
    \caption{(a) System overview and its technical description. (b) The system in an 80cm deep test tank.}
	\label{fig:bbx_system}
\end{figure}

Before introducing the gas flow quantification approach, we first summarize the deep sea in-situ bubble stream characterization instrument.
The instrument follows the wide-baseline setting proposed in \cite{jordt2015bubble} and is a box shaped stereo recording device that can be deployed by a robot arm of an ROV as shown in Fig. \ref{fig:falkor_bbx} or can also be lowered from a surface vessel and then be positioned by divers.
It contains a vertical corridor (64cm$^2$ cross section) in the box center which allows bubble streams to rise through and escape through a hole in the top lid. 
Two deep sea titanium housings with dome ports are mounted at two adjacent sides of the box.
A schematic system overview can be seen in Fig. \ref{fig:bbx_system}. 
Inside each of the housings, there is a high-speed machine vision camera recording images of up to $1024\times800$ pixels resolution and a field of view of around $33^\circ$.
The frame rate of the camera can be set to $80Hz - 100Hz$ (or even faster if smaller image areas are used). Considering a typical bubble rise speed of ~$25-35cm/s$ \cite{leifer2002bubble} each bubble is photographed approximately 40 times until it leaves the vertical field of view of the camera.
While both gas and water are largely transparent, the cameras can observe refraction and reflection effects at the bubble surface; Previous studies\cite{rehder02bubblelifetime,jordt2015bubble} have thereby shown that background illumination (also known as bright-field illumination in microscopy) is advantageous for highlighting these effects and photographing bubbles.

\begin{figure}[!t]
	\begin{center}
			\includegraphics[width=1.0\textwidth]{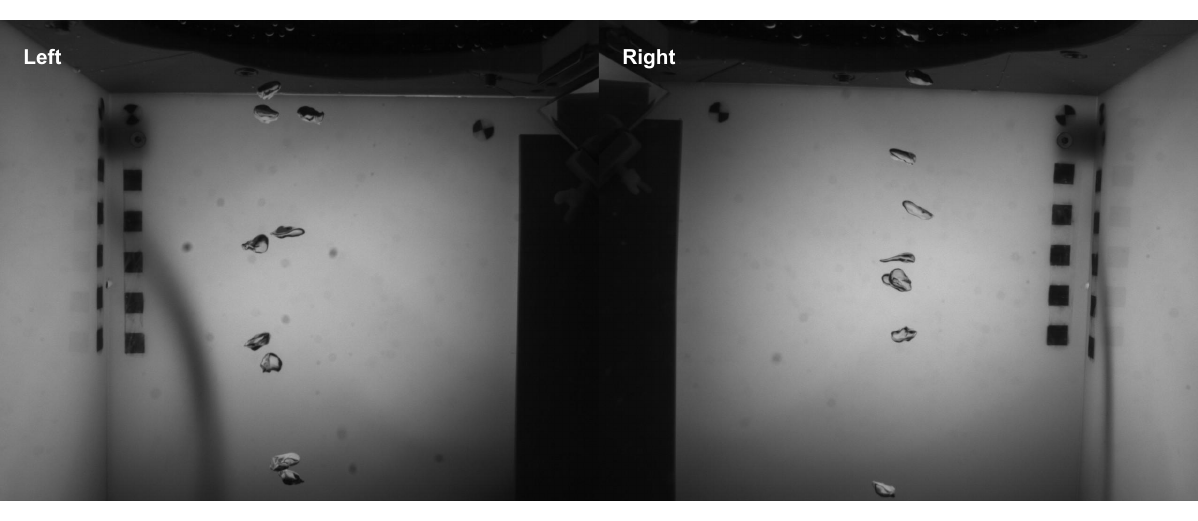}
	\end{center}
	\caption{Sample stereo image pair captured by the instrument. The left image and the right image are concatenated horizontally (If not mentioned particularly, in the later figures, the left image and the right image are shown concatenated).}
\label{fig:sample_image_pair}
\end{figure}

Since blue light suffers from strong scattering and red light from strong absorption, we chose to use green light for the illumination in combination with a (wideband) gray camera without mosaic filter.
Limiting the illumination to almost monochromatic light (or a narrow band of wavelengths) aims at minimizing dispersion effects, and thus potential blur, at the bubble silhouette.
%Therefore, we choose green LEDs as the illumination source (in combination with a white-band gray camera without mosaic filter).
Therefore, two panels of green LEDs (550nm) are mounted at the camera-opposite faces of the box behind acrylic diffusor plates. They provide back-light illumination such that the outline of the bubble produces a dark rim in the image. Since each bubble is observed from 90$^\circ$ different perspectives, no photometric properties such as color or texture can be used for matching, and the outline in one camera only provides a weak hint about the size of the outline in the other camera and no explicit point correspondences can be obtained.
A sample stereo image pair is shown in Fig. \ref{fig:sample_image_pair}, where the left image and the right image are concatenated horizontally.

\subsubsection{Overall System Design}
\begin{figure}[!t]
	\begin{center}
		\includegraphics[width=0.98\columnwidth]{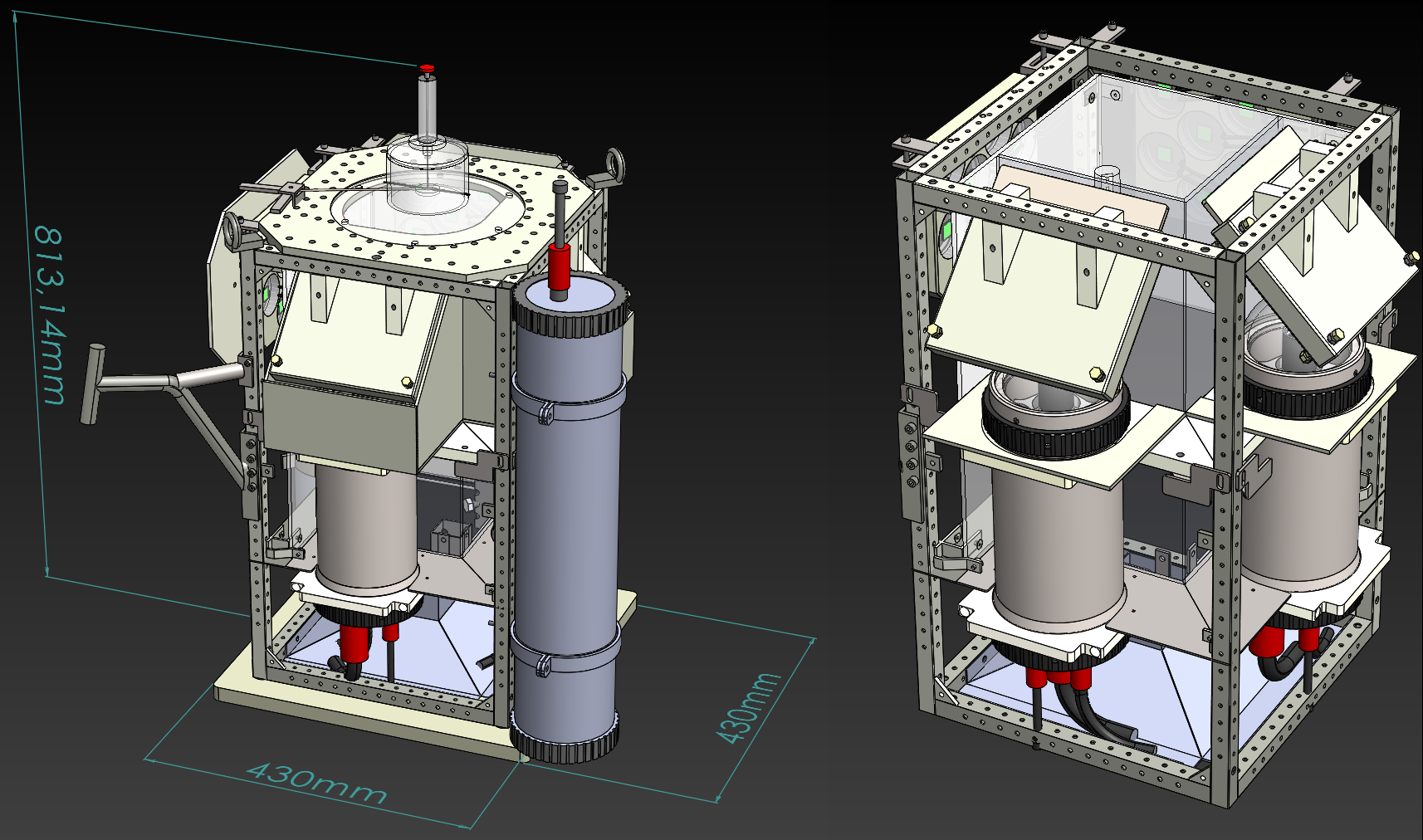}				
	\end{center}
	\caption{Left: Dimensions of instrument, including a removable collector funnel at the top. Right: virtual view into box, with some covers removed. \label{fig:bboxdrawings1} }.
\end{figure}

The system size is 81cm (height, without funnel: 62cm) $\times$ 43cm $\times$ 43cm (see Fig. \ref{fig:bboxdrawings1}). The weight is about 60kg in air and 24kg in water. The box was designed to work at typical depths down to 2000m, but could be updated easily for use in down to 6000m, for which the main components, like the camera housings, are already built. The overall power consumption during 80Hz recording is 70W on average (100W peak during flash). It can be powered remotely or by battery, where the batteries last for 5h-10h of continuous flashing, depending on water temperature. The system can be switched on and off using a heavy-duty switch that can be operated by the ROV. Recording time can be extended as explained in the next paragraph.

\subsubsection{Computers and Synchronization}
\begin{table*}
	\begin{center}
		\caption{Datasheet of the dome port camera systems.}
		\label{tab:cam_sheet}
		\begin{tabularx}{1.0\textwidth}{ccccc}
			\hline\noalign{\smallskip}
			Camera & Image& Pixel & Sen- & Shut-\\
			model & size[px]& size[$\mu$m]& sor & ter\\
			\hline\noalign{\smallskip}
		 	acA1300-200um & 1280 x 1024 & 4.8 x 4.8 & CMOS & Global\\
			\noalign{\smallskip}\hline
			Expo- & Lens & Dome & Dome & Dome \\
			sure[ms] &  & model& radius[cm] & thickness[mm]\\
			\hline\noalign{\smallskip}
			1 & AZURE 8mm C-Mount & Vitrovex & 5 & 7\\	
		    \noalign{\smallskip}\hline
		\end{tabularx}
	\end{center}
\end{table*}
For avoiding high speed data cables outside pressure housings and in the ocean water, as well as compactness and robustness considerations when the system is deployed in more than 1000m water depth, we use two separate housings for the cameras, and each housing contains a dedicated computer to store the images.
Therefore, the BBox holds two pressure housings that each contain an intel NUC computer and a Basler Ace acA1300-200um gray machine vision camera behind a dome port. The 8mm C-Mount lenses (AZURE) have been carefully centered in the domes \cite{she2019adjustment,she2021refractive}.
The datasheet of the cameras can be seen in Table \ref{tab:cam_sheet}.
 
We crop the images to 1024$\times$800 pixels to avoid frame-drops when writing to the 1TB SSDs inside each housing at high frame rates. 
%\todo{Kevin: actually, we should maximize horizontal FOV in the future, but we can potentially reduce vertical fov by half, then even go back to 100Hz ?}
A trigger signal for the entire system is provided by a microcontroller inside one of the camera housings. 
The rising edge of this signal starts the camera exposure time and the background LED flashes; 
the falling edge stops exposure and deactivates the background LEDs. Exposure time is set to 1ms to avoid motion blur. 
Every 5000 images, exposure and flash time is set to 10 microseconds only, which produces a black image in both cameras. 
These \emph{black flash} images are exploited to synchronize the image streams from the two independently recording cameras in post processing. 
Before operation, the computer clocks are synchronized via network time protocol up to one second and the time stamp of the incoming image is written into the filename of the raw image and saved as pgm.
At 80Hz, the stereo system currently produces a data rate of 1 gigabit per second or 0.45 terabyte per hour.
The images can be downloaded after the mission via gigabit ethernet which takes approximately the same time as needed for recording.

\begin{figure}[!h]
	\begin{center}
		\includegraphics[height=43mm]{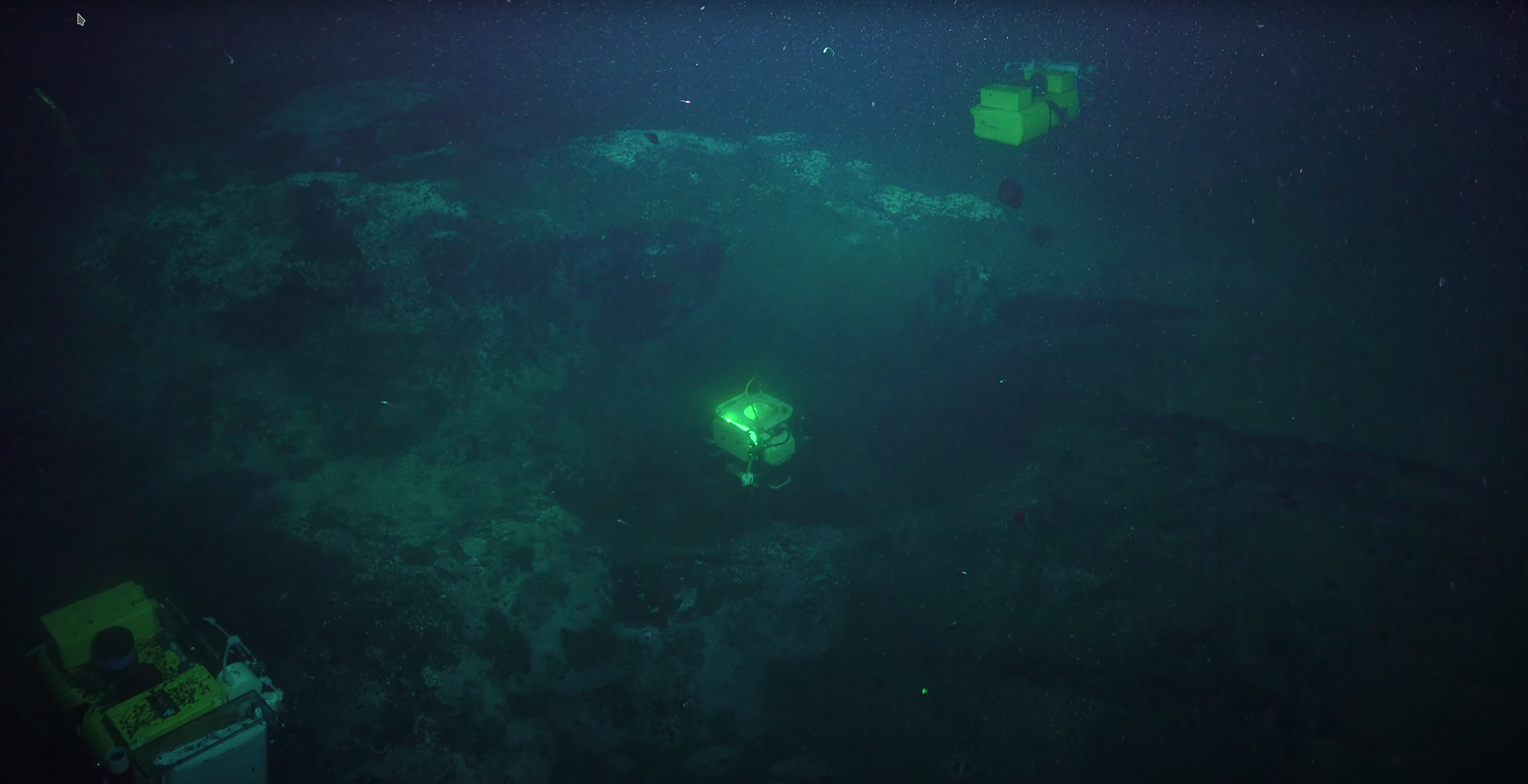}
		\includegraphics[height=43mm]{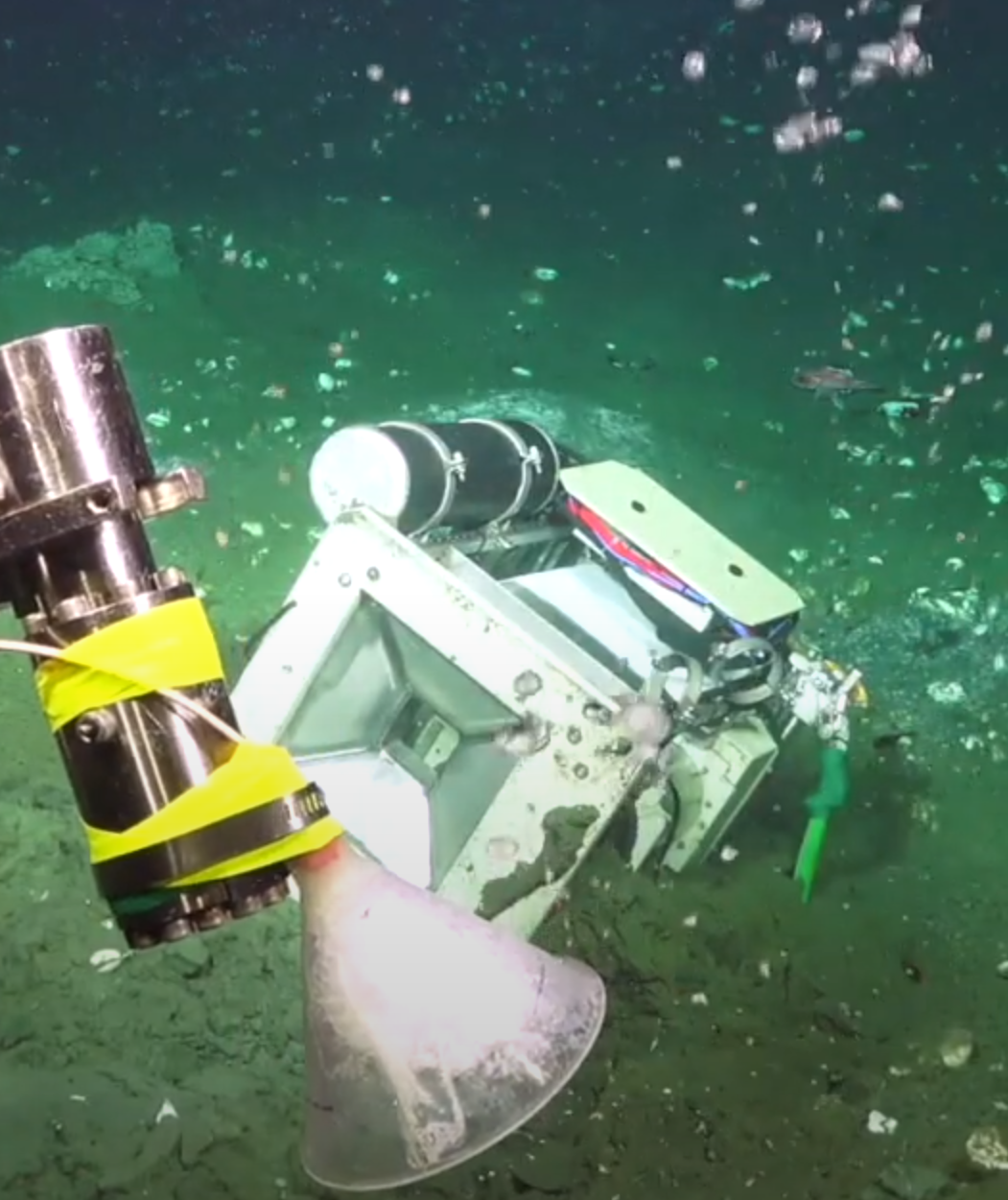}
	\end{center}
	\caption{Left: In long-term standby mode, the LEDs flashes can be helpful for finding the instrument in the total darkness of the deep sea. Right: In uneven terrain the deployment of the BBox by the robot arm fails and the instrument falls over in the sediment. It had to be picked up again for another attempt to position it at the seep spot. Both images captured by Schmidt Ocean Institute's ROV SuBastian.}
\label{fig:green_light_in_dark}
\end{figure}

The BBox is equipped with a long-term-mode that modifies the microcontroller's trigger signal. 
Different to the continuous mode, the microcontroller can create recording intervals such as the first five minutes of each hour. 
Outside these intervals LEDs are only triggered once every few seconds as a standby visualization and image data is disregarded.
This is useful system state information but also proved helpful for finding the instrument back in the total darkness of the deep sea (see Fig. \ref{fig:green_light_in_dark} left). The interval mode reduces the power- and storage needs by a large factor and allows to operate the box for more than one day\cite{veloso2022bboxgquant}.

\subsubsection{Cameras and Mirrors}

\begin{figure}[!t]
	\begin{center}
		\includegraphics[width=0.98\columnwidth]{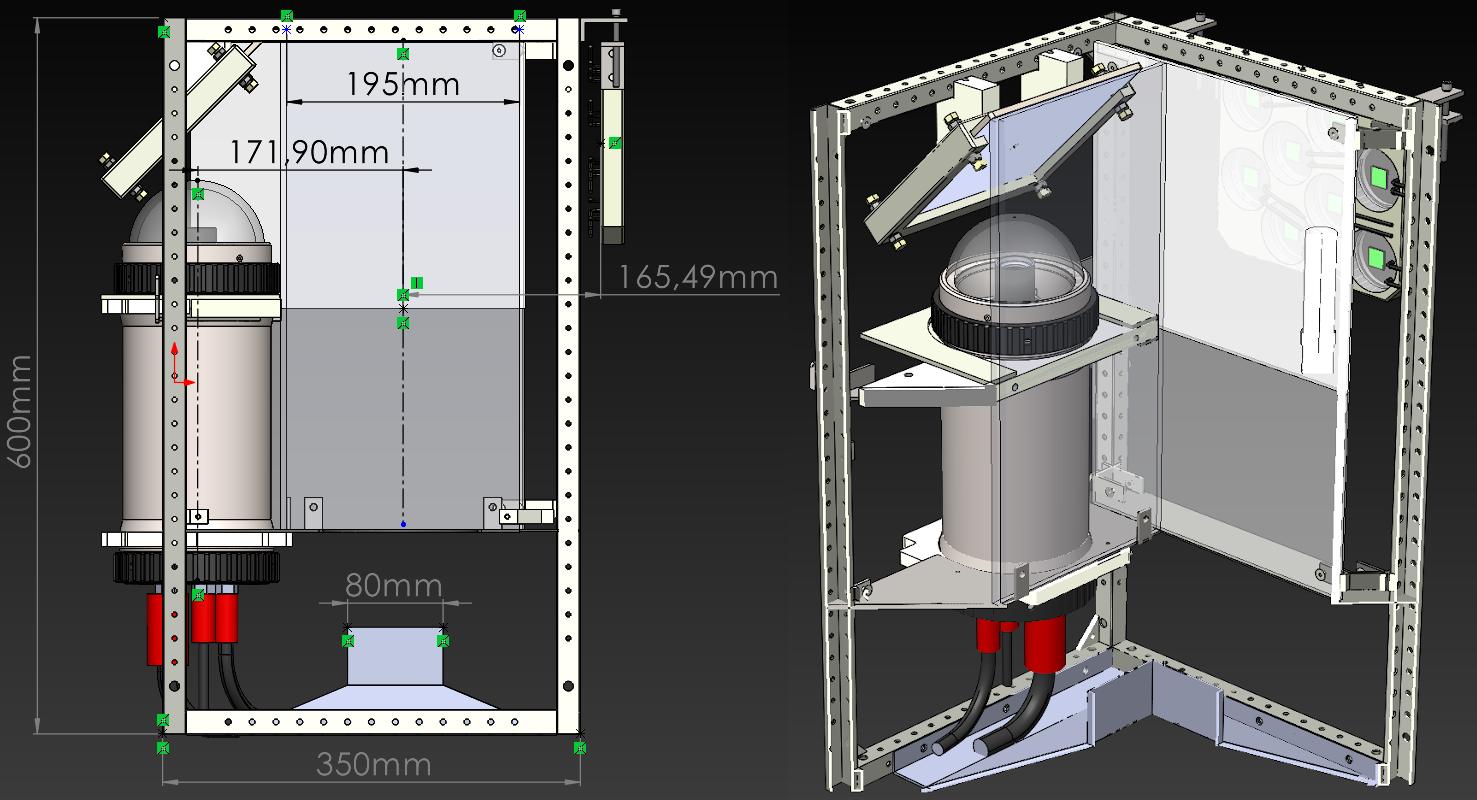}					\end{center}
	\caption{Left: Vertical section of the BBox showing the entrance funnel at the bottom leading to the 80mm wide square-shaped rise corridor. Right: One of the cameras inside the dome port looking through a mirror into the box. Additionally, the background illumination for the other camera (not displayed here) using LEDs behind an acrylic plate is displayed.\label{fig:bboxdrawings2} }.
\end{figure}

As can be seen in the bottom of Fig. \ref{fig:bboxdrawings2} (left), the bubble rise corridor into which the bubbles are directed at the bottom is 8cm wide.
The effective resolution is 5.7 pixel per millimeter in the center of that observation corridor.
Since the cylindrical camera housings are elongated, they are mounted vertically to the frame to reduce the absolute size of the system for an easier handling and ROV-based deployment and to avoid damage of outstanding components.
 The cameras are thus looking upwards through two mirrors at $45^\circ$ angle to create two virtual horizontal views into the bubble rise corridor.

\subsubsection{Adjustment and Calibration}
\label{sec:stereo_calib}
\begin{figure}[!ht]
	\begin{center}
	\includegraphics[height=109pt]{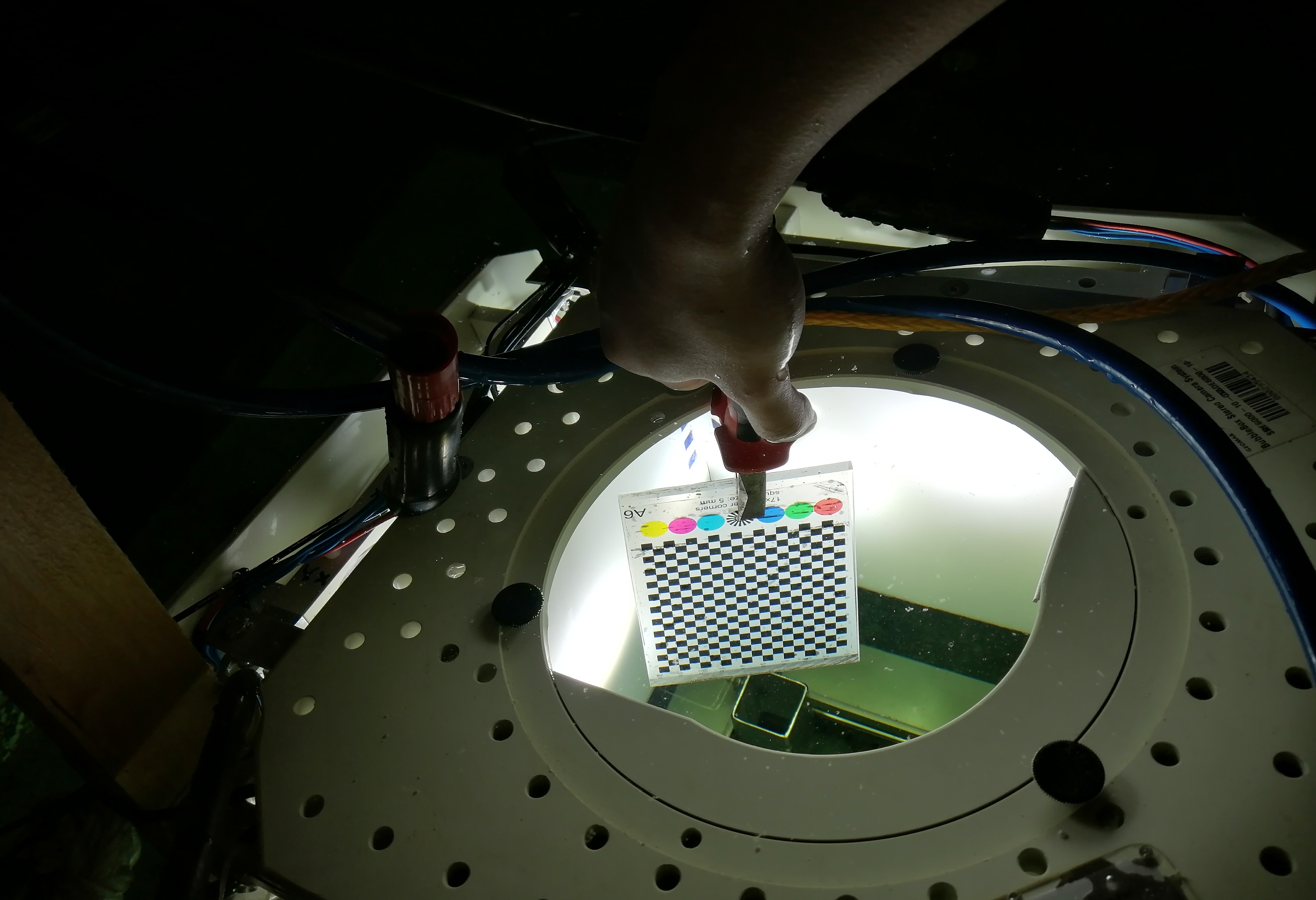}
	\includegraphics[height=109pt]{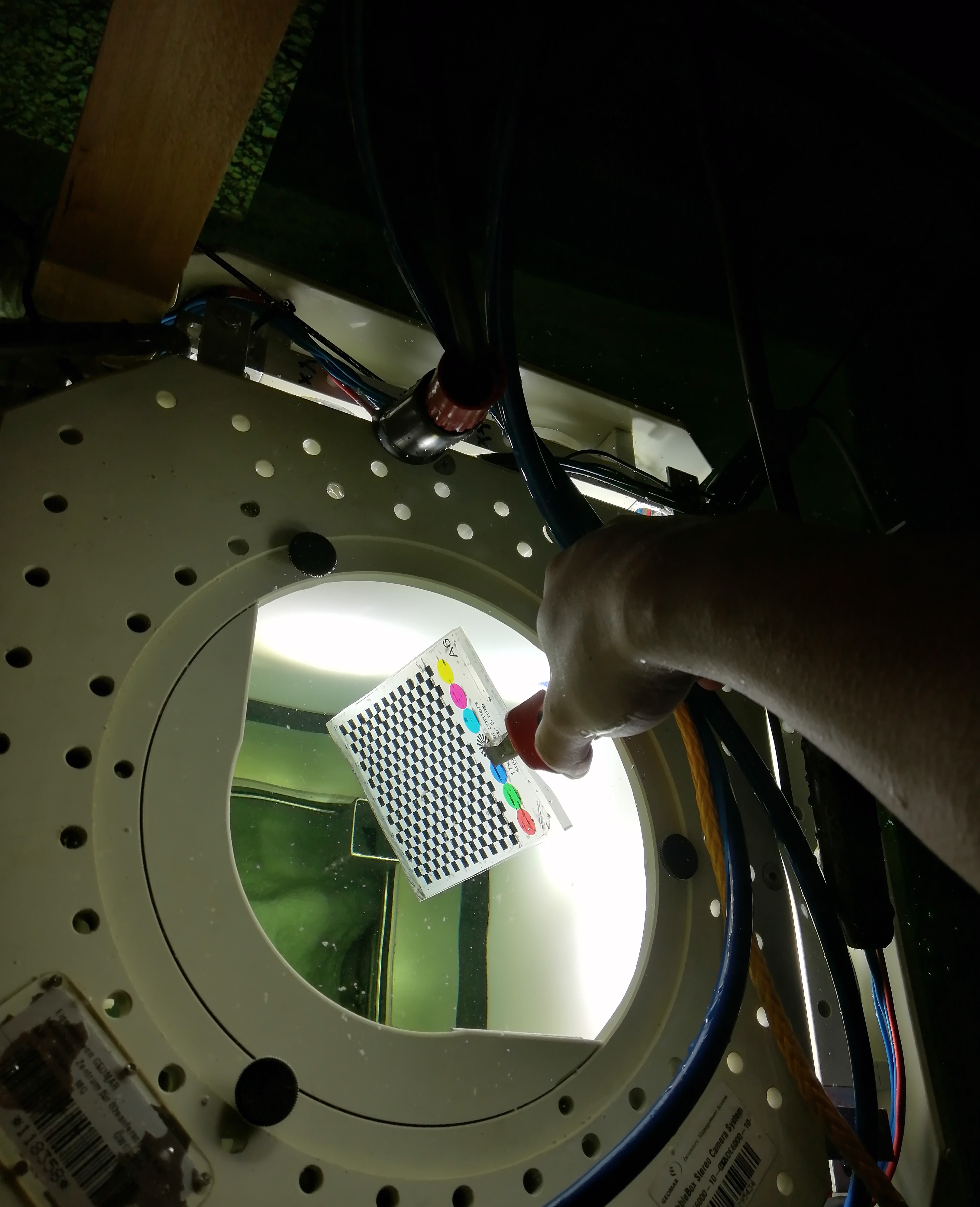}
	\includegraphics[height=109pt]{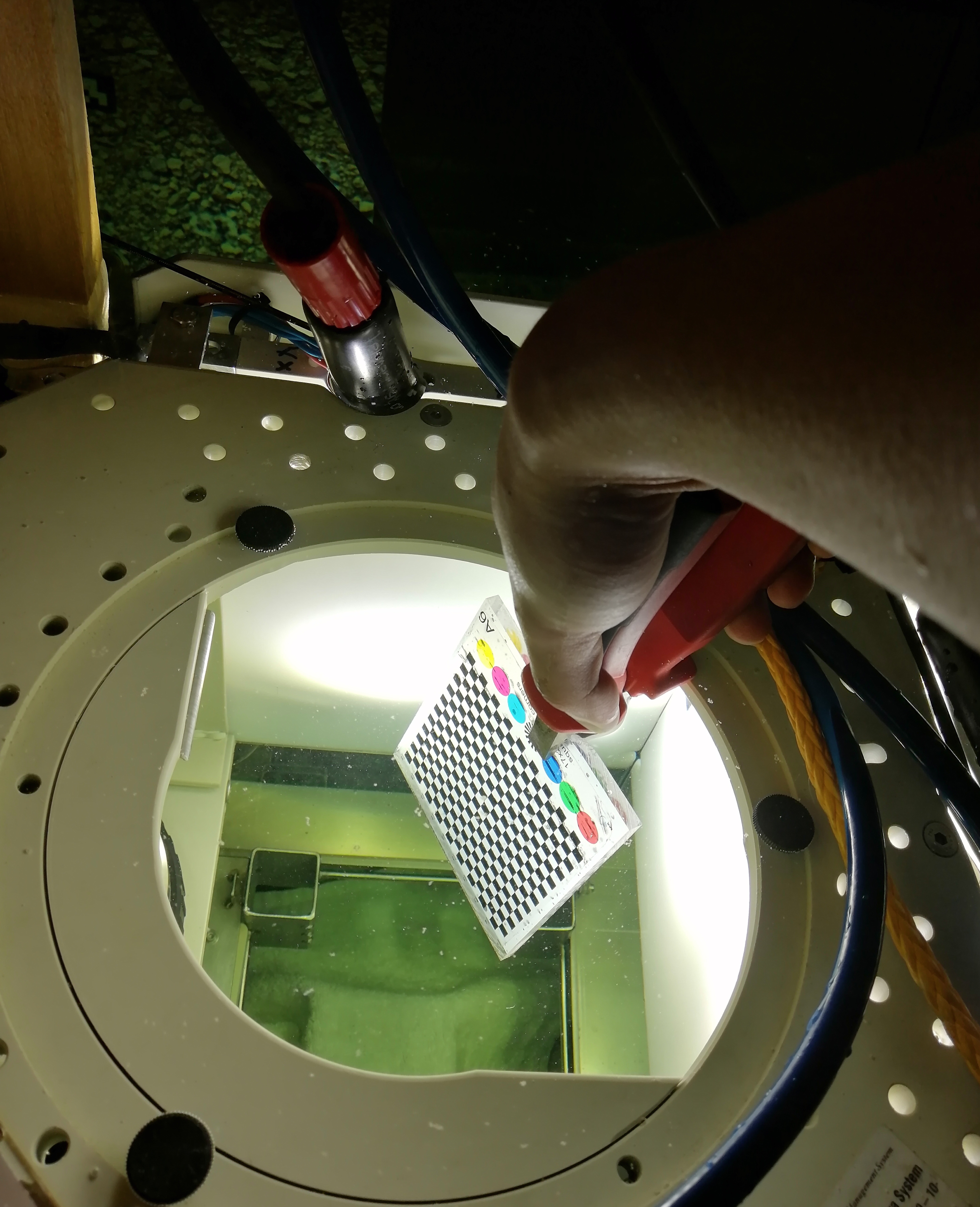}
	\includegraphics[width=1.0\textwidth]{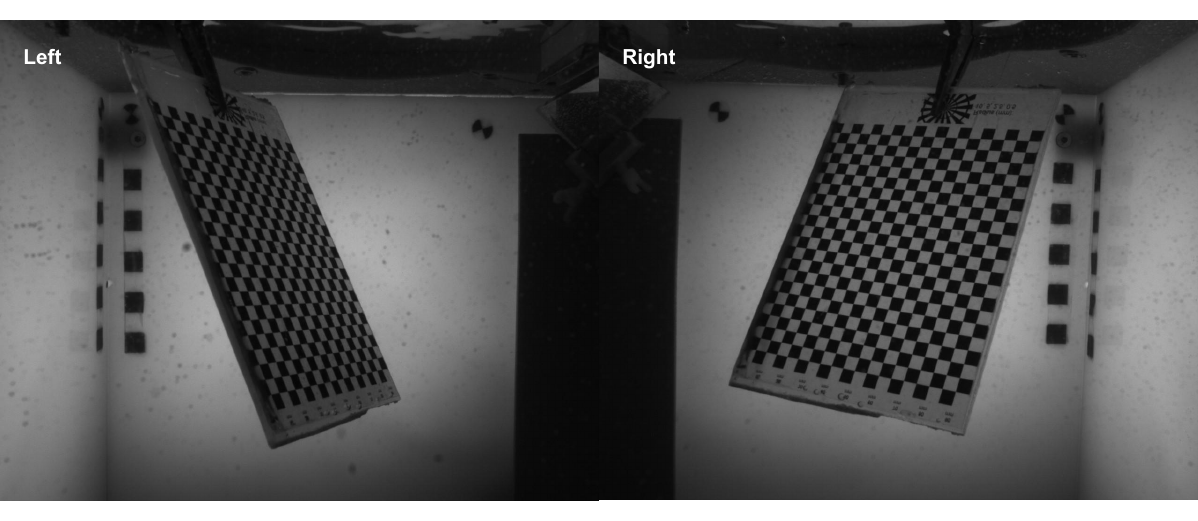}
	\end{center}
\caption{
	Stereo calibration of the instrument. Top: A calibration target is photographed by both two cameras multiple times underwater.
	Hence, the relative position and orientation as well as the camera intrinsics can be obtained.
	Bottom: sample calibration photos.
}
\label{fig:stereo_calib}
\end{figure}
According to the method of \cite{she2019adjustment}, the camera is first centered with the dome port.
To achieve this, the dome port camera is assembled using an adjustable mount and submerged half-way underwater in a pool, such that the camera looks parallel to the water surface.
Next, a chessboard is presented in front of the camera and half-way submerged underwater.
Then the camera is centered, such that no refraction will occur and the underwater part and the above water part of the image will be consistent.
Afterwards, the camera position is fixed and a set of control images of an underwater chessboard is taken in order to calibrate potentially remaining decentering parameters using the method of \cite{she2021refractive}.
After adjustment it is verified that refraction effects are insignificant for observations in the rise corridor\cite{she2021refractive,s2015}, such that a perspective camera model can be used, and the potentially remaining refraction effects can be well absorbed into lens distortion parameters since we are observing objects at a relatively fixed distance. 
Note that this adjustment procedure should be conducted every time the dome port camera is re-assembled.

After centering the cameras with the dome ports, the closed camera housings are rigidly mounted into the BBox and stereo camera calibration is performed.
Due to the bright field setting, stereo calibration is carried out underwater using a transparent calibration target within the rise corridor.
As shown in Fig. \ref{fig:stereo_calib}, both cameras observe the same target multiple times, and the camera intrinsics $\mathcal{K}_1, \mathcal{K}_2$\footnote{Classically, the calibration matrix $\mq K$ represents the perspective camera intrinsics such as focal lengths and principal points. For ease of notation we use the symbol $\mathcal{K}$ to represent all intrinsics, including lens distortion parameters.} are calibrated using the traditional method of Zhang \cite{zhang2000flexible} (see also OceanBestPractises recommendation on underwater camera calibration\cite{oceanbestpractises2016}).
In this contribution, we calibrate the focal lengths and principal points of the camera, and lens distortion parameters such as 2 radial distortion coefficients $k_1, k_2$ and 2 tangential distortion coefficients $p_1, p_2$.
Then, we define the left camera to be the origin of the world coordinate system as $\mq T_1 = [\m I_{3 \times 3} \mid \d 0 ]$. The right camera has the pose of $\mq T_2 = [\m R  \mid \d t]$.
We can then determine the pose of the right camera and do final refinement on both camera intrinsics by projecting the 3D target points onto the stereo images and minimizing the residuals between the projections and the identified corresponding points, as common for target-based calibration:
\begin{equation}
E = \sum_{i}^{n}\sum_{j}^{m}( \Vert  
\d \pi(\q X_{i}, \mq T_1, \mathcal{K}_1) - \q x_{i,j}^{1}
\Vert^{2} + \Vert \d \pi(\q X_{i} , \mq T_2, \mathcal{K}_2) - \q x_{i,j}^{2} \Vert^{2} )
\end{equation}
Here, $\q x_{i,j}^{1}$ and $\q x_{i,j}^{2}$ indicate the $i^{th}$ point on the target photographed by the $j^{th}$ image, and the superscript $^1$ and $^2$ indicate the left camera and the right camera.
$\d \pi()$ represents the perspective projection function.

\subsection{Bubble Stream Characterization Method}
\label{sec:bubble_stream_charac}
\begin{figure}[!ht]
	\begin{center}
		\includegraphics[width=1.0\textwidth]{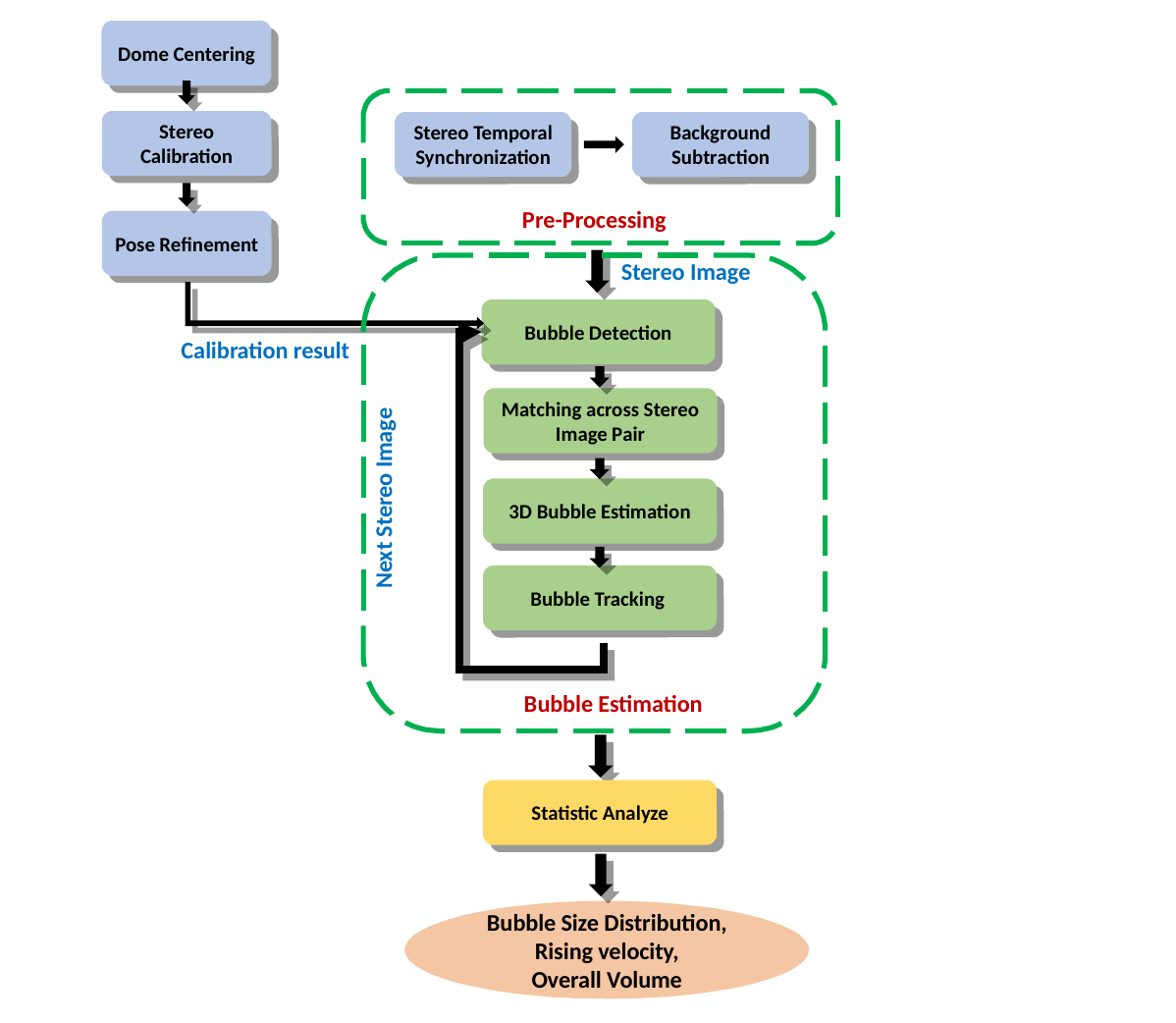}
	\end{center}
\caption{An overview of the automatic bubble stream characterization pipeline.}
\label{fig:bubble_overview}
\end{figure}

During in-situ operation, image data is recorded by the instrument.
Image processing is done afterwards using different modules written in C++ and CUDA.
Low-level image processing is based on OpenCV\footnote{https://opencv.org}, optimization uses the Ceres Solver\footnote{http://ceres-solver.org} but intermediate processing and multiple view geometry reasoning uses our in-house computer vision software libraries or consists of new developments (e.g. sliding window median filering on GPU).
The entire software can also run inside a docker-container\footnote{https://www.docker.com} on a remote server with GPU capabilities. 
 In this case access works through a Jupyter Notebook\footnote{https://jupyter.org} interface. Automated, and efficient, processing is important, since more than half a million images are recorded per hour.
After processing, the software is able to generate a report on the important bubble stream characteristics such as the overall volume of gas released, bubble size distribution and rise velocity.
In this section, we will provide details on the bubble stream characterization approach, and Fig. \ref{fig:bubble_overview} illustrates an overview of the processing pipeline.

\subsubsection{Temporal Synchronization of Stereo Data}
The two cameras are operated by independent computers and record time-stamped data (images).
The time-stamps of the computers agree up to one, or for long-term deployments at most a few, seconds.
Since the goal is to record with 80Hz to 100Hz, we would need a clock agreement of less than 5ms in order to unambiguously associate matching frames in the recorded sequences.
Therefore, every 5000 images, the microcontroller will generate a very short flash time (10 microsecond rather than 1000 microseconds), which leads to a black image. Those black images are now used as identifiers for synchronization of the two photo sequences. To achieve that, the software first iterates through both photo sequences and extracts the timestamp of each image and also detects the black images. This way the damage of potential frame drops in one camera is limited and the moment it happened can be detected easier.
Next, the average time offset between the two computers can be calculated by the timestamps of the aligned black images.
Afterwards, for each image in one of the sequences, we compute the expected timestamp when the corresponding image is captured, and then search the corresponding image in the other sequence by finding the minimum time difference considering the derived time offset from the black image comparison.
Finally, the two photo sequences are aligned and a sequence of stereo image pairs is created.
Particularly, when looking for black images, we sample the main-diagonal and counter-diagonal pixels of the image and check their intensities.
If all pixel intensities are smaller than an empirical threshold (for 8 bit images, currently the value 8 is used), the image is considered a synchronization frame.

\subsubsection{Background Learning and Removal}
\label{sec:BackgroundRemoval}

\begin{figure}[!ht]
	\begin{center}
		\includegraphics[width=1.0\textwidth]{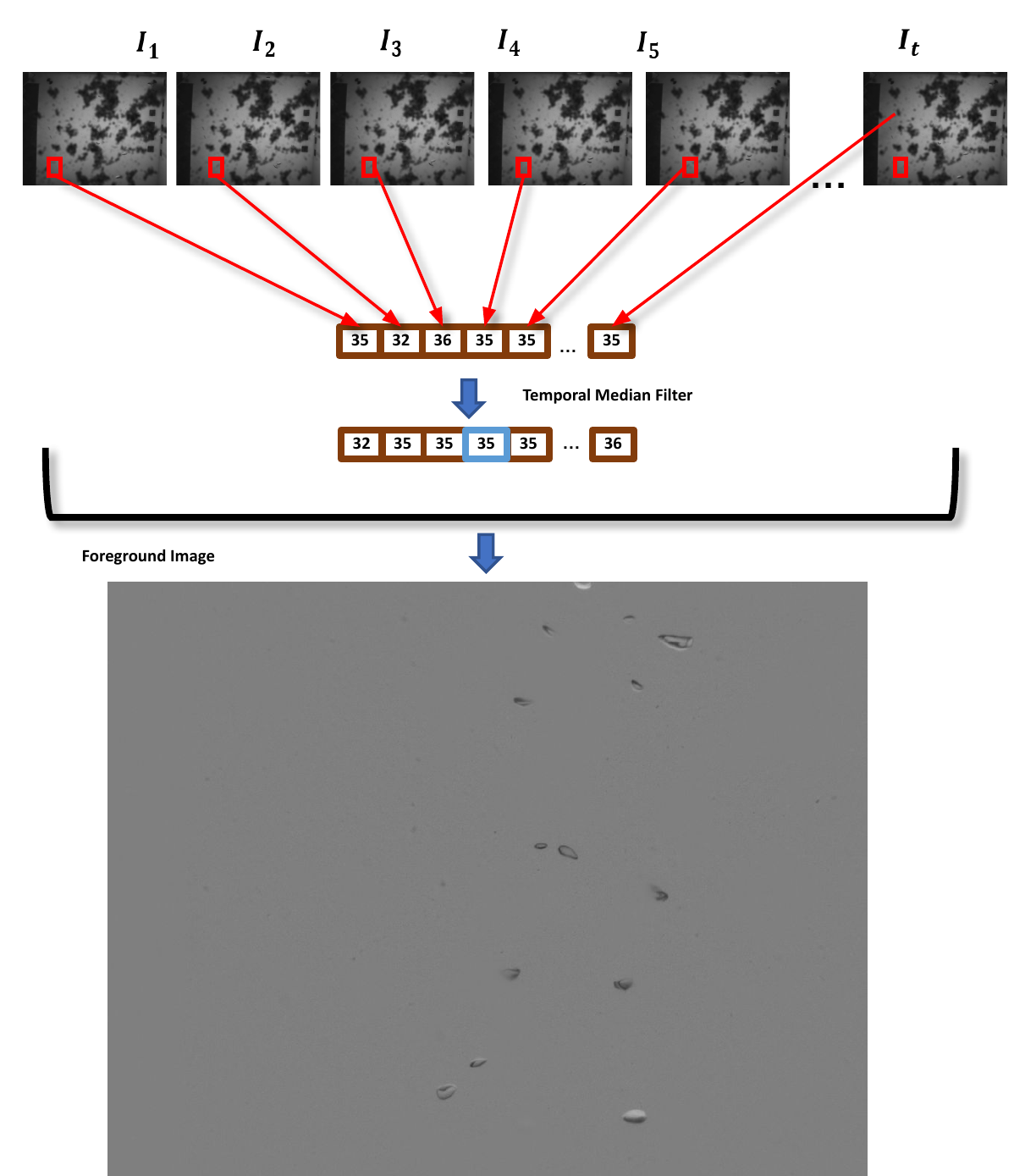}
	\end{center}
\caption{Principal of the temporal median filtering for background learning.
	If the background structures are stable for a certain time interval, they can be learned and removed by estimating a \emph{background image}.
	For each pixel in the background image, the intensity value is determined by taking the median value along the temporal time series.
}
\label{fig:background_subtract}
\end{figure}

The original images usually contain complex background structures, for instance, sediment stuck on the dome ports, bubbles trapped and the structures and markers on the frame of the instrument itself, which makes bubble detection complicated in the raw images.
But the background information can be learned and therefore removed if the background objects stay static over a certain time interval.
Assuming that we have only sparse bubble observations, at each pixel position in the image we will see the background in the majority of the images. 
Since the Median is a robust estimator with a breakdown point of 50\%, it can be employed to robustly estimate the background per pixel (from many subsequent images at the same image position).
Consequently, a temporal median filter is leveraged to compute a background image from a series of images even with bubbles(see Fig. \ref{fig:background_subtract}). 
Since the background can change over time, we apply the Median background estimation using a sliding-window approach. We then subtract the 'learned' background image from each raw bubble image to obtain only the moving objects, i.e. the bubbles.
In this step the images can also be undistorted to remove the lens distortion, in case lens distortion is present.
These processing steps were implemented efficiently in CUDA using a streaming architecture.

\subsubsection{Bubble Detection}
\label{sec:bubble_detection}
\begin{figure}[!h]
	\begin{center}
		\subfloat[Foreground image]{
			\includegraphics[width=0.24\textwidth]{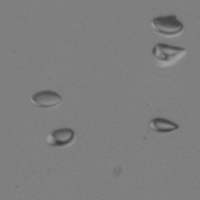}
		}
		\subfloat[Canny edges]{
			\includegraphics[width=0.24\textwidth]{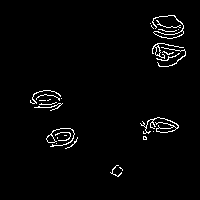}
		}
		\subfloat[Ellipse fitting]{
			\includegraphics[width=0.24\textwidth]{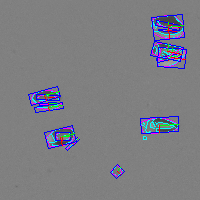}
		}
		\subfloat[Contours merging]{
			\includegraphics[width=0.24\textwidth]{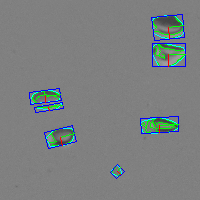}
		}\newline
		\subfloat[Bubble cleaning]{
			\includegraphics[width=0.45\textwidth]{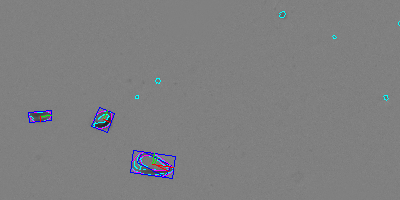}
			\label{fig:bubble_detection_pipeline-cleaning}
		}
		\subfloat[Final result]{
			\includegraphics[width=0.45\textwidth]{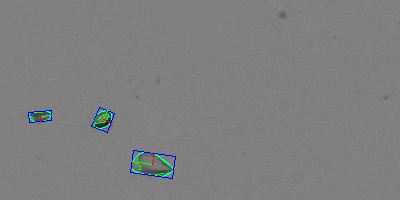}
		}
	\end{center}
	\caption{The workflow of bubble detection using image processing techniques.}
	\label{fig:bubble_detection_pipeline}
\end{figure}
Bubble detection aims at finding positions and 2D contours of the bubbles in an image, more specifically, the bubble contours. 
Due to the back illumination, the bubbles appear as dark rims with a brighter area inside the contour in the foreground images.
The image processing workflow used here is similar to \cite{jordt2015bubble, thomanek2010automated}. First, the Canny edge detector is performed to find those pixels that belong to the contour and group them into edgels. Then, the convex hull is determined for each bubble and an ellipse is fitted to the convex hull afterwards.
The workflow of bubble detection is shown in Fig. \ref{fig:bubble_detection_pipeline}.
Since there are always single noisy pixels or entire dirt particles inside the water, a threshold for a minimum size of bubbles has to be used. Since a 1mm-diameter spherical bubble corresponds to a circle of about 31 pixels circumference, the default setting is to reject contours with less than 30 pixels length as moving particles; this reduces mis-detection (see Fig. \ref{fig:bubble_detection_pipeline-cleaning}). The threshold can be adapted, e.g. when the goal is to measure very small bubbles in very clear water.
\begin{figure}
	\begin{center}
		\def\svgwidth{0.9\textwidth}
		{
			%% Creator: Inkscape 1.1.1 (1:1.1+202109281949+c3084ef5ed), www.inkscape.org
			%% Accompanies image file 'epigeometry.pdf' (pdf, eps, ps)
			\begingroup%
			\makeatletter%
			\providecommand\color[2][]{%
				\errmessage{(Inkscape) Color is used for the text in Inkscape, but the package 'color.sty' is not loaded}%
				\renewcommand\color[2][]{}%
			}%
			\providecommand\transparent[1]{%
				\errmessage{(Inkscape) Transparency is used (non-zero) for the text in Inkscape, but the package 'transparent.sty' is not loaded}%
				\renewcommand\transparent[1]{}%
			}%
			\providecommand\rotatebox[2]{#2}%
			\newcommand*\fsize{\dimexpr\f@size pt\relax}%
			\newcommand*\lineheight[1]{\fontsize{\fsize}{#1\fsize}\selectfont}%
			\ifx\svgwidth\undefined%
			\setlength{\unitlength}{345bp}%
			\ifx\svgscale\undefined%
			\relax%
			\else%
			\setlength{\unitlength}{\unitlength * \real{\svgscale}}%
			\fi%
			\else%
			\setlength{\unitlength}{\svgwidth}%
			\fi%
			\global\let\svgwidth\undefined%
			\global\let\svgscale\undefined%
			\makeatother%
			\begin{picture}(1,0.57971012)%
			\lineheight{1}%
			\setlength\tabcolsep{0pt}%
			\put(0,0){\includegraphics[width=\unitlength,page=1]{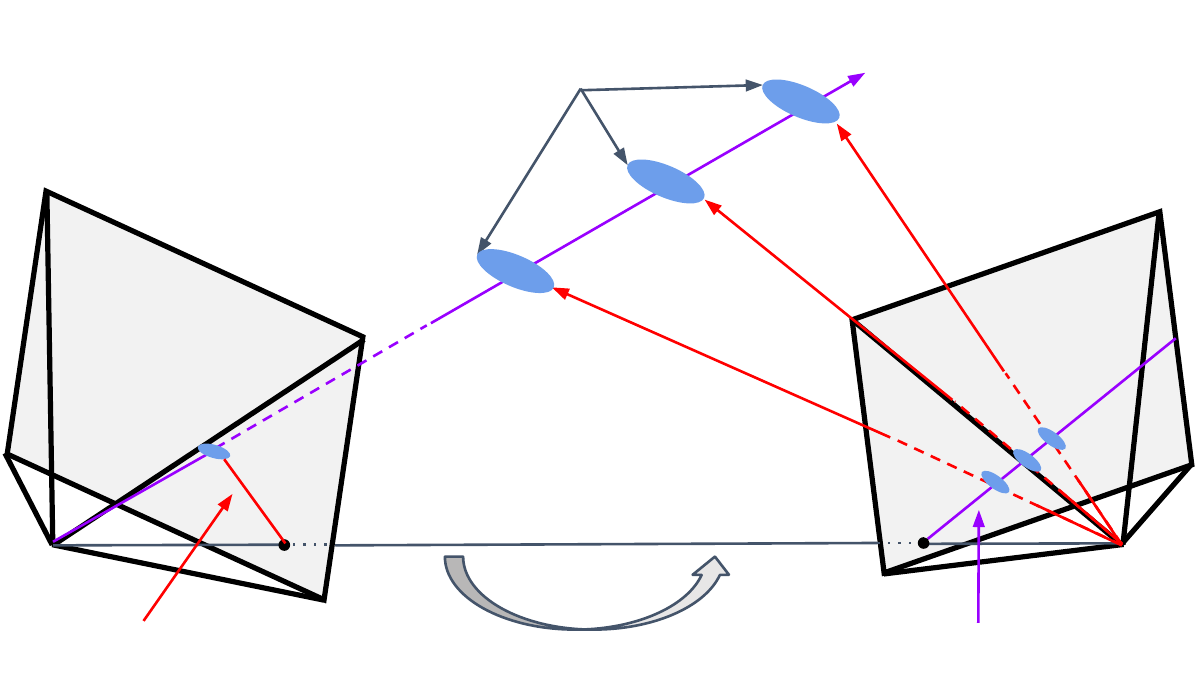}}%
			\put(0.75218632,0.14997996){\color[rgb]{0,0,0}\makebox(0,0)[lt]{\lineheight{1.25}\smash{\begin{tabular}[t]{l}$\q e_2$\end{tabular}}}}%
			\put(0.23921124,0.14854471){\color[rgb]{0,0,0}\makebox(0,0)[lt]{\lineheight{1.25}\smash{\begin{tabular}[t]{l}$\q e_1$\end{tabular}}}}%
			\put(0.10643515,0.04025683){\color[rgb]{0,0,0}\makebox(0,0)[lt]{\lineheight{1.25}\smash{\begin{tabular}[t]{l}Epipolar line $\q l_1$ \end{tabular}}}}%
			\put(0.02962174,0.08871616){\color[rgb]{0,0,0}\makebox(0,0)[lt]{\lineheight{1.25}\smash{\begin{tabular}[t]{l}$\d C_1$\end{tabular}}}}%
			\put(0.92968956,0.10246075){\color[rgb]{0,0,0}\makebox(0,0)[lt]{\lineheight{1.25}\smash{\begin{tabular}[t]{l}$\d C_2$\end{tabular}}}}%
			\put(0.74664824,0.03774802){\color[rgb]{0,0,0}\makebox(0,0)[lt]{\lineheight{1.25}\smash{\begin{tabular}[t]{l}Epipolar line $\q l_2$\end{tabular}}}}%
			\put(0.47083302,0.02888326){\color[rgb]{0,0,0}\makebox(0,0)[lt]{\lineheight{1.25}\smash{\begin{tabular}[t]{l}$\m R, \d t$\end{tabular}}}}%
			\put(0.44510823,0.52015774){\color[rgb]{0,0,0}\makebox(0,0)[lt]{\lineheight{1.25}\smash{\begin{tabular}[t]{l}Bubble $\mathcal{B}$ ?\end{tabular}}}}%
			\put(0.1464242,0.22308197){\color[rgb]{0,0,0}\makebox(0,0)[lt]{\lineheight{1.25}\smash{\begin{tabular}[t]{l}$\mathcal{B}_1$\end{tabular}}}}%
			\put(0.87396897,0.24654767){\color[rgb]{0,0,0}\makebox(0,0)[lt]{\lineheight{1.25}\smash{\begin{tabular}[t]{l}$\mathcal{B}_2$\end{tabular}}}}%
			\end{picture}%
			\endgroup%
			
		}
	\end{center}
	\caption{Epipolar geometry constraint for stereo bubble matching. 
		The epipolar lines $\q l_1$ and $\q l_2$ are the projections of the light rays sensed by the pixel $\mathcal{B}_2$ and $\mathcal{B}_1$, respectively. $\d C_1$ and $\d C_2$ are the camera centers. The baseline between two camera centers and the light rays of a corresponding bubble span an epipolar plane. The epipolar constraint reduces the search space of the corresponding bubble.}
	\label{fig:epigeometry}
\end{figure}
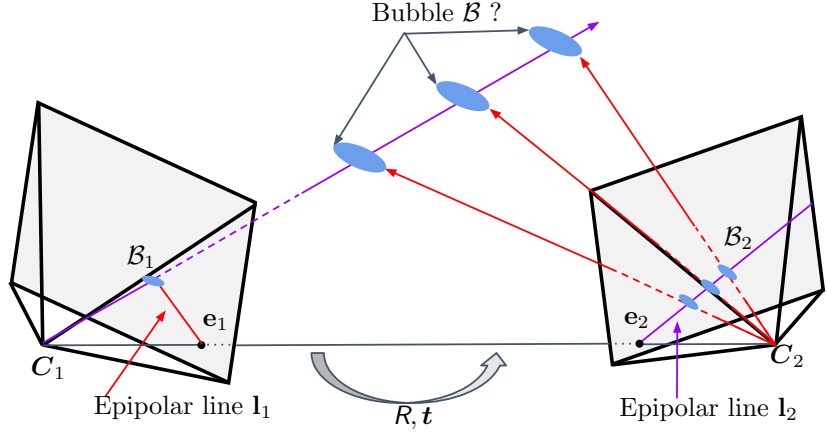

\subsubsection{Epipolar Geometry and Stereo Matching}
\label{sec:epigeometry}
To estimate the bubble volumes, it is required to find the corresponding bubble outlines across the stereo image pairs.
 Traditional feature-based \cite{Lowe-2004-Features} or pixel-wise \cite{hirschmuller2007stereo} matching approaches find correspondences by computing similarity of local image patches. 
 Since bubbles lack texture or other specific appearance information, only geometric constraints can be used for matching, such as epipolar geometry.
As shown in Fig. \ref{fig:epigeometry}, a bubble is observed by two cameras, and the pose of the second camera with respect to the first camera is related by a rotation and translation $\m R, \d t$, which is obtained from the stereo calibration.
 The second bubble's projection is therefore constrained by epipolar geometry. The matching we use is the same as described in \cite{jordt2015bubble}, searching for candidates in a corridor around the epipolar line, and then solving the matching problem of all bubbles in an image pair at once using a bipartite graph \cite{kuhn1955hungarian}.
After finding the bubble correspondences in the stereo image pair, we can estimate the bubble shape.
%Empirically, air, methane and CO$_2$ bubbles in sub-millimeter range are spherical, and become ellipsoidal in the several millimeters range, in the centimeters range they deform to spherical caps or become unstable.
\subsubsection{Ellipsoid Initialization}
We model all bubbles in 3D as ellipsoidal, and initialize a 3D ellipsoid from the 2D ellipse parameters in the two images, before we optimize the 3D ellipsoid position and axes to fit both projections.

\begin{figure}[!h]
	\begin{center}
		\def\svgwidth{0.98\textwidth}
		{
			%% Creator: Inkscape 1.1.1 (1:1.1+202109281949+c3084ef5ed), www.inkscape.org
			%% Accompanies image file 'bubble3D.pdf' (pdf, eps, ps)
			\begingroup%
			\makeatletter%
			\providecommand\color[2][]{%
				\errmessage{(Inkscape) Color is used for the text in Inkscape, but the package 'color.sty' is not loaded}%
				\renewcommand\color[2][]{}%
			}%
			\providecommand\transparent[1]{%
				\errmessage{(Inkscape) Transparency is used (non-zero) for the text in Inkscape, but the package 'transparent.sty' is not loaded}%
				\renewcommand\transparent[1]{}%
			}%
			\providecommand\rotatebox[2]{#2}%
			\newcommand*\fsize{\dimexpr\f@size pt\relax}%
			\newcommand*\lineheight[1]{\fontsize{\fsize}{#1\fsize}\selectfont}%
			\ifx\svgwidth\undefined%
			\setlength{\unitlength}{345bp}%
			\ifx\svgscale\undefined%
			\relax%
			\else%
			\setlength{\unitlength}{\unitlength * \real{\svgscale}}%
			\fi%
			\else%
			\setlength{\unitlength}{\svgwidth}%
			\fi%
			\global\let\svgwidth\undefined%
			\global\let\svgscale\undefined%
			\makeatother%
			\begin{picture}(1,0.56231882)%
			\lineheight{1}%
			\setlength\tabcolsep{0pt}%
			\put(0,0){\includegraphics[width=\unitlength,page=1]{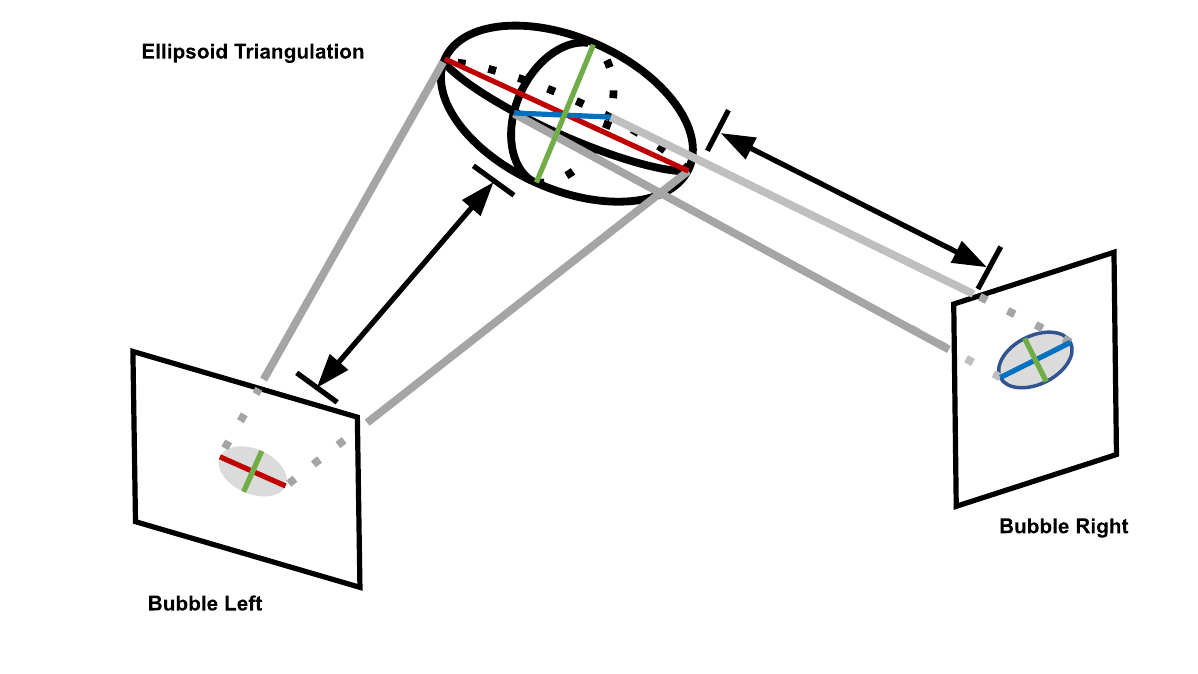}}%
			\put(0.3213152,0.34401756){\color[rgb]{0,0,0}\makebox(0,0)[lt]{\lineheight{1.25}\smash{\begin{tabular}[t]{l}$d_1$\end{tabular}}}}%
			\put(0.72020688,0.40872571){\color[rgb]{0,0,0}\makebox(0,0)[lt]{\lineheight{1.25}\smash{\begin{tabular}[t]{l}$d_2$\end{tabular}}}}%
			\put(0.57870198,0.39606016){\color[rgb]{0,0,0}\makebox(0,0)[lt]{\lineheight{1.25}\smash{\begin{tabular}[t]{l}$A_2$\end{tabular}}}}%
			\put(0.49653964,0.53803072){\color[rgb]{0,0,0}\makebox(0,0)[lt]{\lineheight{1.25}\smash{\begin{tabular}[t]{l}$C_1$\end{tabular}}}}%
			\put(0.43119714,0.36562601){\color[rgb]{0,0,0}\makebox(0,0)[lt]{\lineheight{1.25}\smash{\begin{tabular}[t]{l}$C_2$\end{tabular}}}}%
			\put(0.38569859,0.45841842){\color[rgb]{0,0,0}\makebox(0,0)[lt]{\lineheight{1.25}\smash{\begin{tabular}[t]{l}$B_1$\end{tabular}}}}%
			\put(0.51893785,0.46522981){\color[rgb]{0,0,0}\makebox(0,0)[lt]{\lineheight{1.25}\smash{\begin{tabular}[t]{l}$B_2$\end{tabular}}}}%
			\put(0.15591123,0.1815303){\color[rgb]{0,0,0}\makebox(0,0)[lt]{\lineheight{1.25}\smash{\begin{tabular}[t]{l}$a_1$\end{tabular}}}}%
			\put(0.32856073,0.51685643){\color[rgb]{0,0,0}\makebox(0,0)[lt]{\lineheight{1.25}\smash{\begin{tabular}[t]{l}$A_1$\end{tabular}}}}%
			\put(0.25339384,0.14865099){\color[rgb]{0,0,0}\makebox(0,0)[lt]{\lineheight{1.25}\smash{\begin{tabular}[t]{l}$a_2$\end{tabular}}}}%
			\put(0.19573033,0.13052748){\color[rgb]{0,0,0}\makebox(0,0)[lt]{\lineheight{1.25}\smash{\begin{tabular}[t]{l}$c_2$\end{tabular}}}}%
			\put(0.2156049,0.18820715){\color[rgb]{0,0,0}\makebox(0,0)[lt]{\lineheight{1.25}\smash{\begin{tabular}[t]{l}$c_1$\end{tabular}}}}%
			\put(0.84386437,0.29356652){\color[rgb]{0,0,0}\makebox(0,0)[lt]{\lineheight{1.25}\smash{\begin{tabular}[t]{l}$c_1$\end{tabular}}}}%
			\put(0.87077484,0.21898473){\color[rgb]{0,0,0}\makebox(0,0)[lt]{\lineheight{1.25}\smash{\begin{tabular}[t]{l}$c_2$\end{tabular}}}}%
			\put(0.80874735,0.22929254){\color[rgb]{0,0,0}\makebox(0,0)[lt]{\lineheight{1.25}\smash{\begin{tabular}[t]{l}$b_1$\end{tabular}}}}%
			\put(0.90066542,0.27505231){\color[rgb]{0,0,0}\makebox(0,0)[lt]{\lineheight{1.25}\smash{\begin{tabular}[t]{l}$b_2$\end{tabular}}}}%
			\end{picture}%
			\endgroup%
			
		}
	\end{center}
	\caption{The principal of 3D ellipsoid triangulation from the stereo image pair.}
	\label{fig:ellipsoid_triangulation}
\end{figure}
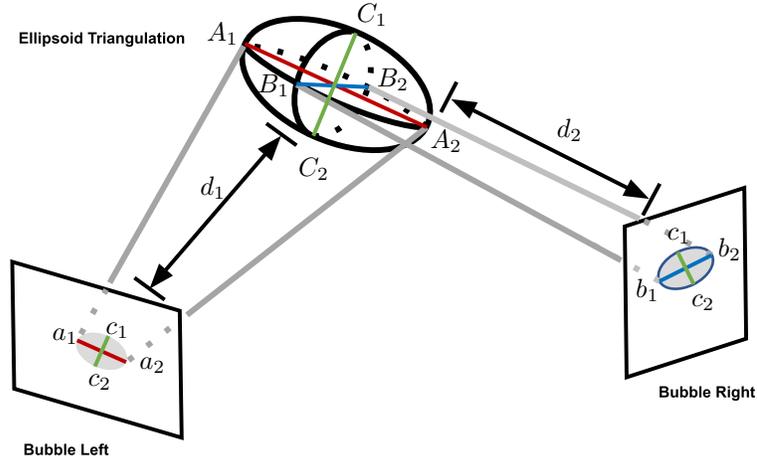

As can be seen in Fig. \ref{fig:ellipsoid_triangulation}, the center of the ellipsoid is triangulated from the center of the ellipses, therefore, the distances of the bubble center to the cameras are also computed as $d_1$ and $d_2$.
We specify a set of feature points $A_1,A_2,B_1,B_2,C_1,C_2$ as the endpoint of the ellipsoid axes, among which $A_1, A_2$ are back-projected from $a_1, a_2$ in the left image with a distance of $d_1$.
Similarly, $B_1, B_2$ are back-projected from $b_1, b_2$ in the right image with a distance of $d_2$.
The vector $\overrightarrow{A_1A_2}$ and $\overrightarrow{B_1B_2}$ form a plane, and the third vector $\overrightarrow{C_1C_2}$ is constructed as the cross product of the other two vectors, such that $\overrightarrow{C_1C_2}$ is orthogonal to the plane.
Since $\overrightarrow{A_1A_2}$ and $\overrightarrow{B_1B_2}$ are not orthogonal, we setup a $3\times3$ matrix $\m R$ with three columns being these vectors, and map to a rotation matrix where the three columns are orthonormal.
Then, we obtain the axes of the ellipsoid.
Finally, we recompute the lengths of the axes by mapping the pre-computed feature points to the axis vectors,
and the volume of the bubble can be obtained as:
\begin{equation}
	V = \frac{4}{3}\pi abc`
\end{equation}
where $a,b,c$ are the lengths of the ellipsoid axes.

However, the center of the 2D ellipses in the left and right image is actually not a corresponding point, and their back-projected rays do not intersect in 3D.
In addition, the same bubble distance is used to initialize the endpoints of the ellipsoid axes.
Consequently, the triangulated ellipsoid is only an initial solution which should be further refined using 2D image observations, which in this case, are the bubble contour constraints in the stereo images instead of point correspondences, which will be described in the next subsection.

\subsubsection{Bubble Adjustment without Point Correspondences}
\label{sec:ba_bubble}
We first represent the ellipsoid surface as a quadric, which is a $4\times4$ symmetric matrix $\mq Q$ in the 3D projective space $\mathbb{P}^3$ \cite{Hartley2004_MultipleView}.
The quadric can be initialized with the triangulated ellipsoid parameters as following:
\begin{equation}
	\mq Q = \mq H^{-\mathrm{T}} \mq Q_u \mq H^{-1}
\end{equation}
where $\mq Q_u = diag\{1, 1, 1, -1\}$ denotes the unit sphere.
$\mq H$ is the point transformation matrix which is composed by the orientation, translation, and the lengths of the ellipsoid axes:
\begin{equation}
	\mq H = \begin{bmatrix}
	\m D_e & \d t_e \\
	\d 0\trans & 1
	\end{bmatrix}
\end{equation}
where $\d t_e$ is the translation of the ellipsoid center, and $\m D_e$ is: 
\begin{equation}
	\m D_e = \begin{bmatrix}
	a & 0 & 0\\
	0 & b & 0\\
	0 & 0 & c
	\end{bmatrix} \cdot
	\m R_e
\end{equation}
with $\m R_e$ being the orientation of the ellipsoid.
Since we have calibrated the stereo camera system, we have obtained the camera intrinsics, as well as the camera poses, which are $\mq K_1, \mq K_2, \mq T_1, \mq T_2$.
Therefore, we can construct the projection matrices of the stereo cameras as $\mq P_1 = \mq K_1 \mq T_1$ and $\mq P_2 = \mq K_2 \mq T_2$.%, and augment them to $4\times4$ matrices. 
Now, given the projection matrix $\mq P$ of a camera, the quadric can be projected onto the image as:
\begin{equation}
	\mq C^\star = \mq P \mq Q^\star \mq P\trans 
\end{equation}
where $\mq C^\star, \mq Q^\star$ is the dual conic of $\mq C$, and dual quadric of $\mq Q$ respectively, and they can be obtained through:
\begin{equation}
	\mq Q^\star = \mq Q^{-1}\;\;\;\;\;\;\;\;\;\;\;\;\;\;\;\;\;\;\;\;\;\;	\mq C^\star = \mq C^{-1}
\end{equation}

Therefore, we can project the triangulated ellipsoid to the left image and the right image and obtain their 2D conics $\mq C_1$ and $\mq C_2$.
To optimize the quadric, we minimize the difference between the projected conics and the detected conics.
Since we have detected the bubble and fit an ellipse around the contour in Sect. \ref{sec:bubble_detection},
we sample points on the ellipse uniformly and define the cost function as the Mahalanobis distance of the sampled points $\q x$ (lens distortion was removed in the background removal step) to the conic:
\begin{equation}
	\q x \trans \mq C \q x
\end{equation}
Therefore, the optimal quadric $\mq Q$ can be solved via minimizing the following:
\begin{equation}
E = \sum_{i} \Vert \q x_{i,1}\trans \mq C_1 \q x_{i,1} \Vert^2 + \Vert \q x_{i,2}\trans \mq C_2 \q x_{i,2}\Vert^2
\label{eq:quadricenergy}
\end{equation}
where $\q x_{i,1}$ and $\q x_{i,2}$ are the sampled points on the detected ellipse in the left image and the right image.
After optimizing the quadric, we retrieve the ellipsoid parameters from the quadric representation.
\subsubsection{Bubble Tracking}
\begin{figure}[!h]
	\includegraphics[width=0.9\textwidth]{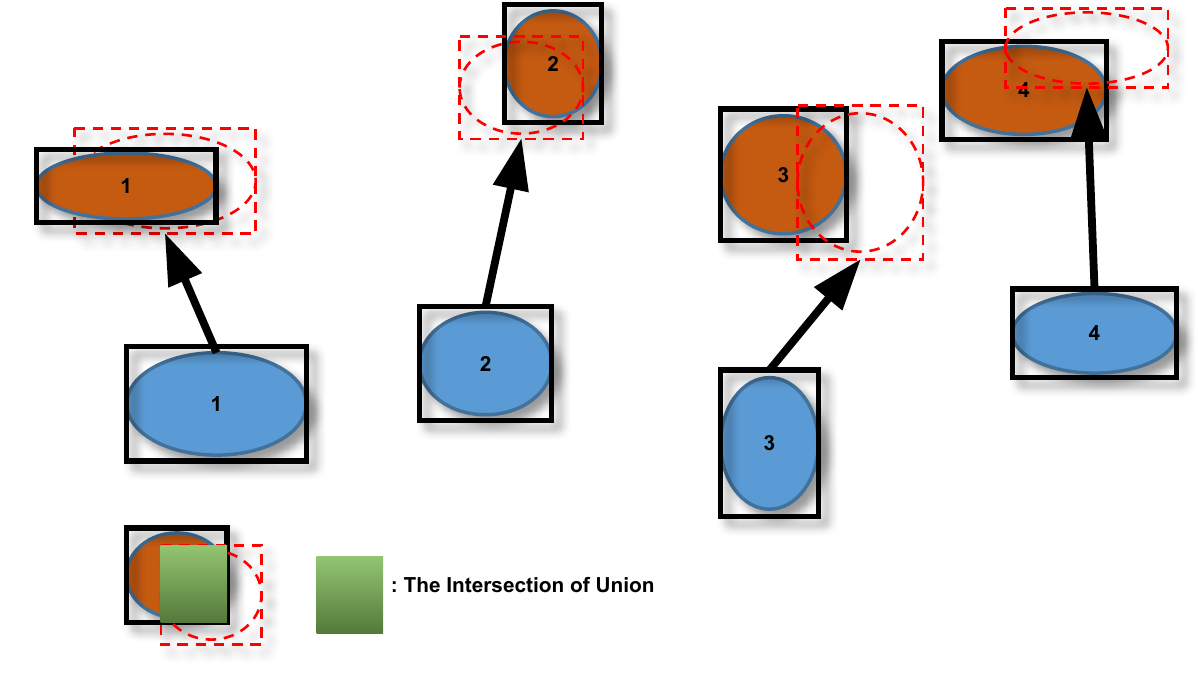}
\caption{The principal of the data association step in the bubble tracking.
	The blue ellipses denote the bubbles in the current frame, the brown
	ellipses represent the identified bubbles in the next frame. The IoU is
	computed between the predicted and detected bounding box.
}
\label{fig:IoU_tracking}
\end{figure}
To measure the rise speed of each bubble, and also to avoid counting bubbles multiple times, the bubbles need to be tracked over the image sequence.
We utilize the Tracking-by-Detection \cite{leal2017tracking} framework to address this issue where the interesting target is identified in each frame and associated with its previous trajectory. 
Therefore, the tracking problem is essentially cast as a matching problem which is similar to Sect. \ref{sec:epigeometry}.
The bubbles always rise upwards with a small oscillation in the sideward direction, which can be used as a constraint to reduce the search space in the data association step.
In this contribution, we employ the SORT (Simple Online and Realtime Tracking) tracker \cite{bewley2016simple}.
The bubble position in the next frame is first predicted by a Kalman Filter \cite{kalman1960new} that holds each bubble's parameters, then the Intersection of Union (IoU) is utilized as the weight for constructing the bipartite graph.

To assign the newly identified bubbles to their previous trajectories, we treat the bubbles in the current frame as the source group and the new bubbles in the next frame as the target group, and construct a bipartite matching graph like in Sect. \ref{sec:epigeometry}.
To impose the upwards and sideward motion constraints, we discard the edges where the later bubble is below the old bubble, and also discard the edges where the sideward motion exceeds a certain threshold.
For each bubble in the current frame, its expected position and bounding box are predicted by the Kalman Filter, and we compute the IoU between the predicted bounding box and the later bubble, and use the IoU as the weight of the edge in the graph.
As illustrated in Fig. \ref{fig:IoU_tracking}, the IoU expresses the similarity between the predicted bubble and the later detected bubble.
Finally, the best association set can be found via the Hungarian algorithm \cite{kuhn1955hungarian}.
For those bubbles which have no assignment to the previous trajectories, new trajectories are initialized.

\subsubsection{Counting at Reference Surface}
To avoid multiple counting of the bubbles, and to treat fast and slow rising bubbles in a consistent manner, we only count a bubble as valid and calculate its characteristics when its trajectory passes a virtual horizontal plane, the counting reference surface. This surface is defined by selecting a certain horizontal image row in the first camera. 
Consequently, we disregard bubble trajectories that start above the counting reference surface or bubbles that dissolve or cannot be tracked anymore before reaching the reference surface.

\section{Evaluation}
\label{sec:evaluation}

\subsection{Stereo Calibration}
\begin{table*}
	\begin{center}
		\caption{Camera intrinsic calibration results.}
		\label{tab:intrinsic_calib_result}
		\begin{tabularx}{0.7\textwidth}{lcc}
			\hline\noalign{\smallskip}
			Calibration parameters& Camera left & Camera right\\
			Focal length $f_x$ 	  &  1723.189   & 1711.854 	  \\
			Focal length $f_y$    &  1737.865   & 1719.751	  \\
			Principal point $c_x$ &  584.490  	& 507.474 	  \\
			Principal point $c_y$ &  362.619  	& 349.812 	  \\
			$k_1$				  &  -0.1087   	& -0.0716	  \\	
			$k_2$				  &  0.1184    	& 0.0106	  \\
			$p_1$				  &  -0.0031    & -0.0136	  \\
			$p_2$				  &  -0.0021 	& -0.0128	  \\
			\noalign{\smallskip}\hline
		\end{tabularx}
	\end{center}
\end{table*}

\begin{table*}
	\begin{center}
		\caption{Recalibration results of the relative orientation and translation of the stereo camera system. The last column shows the mean distance of the hand-picked photogrammetric marker points to their corresponding epipolar lines.}
		\label{tab:relative_rot_trans}
		\begin{tabularx}{1.0\textwidth}{lccc}
			\hline\noalign{\smallskip}
 			& Orientation(Quat.)        & Translation[mm] & Epi. error[px]	 \\
			Before &  (0.694, -0.020, 0.718, 0.033)		& (-287.11, -17.11, 303.88) & 5.59 $\pm$ 1.19 \\
			After  &  (0.698, -0.021, 0.715, 0.041)    	& (-286.52, -22.46, 302.79) & 0.96 $\pm$ 0.76 \\
			\noalign{\smallskip}\hline
		\end{tabularx}
	\end{center}
\end{table*}
After centering the cameras with the dome ports, we perform stereo calibration following the procedure described in Sect. \ref{sec:stereo_calib}. 
The camera intrinsic calibration results are presented in Table \ref{tab:intrinsic_calib_result} and the relative orientation and translation are presented in Table \ref{tab:relative_rot_trans} (first row).
The relative orientation is represented by a quaternion, where the first entry is the real part of the quaternion.
We have obtained an average reprojection error of 0.24 pixels over all chessboard corners in all stereo images.
Then, we back-project all image observations to the 3D calibration target plane using the calibrated parameters, we observed that the average error in 3D is 0.049mm.

\subsection{Recalibration by Bubble Adjustment}
\begin{figure}[!h]
	\begin{center}
		\subfloat[Drawing of the photogrammtric reference markers. ]{
			\includegraphics[width=0.6\textwidth]{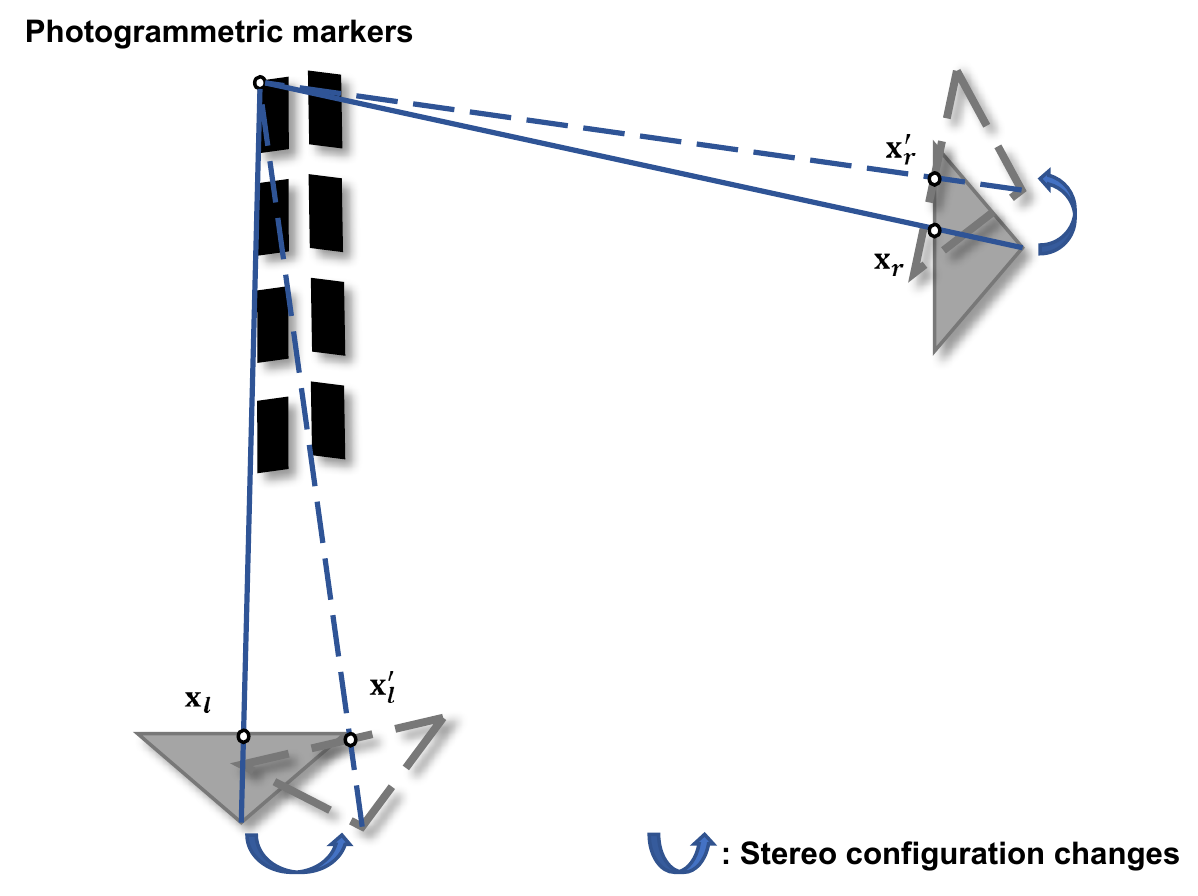}
		\label{fig:ba_recalib_markers_a}
		}\\
		\subfloat[Stereo images of the photogrammtric reference markers. ]{
			\includegraphics[width=1.0\textwidth]{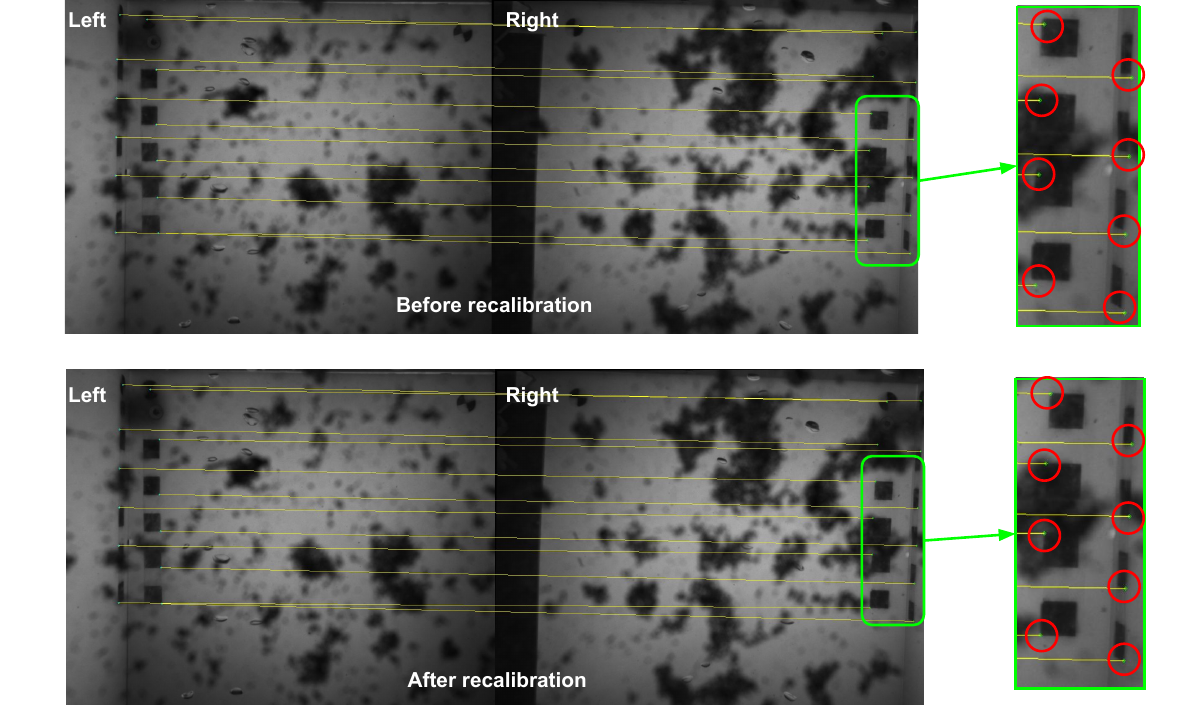}
		\label{fig:ba_recalib_markers_b}
		}
	\end{center}
\caption{(a), The drawing shows the reference markers (black squares) that are attached to the acrylic panels, seen by both cameras. (b), Stereo images captured in the deep sea where sediment particles and aggregates obscure the view and require  removal as background (see Sect. \ref{sec:BackgroundRemoval}). The yellow lines indicate the projection of the markers' corners from the left to right images. Top: Before recalibration, the projected corners in both left and right images are off from their true positions (only magnify the right image for better visualization). Bottom: After recalibration, the projections match their actual positions. }
\end{figure}

\begin{figure}[!h]
\begin{center}
	\subfloat[Before Recalibration (background removed)]{
		\includegraphics[width=1.0\textwidth]{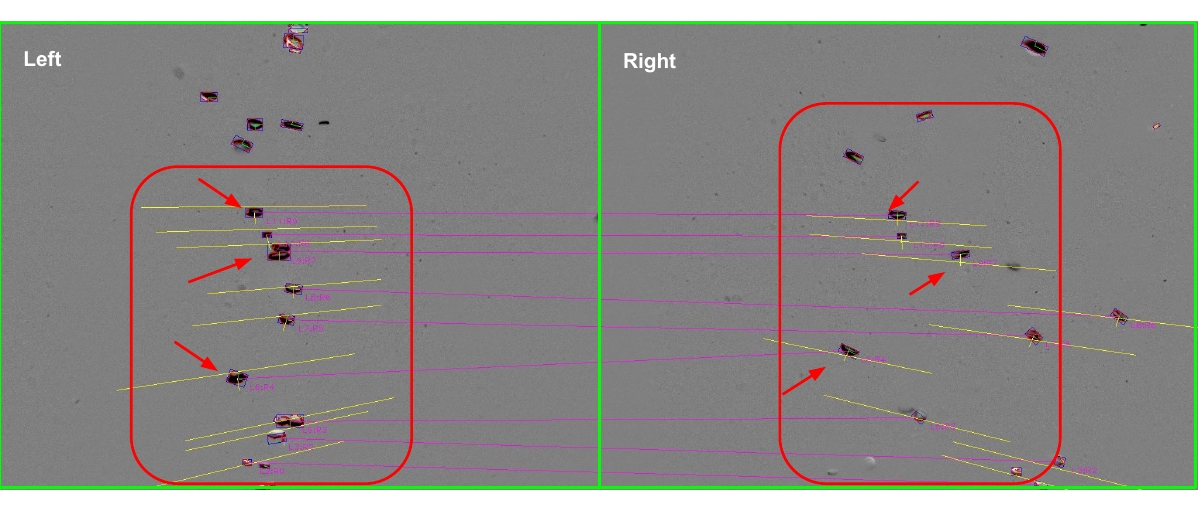}
	\label{fig:ba_recalib_before}
	}\newline
	\subfloat[After Recalibration (background removed)]{
		\includegraphics[width=1.0\textwidth]{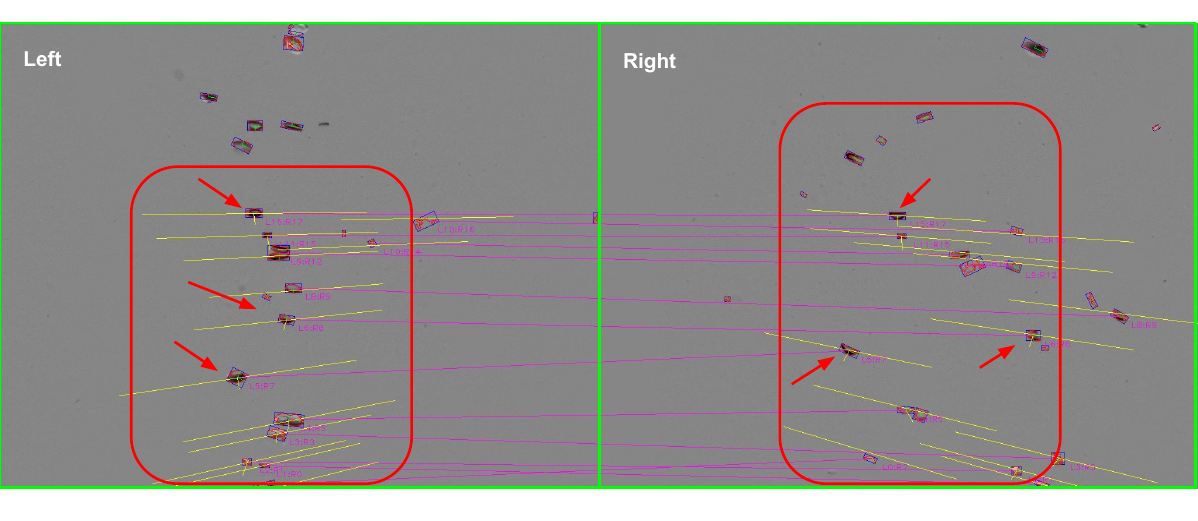}
	\label{fig:ba_recalib_after}
	}
\end{center}

\caption{Epipolar lines of corresponding bubbles in a pair of background removed stereo images. Top: Before recalibration, the epipolar lines (yellow) are off from the corresponding bubbles. Bottom: After recalibration, the epipolar lines intersect with the corresponding bubbles' silhouette.}

\end{figure}

One lesson learnt from the actual in-situ experiment is that the relative orientation and position of the stereo camera system does vary due to environmental pressure and temperature changes.
The metal frames and the plastic panels of the cameras and mirror mountings behave differently compared to a water pool at room temperature, where we took stereo camera calibration images.
Therefore tank/laboratory calibration results are not directly applicable to images acquired in the field and need to be treated carefully.
As shown in Fig. \ref{fig:ba_recalib_markers_a}, black square photogrammetric markers are attached to the acrylic panels and are visible in both cameras.
We first hand-picked points on the corners of the photogrammetric markers from the tank/laboratory calibration images,
and triangulate 3D locations of the marker points since we are certain that the stereo calibration is accurate enough.
Then, we project the triangulated marker points onto the in-situ deep sea images using the same set of stereo calibration parameters, however, we observe that these projections do not fit (see the red circles in Fig. \ref{fig:ba_recalib_markers_b}, top).
In addition, we calculate the mean epipolar distance (distance of a point to its corresponding epipolar line) of the marker points and obtained an eipipolar error of 5.59 pixels.
Similarly, the stereo epipolar matching results have shown that the epipolar lines of the bubble centers are significantly off\footnote{When back-projecting a pixel that lies inside the bubble contour, the back-projected viewing ray cuts through the 3D ellipsoid. Projecting this ray into the other image, the resulting epipolar line must clearly cut though the corresponding contour in the other image. If it is only tangent, or in case it does not even touch the corresponding contour, indicates that calibration is incorrect.} from the corresponding bubble centers in the other image (see Fig. \ref{fig:ba_recalib_before}).
Consequently, a recalibration is necessary, however, performing in-situ calibration with a moving small board inside the BBox is not feasible.
Therefore, we propose a self-calibration approach to refine the relative orientation and translation of the stereo cameras by directly utilizing the in-situ images.

The self-calibration is essentially an extension to Sect. \ref{sec:ba_bubble} where we use bubble adjustment with ellipse constraints to optimize the quadric representation of the bubble.
Extending the cost function of eq. \ref{eq:quadricenergy}, we now sum over multiple bubbles and jointly optimize all quadric projections $\mq C^k_*$, with respect to the quadric parameters in $\mq Q^k_*$ but also to the common relative orientation and position of the stereo camera at the same time:
\begin{equation}
E = \sum_{k} \sum_{i} \Vert \q x_{i,1}\trans \mq C^k_1 \q x_{i,1} \Vert^2 + \Vert \q x_{i,2}\trans \mq C^k_2 \q x_{i,2}\Vert^2	
\end{equation}
Again, $\mq C^k_1$ and $\mq C^k_2$ are the 2D conics projected from $k^{th}$ bubble (quadric) in the left and right image.
By accumulating multiple bubbles from multiple frames, a recalibration of the stereo camera system can be achieved using in-situ data.
This way the calibration adapts to the environment where the measurements are conducted. Note that different bubbles can be used for each frame such that bubble tracking over multiple frames is not necessary for applying this method.
The recalibrated relative orientation and translation are presented in Table \ref{tab:relative_rot_trans} (second row).
The orientation change in Euler angles (XYZ order) is $(0.596^\circ, -0.557^\circ, 0.708^\circ)$, and the translation change is (3.256mm, 1.391mm, 1.910mm).
After recalibration, the mean epipolar distance of the marker points is reduced to 0.96 pixels.
Note that the corner points on the photogramemtric markers are hand-picked, which do not have a sub-pixel level of accuracy.
As also can be seen from Fig. \ref{fig:ba_recalib_markers_b} (bottom) and Fig. \ref{fig:ba_recalib_after} that the projections of the photogrammetric marker points are in the correct positions, and the epipolar lines of the bubble contour masses now intersect with their corresponding bubble outlines.

\subsection{Dark Frames}
\begin{figure}[!th]
	\begin{center}
		\includegraphics[width=0.99\columnwidth]{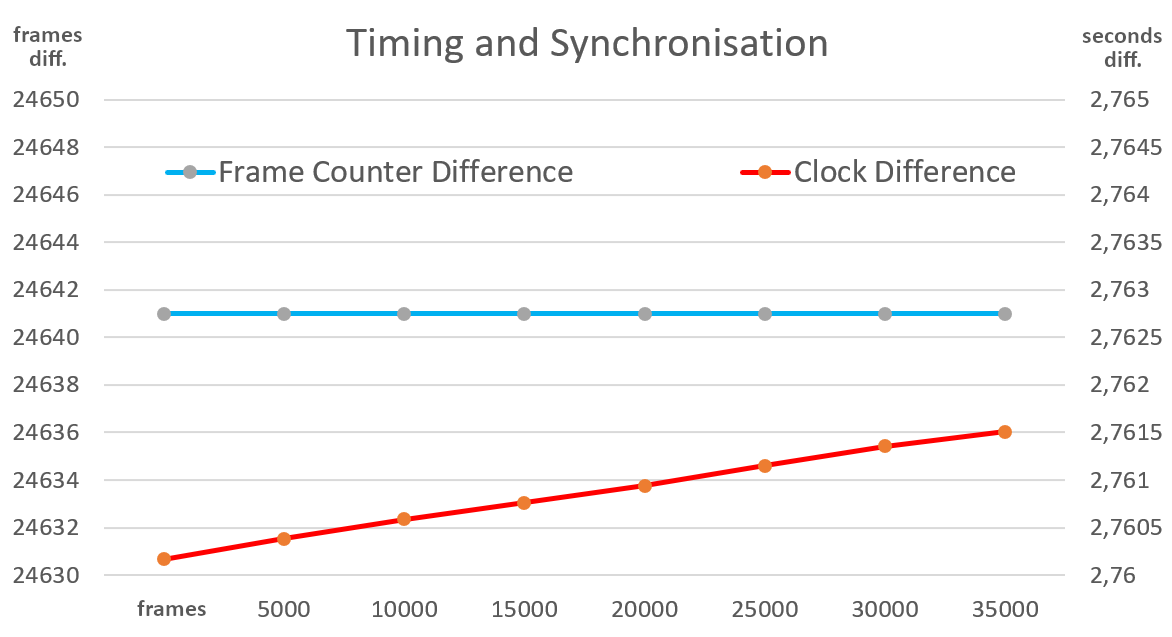}
	\end{center}
	\caption{Frame offset measured by black frames (top / blue curve) and clock difference of the computers in seconds (bottom / red curve). We observe that the black frames are detected reliably and no frames are lost. However, the recorded time stamp difference increases slightly due to clock drift.}
	\label{fig:clockdrift}
\end{figure}

We have verified that the dark frame detection works robustly and reliably. Using a frame rate of 80Hz we do not observe frame drops; which is verified by a dark frame appearing exactly every 5000 images for both camera image streams. At 100Hz sporadic frame drops were observed to appear when using the highest image resolution.
In Fig. \ref{fig:clockdrift} we exemplify the dark frame detection and the timing during a short experiment of taking about 40000 images with both cameras at 80Hz.
 Before the experiment, we deliberately did not synchronize the clocks of the two computers such that they differed by about 2.7 seconds.
We subtract both the frame counters and the timestamps recorded independently by both computers when a black frame is detected. It can be seen that the counter difference is constant, whereas the time is drifting slightly. 
For this short experiment of 437.5 seconds, we observe a clock drift between the two computers by 1.3ms, which means 0.27 seconds per day. Since the frame counter offset stays stable for the detected dark frames and the images were triggered at the same time from an external micro controller we can still associate the correctly corresponding images from both cameras.
A significant drift of the computer clocks compared to the trigger signal could theoretically influence computing the bubble rise speeds and flow rates.
 However, the rather consistent clock drift found for the computers ( +0.0005\% and +0.0002\%) will not cause errors that are significant compared to other uncertainties of the proposed method.

\subsection{Background Removal}
\begin{figure}[!ht]
	\begin{center}
		\subfloat[a]{
			\includegraphics[width=0.7\textwidth]{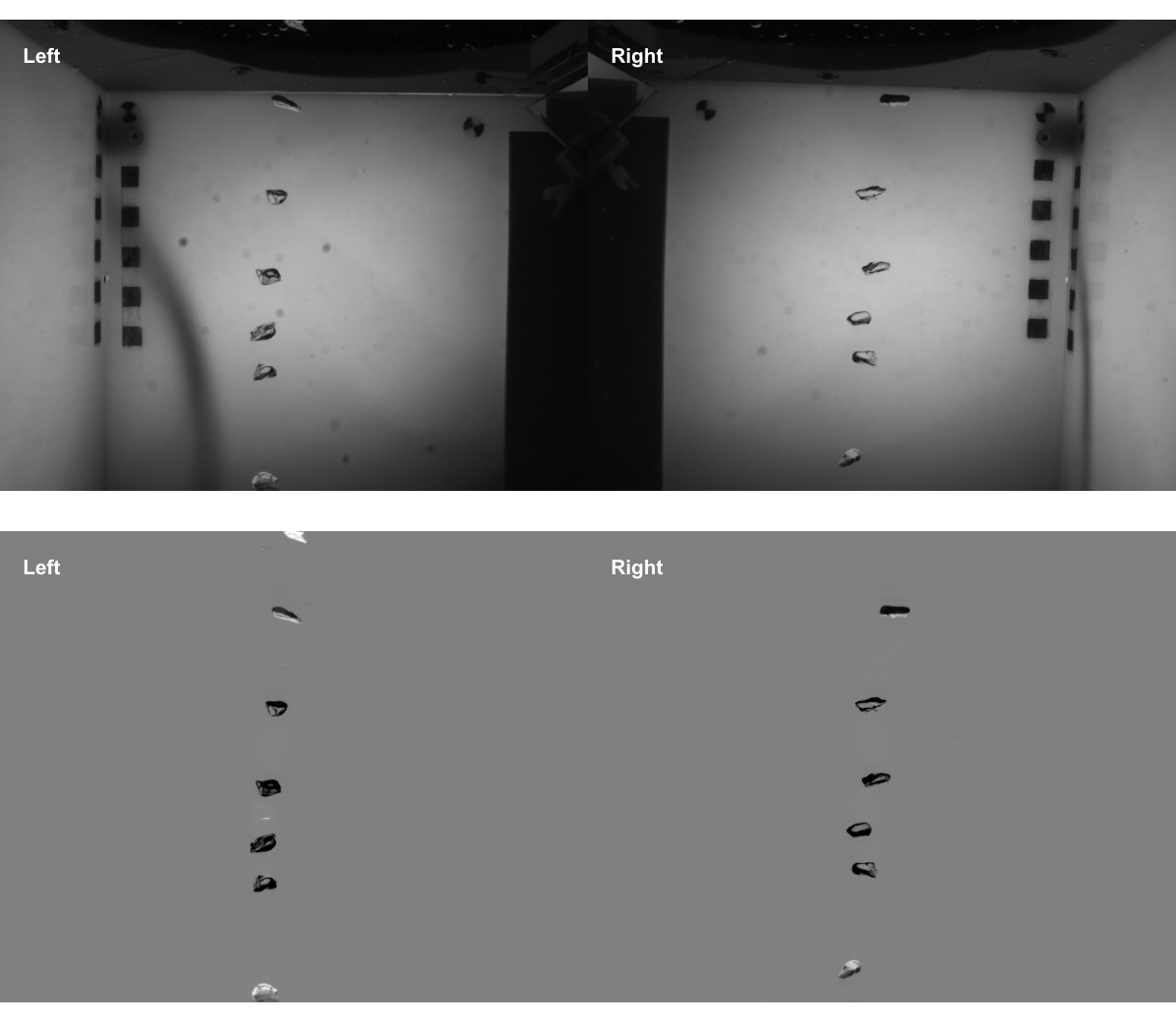}
		\label{fig:background_rm_a}
		}\\
		\subfloat[b]{
			\includegraphics[width=0.7\textwidth]{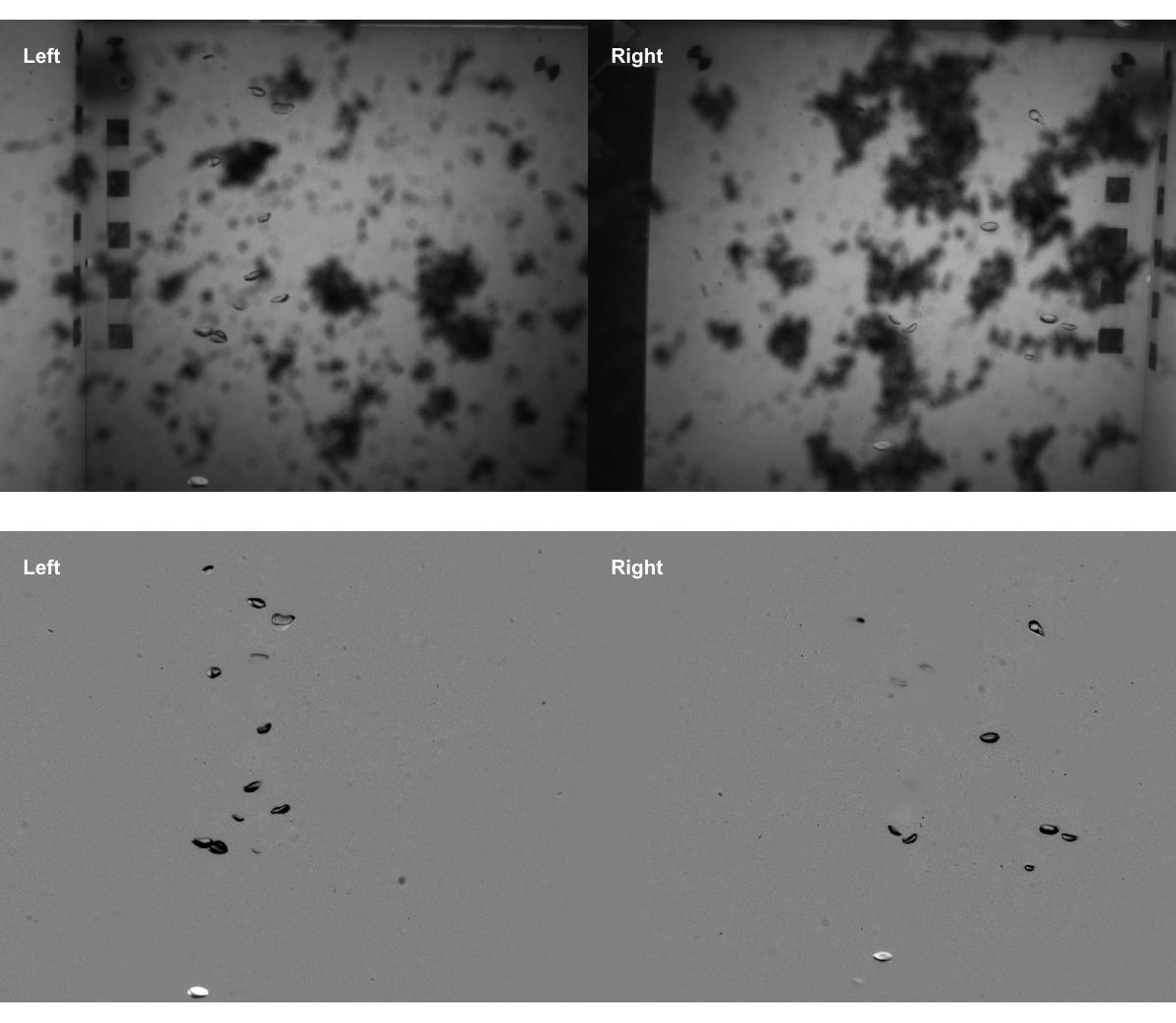}
		\label{fig:background_rm_b}
		}
	\end{center}
	\caption{Two sample results of the background removal. In each sub-figure, the top row shows the original stereo images and the bottom row shows the resulting foreground images.}
	\label{fig:background_rm}
\end{figure}

Fig. \ref{fig:background_rm} shows two sample results of the background removal from two different data sets.
It can be seen that the temporal median filtering algorithm works perfectly when the water inside the BBox is clear (see Fig. \ref{fig:background_rm_a}), and it also works robustly enough with real ocean images, even though some sediments stuck on the camera glass dome and block part of the cameras' view (Fig. \ref{fig:background_rm_b}). 
If the occluding object is too large (see Fig. \ref{fig:failure_case_1}), we may not be able to record the bubble completely, and the bubble detection could fail.
 An automatic dome port and mirror cleaning system could be installed in a future improved BBox version.

\subsection{Known Reference Objects}
\begin{figure}[!th]
	\begin{center}
		\includegraphics[height=72pt]{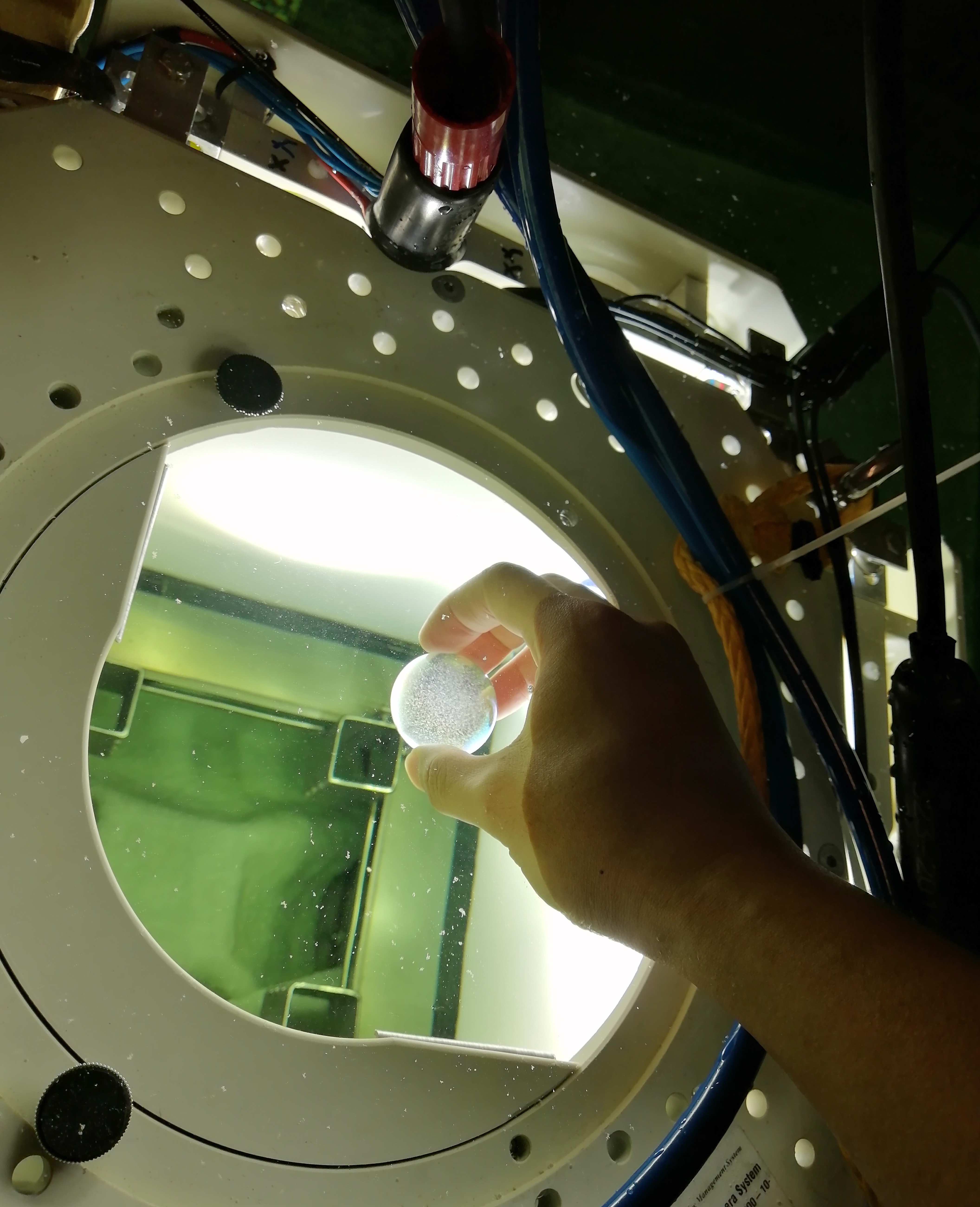}
		\includegraphics[height=72pt]{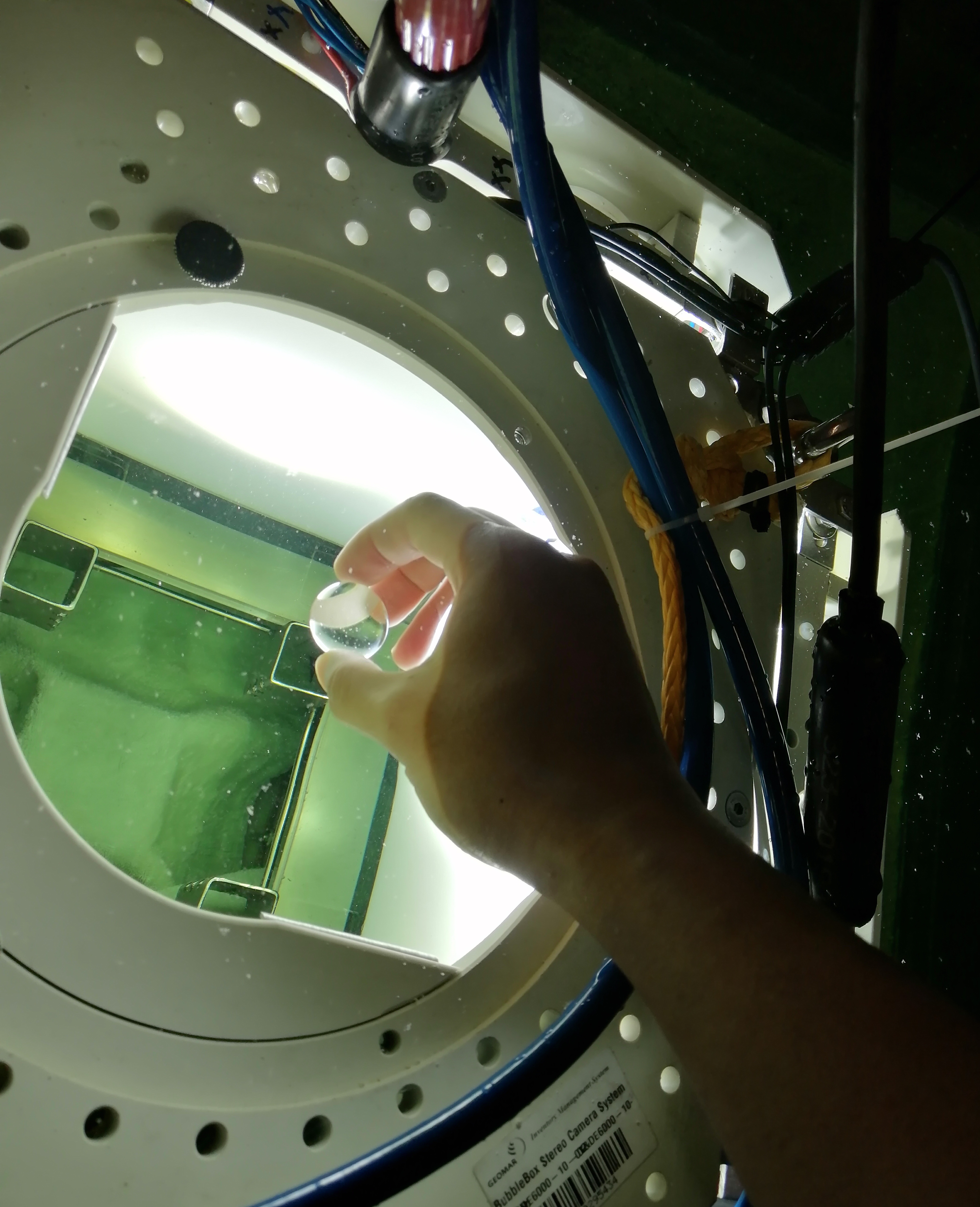}
		\includegraphics[height=72pt]{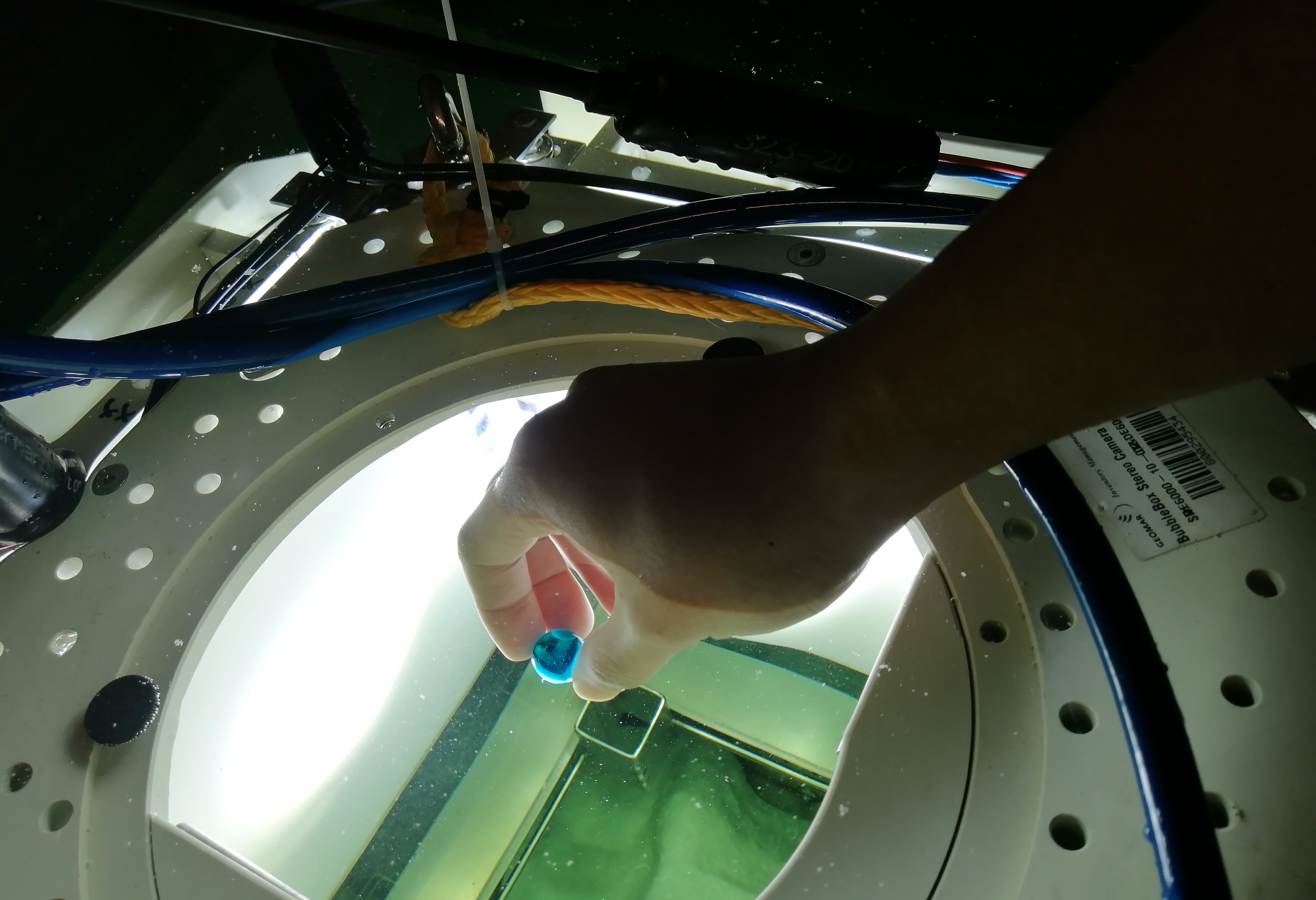}
		\includegraphics[height=72pt]{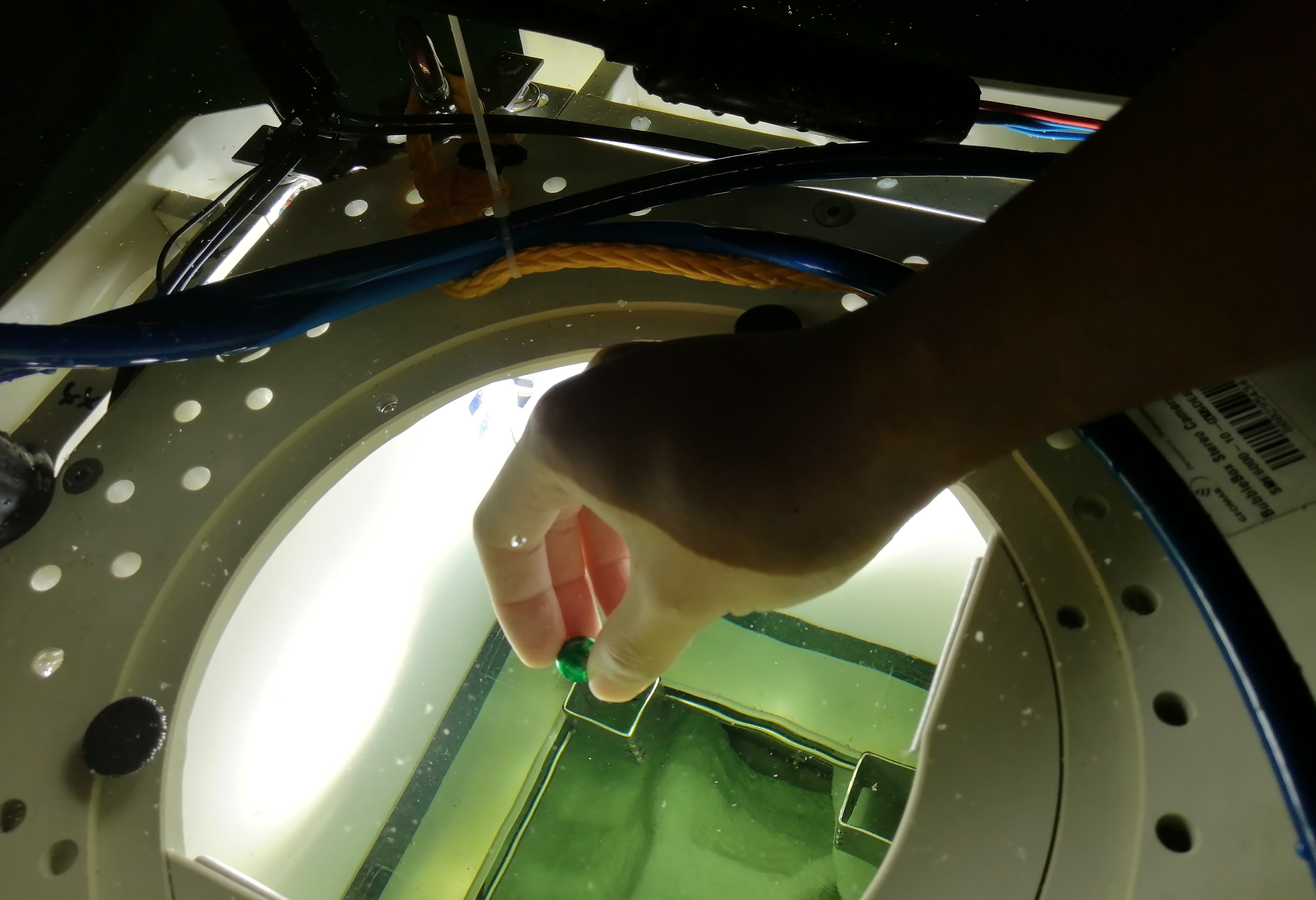}
	\end{center}
\caption{Evaluation using known reference objects (4 glass marbles with different radii).}
\label{fig:glass_marbles}
\end{figure}
To evaluate the accuracy of the proposed ellipsoid triangulation technique and later bubble adjustment with ellipse constraints, we conduct measurements using the BBox to photograph glass marbles of approximate radii 6mm, 8mm, 13mm and 17mm, falling through the rise corridor in water (see Fig. \ref{fig:glass_marbles}).
The diameter of each marble was measured by a vernier caliper.
Since the marbles are not perfectly spherical, to obtain an accurate reference, we measure each glass marble 10 times while rotating it and use the average diameter as reference value.
Next, we drop each glass marble through the corridor of the instrument 10 times to obtain over 100 sample measurements per marble size.
We perform bubble detection, epipolar geometry matching, ellipsoid triangulation and bubble adjustment on images where the marble could be seen, and the evaluated results of the equivalent diameter \footnote{The equivalent diameter of a non-spherical particle is equal to a diameter of a spherical particle that exhibits identical volume to that of the investigated non-spherical particle.} are shown in Table \ref{tab:glass_marbles}, together with standard deviations. 

\begin{table*}
	\begin{center}
		\caption{Evaluation results of the glass marbles.}
		\label{tab:glass_marbles}
		\begin{tabularx}{1.0\textwidth}{lccc}
			\hline\noalign{\smallskip}
			& Proposed method [mm]& Vernier caliper [mm] & Err. [\%]\\
			Marble R17 &  34.33 $\pm$ 0.09  & 34.72 $\pm$ 0.10& 1.12 \\
			Marble R13 &  25.00 $\pm$ 0.20 & 25.10 $\pm$ 0.01& 0.39\\
			Marble R8 &  16.01 $\pm$ 0.16 & 15.91 $\pm$ 0.25& 0.63 \\
			Marble R6 &  12.68 $\pm$ 0.16 & 12.41 $\pm$ 0.13& 2.2 \\
			\noalign{\smallskip}\hline
		\end{tabularx}
	\end{center}
\end{table*}

\begin{figure}[!h]
	\begin{center}
		\subfloat[Glass marble R17 (radius $\approx$ 17mm)]{
			\includegraphics[width=\textwidth]{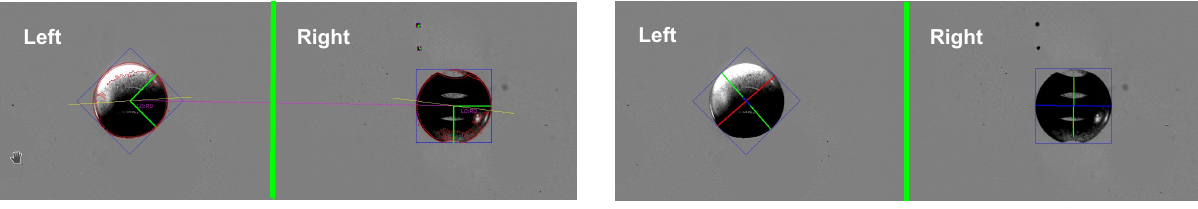}
		}\newline
		\subfloat[Glass marble R13 (radius $\approx$ 13mm)]{
			\includegraphics[width=\textwidth]{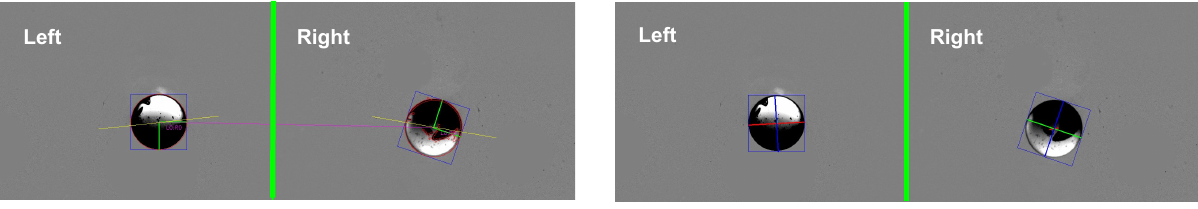}
		}\newline	
		\subfloat[Glass marble R8 (radius $\approx$ 8mm)]{
			\includegraphics[width=\textwidth]{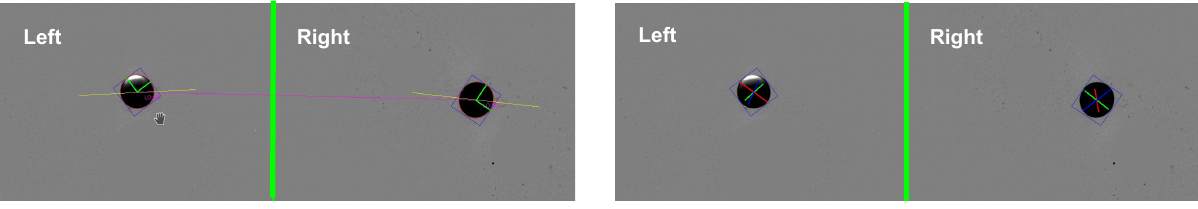} 
		}\newline
		\subfloat[Glass marble R6 (radius $\approx$ 6mm)]{
			\includegraphics[width=\textwidth]{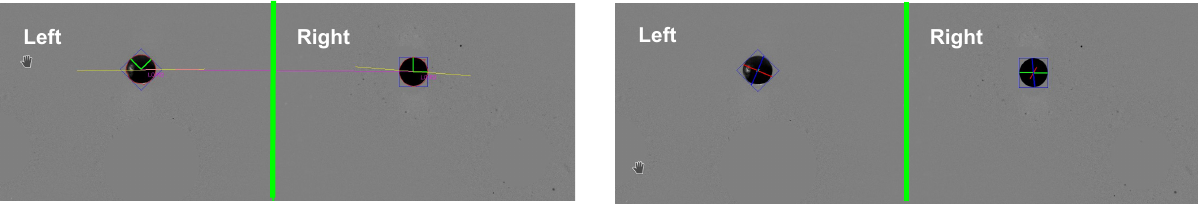}
		}
	\end{center}
	\caption{Image samples of the intermediate results of the known reference objects (stereo images are concatenated into one image). From top to bottom:  samples of differently sized glass marbles. From left to right: stereo epipolar geometry matching and the re-projected ellipsoid. In the left part of each sub-figure, the glass marbles are identified and marked as red outlines and a blue bounding box; The violet lines connect the bubble correspondences; the yellow lines are the epipolar lines of the contour in the other image (note that the contours of the corresponding bubbles are not sets of corresponding points, they describe tangent rays to the ellipsoid from entirely different perspectives). In the right part of each sub-figure, the final reconstructed 3D ellipsoid is projected onto the image, and its $X-$, $Y-$, $Z-$ axis are shown in red, green, and blue lines respectively.}
	\label{fig:glass_marbels_intermediate}
\end{figure}

It is shown that the estimation accuracy of the reference objects is in the range of $1 - 2\%$ in the equivalent diameter under ideal conditions, which is also in an agreement with the calibration residuals.
Some results of intermediate steps can be found in Fig. \ref{fig:glass_marbels_intermediate}, where the epipolar geometry matching results are shown in the left part of the sub-figures, while the re-projected ellipsoids are shown in the right.
Here, to re-project an ellipsoid, first its center point is projected onto the image, followed by projecting the 6 endpoints of the ellipsoid axes and connecting them.

\subsection{Controlled Experiments}
\begin{figure}[!t]
	\includegraphics[width=1.0\textwidth]{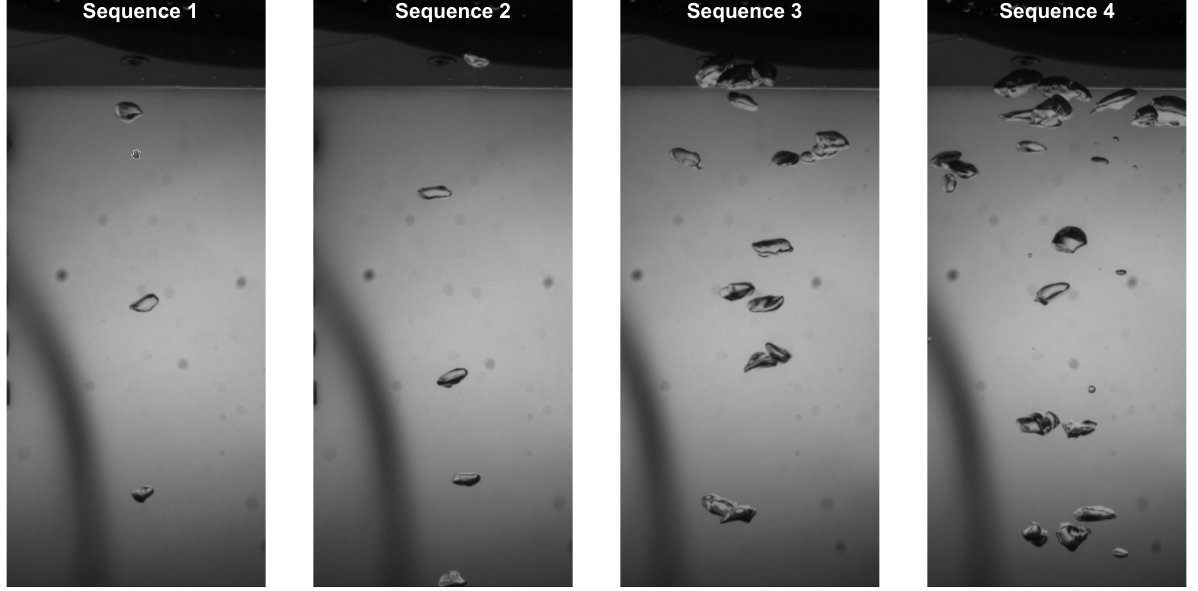}
\caption{An overview of the sequences with different gas flow rates for the controlled experiments.}
\label{fig:tlz_data_overview}
\end{figure}

To demonstrate the effectiveness of the complete workflow also for actual bubbles, we conduct a controlled experiment.
We set up the instrument in a water pool with an air-bubble generator placed underneath the instrument.
The generator is able to produce air bubbles of different sizes and at different flow rates to control the bubble density of the upward rising stream.
A cylinder is used to collect the released gas bubbles on top of the instrument and to measure the overall trapped gas volume.
The instrument records photos of the rising bubbles as during in-situ observations in the ocean, and the images are analyzed by the bubble stream characterization approach as outlined in Sect. \ref{sec:bubble_stream_charac}.
The overall bubble volume measured using the stereo camera bubble stream characterization approach in comparison to the cylinder trapping measurement is used to evaluate the accuracy and robustness of the method.
We start with low bubble stream density and gradually increase the gas flow of the bubble generator.
An overview of the sequences can be seen in Fig. \ref{fig:tlz_data_overview}, and the experiment results are shown in Table \ref{tab:control_experiment}.

\begin{table*}[!t]
	\begin{center}
		\caption{Bubble stream characterization results of the controlled experiments.}
		\label{tab:control_experiment}
		\begin{tabularx}{1.0\textwidth}{lccccccc}
			\hline\noalign{\smallskip}
			Seq-& Gas  & Occlu- & Duration & Cylinder & Estimated & Volume& Eq.radius\\
			ence& flow &sions & & read [ml] & vol.[ml] & err.[\%]& err.[\%]\\
			\hline
			1 & low  &no &7 min  & 81  & 84.4 & 4.2&  1.4\\ 
			2 & med.  &no &1 min & 105 & 109.2  & 4.0&  1.3\\
			3 & high   &yes &50 sec & 110 & 121.1  & 10.1&  3.3\\
			4 & higher &many &30 sec & 135 & 188.12  & 39.35&  11.2 \\
			\noalign{\smallskip}\hline
		\end{tabularx}
	\end{center}
\end{table*}

\begin{figure}
	\begin{center}
		\subfloat[Medium bubble density (sequence 2)]{
			\includegraphics[width=0.78\textwidth]{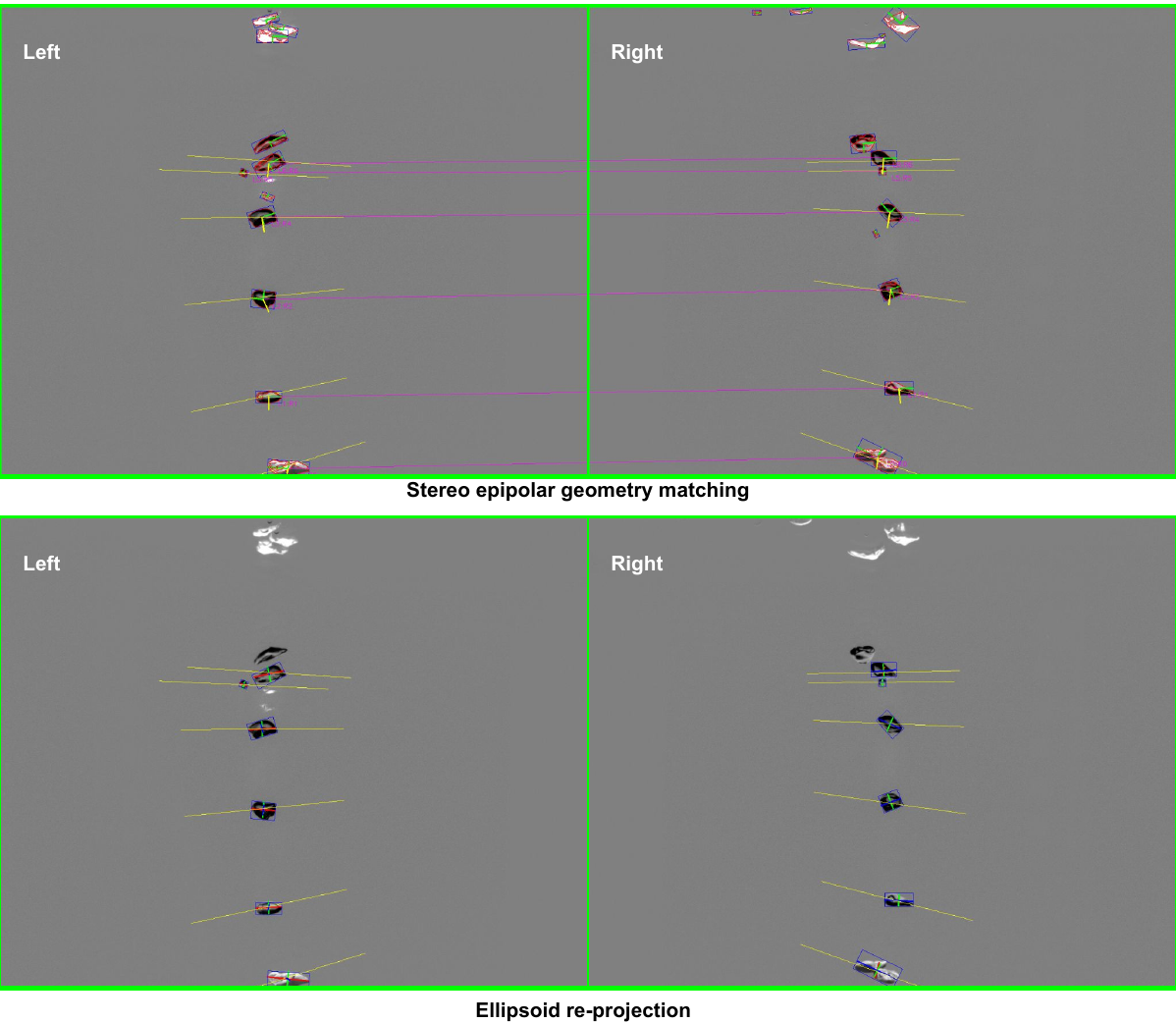}
			\label{fig:TLZ_result_1}
		}\\
		\subfloat[High bubble density (sequence 4)]{
			\includegraphics[width=0.78\textwidth]{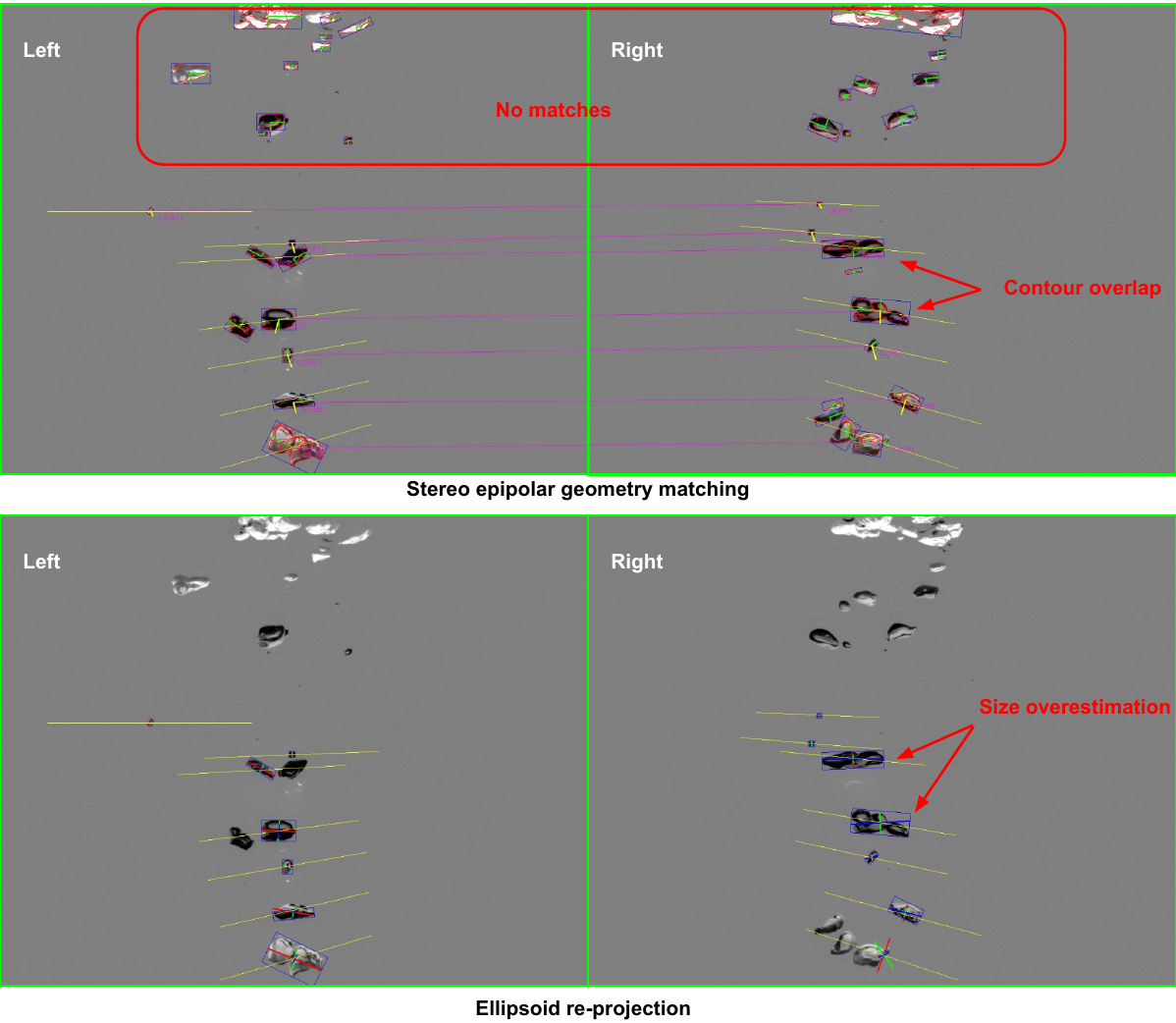}
			\label{fig:TLZ_result_2}
		}
		
	\end{center}
\caption{Sample results of stereo epipolar geometry matching and ellipsoid re-projection from the controlled experiments.}
\end{figure}

As can be seen from Table \ref{tab:control_experiment}, with only one single bubble stream, the overall volume estimation shows a relative volume error of 4.2\% (sequence 1) and 4.0\% (sequence 2).
Since the bubble volume increases with the third power of the radius, this can be interpreted as an equivalent radius error of slightly more than 1\%. 
This is close to the glass marble experiment results and shows that the ellipsoid model assumption works very well for the present bubble size range.
Results from the intermediate steps are shown in Fig. \ref{fig:TLZ_result_1}, and show that the ellipsoid re-projections well fit the identified bounding boxes of the bubbles.
However, we can also see that the accuracy decreases with increasing flow rate due to bubble contours overlapping in the image, especially in sequence 4 where the bubble stream density is very high.
One explanation is that the overlapping bubble contours in the image are merged into a bigger contour which encloses the bubble cluster such that the ellipsoid size of this cluster is overestimated.
In addition, it also introduces ambiguity both in the stereo (epipolar) matching and the bubble tracking (non-equal number of bubbles identified, see Fig. \ref{fig:TLZ_result_2}), which adds an additional error to the final volume estimation.
Nevertheless, for a moderate amount of overlapping bubbles in the image (like in sequence 3 of Fig. \ref{fig:tlz_data_overview}), we still obtain a result of an overall volume error of 10\% (or, equivalent radius error of around 3\%).
Note that also the physical collection of the gas using the trapping cylinder comes with a relevant measurement uncertainty that however is difficult to quantify for our experiment.
 In particular for the more turbulent high flow rates as in sequences 3 and 4 some bubbles stuck to the cylinder entrance and did not get collected. 
 Additional uncertainty arises in reading the cylinder measurement (surface tension of collected bubble volume) and due to the different water pressures (clearly below 5\%) at the reference surface of the photo evaluation and the cylinder reservoir, potentially leading to different gas expansions. We therefore consider an overall volume difference of about 10\% still a reasonable agreement.

\subsection{Data analysis from in-situ measurements from offshore Oregon, Pacific}
%\begin{figure}
%	\begin{center}
%	\includegraphics[width=0.6\textwidth]{eval/research_area.png}
%	\end{center}
%
%\caption{Research area of the Falkor cruise. Image courtesy: Schmidt Ocean Institute.}
%\label{fig:research_area}
%\end{figure}
The instrument has been deployed at the Cascadia Margin offshore Oregon in the Pacific Ocean during Falkor Cruise 'Observing Seafloor Methane Seeps at the Edge of Hydrate Stability' (FK190612), jointly with several other instruments to analyze bubble flow rates. A detailed analysis of the observations of the BBox's and other observations is given in \cite{veloso2022bboxgquant}.
% Fig. \ref{fig:research_area} shows the research area where the green circles indicate the deployment locations.
Instead, we only exemplify some results here.
The instrument was deployed on top of seep sites by an ROV (also see Fig. \ref{fig:falkor_bbx}).
We selected 6 sequences from a number of interesting time series and report the evaluation results in Table \ref{tab:result_pacific_ocean}. In this manuscript, $\pm$ indicates standard deviation. 
\begin{table*}
	\begin{center}
		\small
		\caption{Sample results of bubble stream characterization on data from the Pacific Ocean.}
		\label{tab:result_pacific_ocean}
		\begin{tabularx}{1.0\textwidth}{lcccccc}
			\hline\noalign{\smallskip}
			Date, time& Count& Volume [ml] & Flow rate [ml/s] & Diameter [mm] & Velocity [cm/s]\\
			06/19, 16:50& 370& 40.33 & 0.646 & 5.71 $\pm$ 0.51 & 26.21\\
			06/19, 17:05& 349& 36.78 & 0.591 & 5.70 $\pm$ 0.50 & 26.33\\
			06/20, 09:09& 453& 37.31 & 0.591 & 5.20 $\pm$ 0.62 & 28.22\\
			06/20, 09:24& 462& 33.93 & 0.565 & 5.00 $\pm$ 0.60 & 28.65\\
			06/20, 19:11& 544& 41.97 & 0.699 & 5.03 $\pm$ 0.56 & 27.79\\
			06/20, 19:26& 520& 41.13 & 0.685 & 5.08 $\pm$ 0.54 & 27.86\\
			\noalign{\smallskip}\hline
		\end{tabularx}
	\end{center}
\end{table*}
\begin{figure}[!h]
\def\svgwidth{1.0\textwidth}
{
	%% Creator: Inkscape 1.1.1 (1:1.1+202109281949+c3084ef5ed), www.inkscape.org
	%% Accompanies image file 'bubble_report.pdf' (pdf, eps, ps)
	\begingroup%
	\makeatletter%
	\providecommand\color[2][]{%
		\errmessage{(Inkscape) Color is used for the text in Inkscape, but the package 'color.sty' is not loaded}%
		\renewcommand\color[2][]{}%
	}%
	\providecommand\transparent[1]{%
		\errmessage{(Inkscape) Transparency is used (non-zero) for the text in Inkscape, but the package 'transparent.sty' is not loaded}%
		\renewcommand\transparent[1]{}%
	}%
	\providecommand\rotatebox[2]{#2}%
	\newcommand*\fsize{\dimexpr\f@size pt\relax}%
	\newcommand*\lineheight[1]{\fontsize{\fsize}{#1\fsize}\selectfont}%
	\ifx\svgwidth\undefined%
	\setlength{\unitlength}{343.71109009bp}%
	\ifx\svgscale\undefined%
	\relax%
	\else%
	\setlength{\unitlength}{\unitlength * \real{\svgscale}}%
	\fi%
	\else%
	\setlength{\unitlength}{\svgwidth}%
	\fi%
	\global\let\svgwidth\undefined%
	\global\let\svgscale\undefined%
	\makeatother%
	\begin{picture}(1,0.4)%
	\lineheight{1}%
	\setlength\tabcolsep{0pt}%
	\put(0,0){\includegraphics[width=\unitlength,page=1]{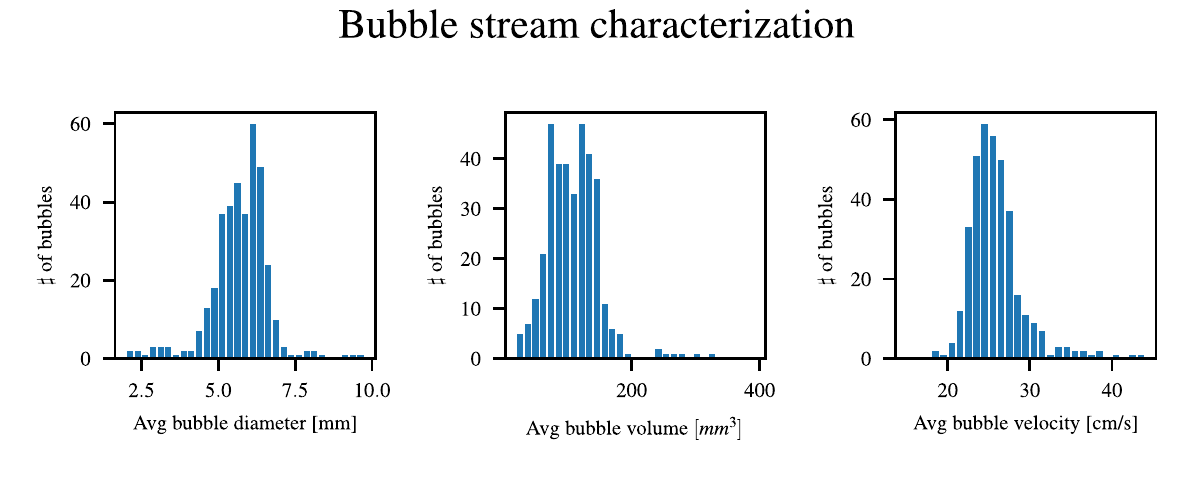}}%
	\end{picture}%
	\endgroup%
	
}
\caption{Bubble stream characterization results for a sequence on 2019/06/19, 16:50. Note that the axes of the bubble diameter and bubble volume histograms are related by a cubic stretch, which also lets the histograms appear slightly deformed, on top of different bin boundaries.}
\label{fig:bubble_stream_plot}
\end{figure}

\begin{figure}[!h]
	\begin{center}
		\includegraphics[width=0.78\textwidth]{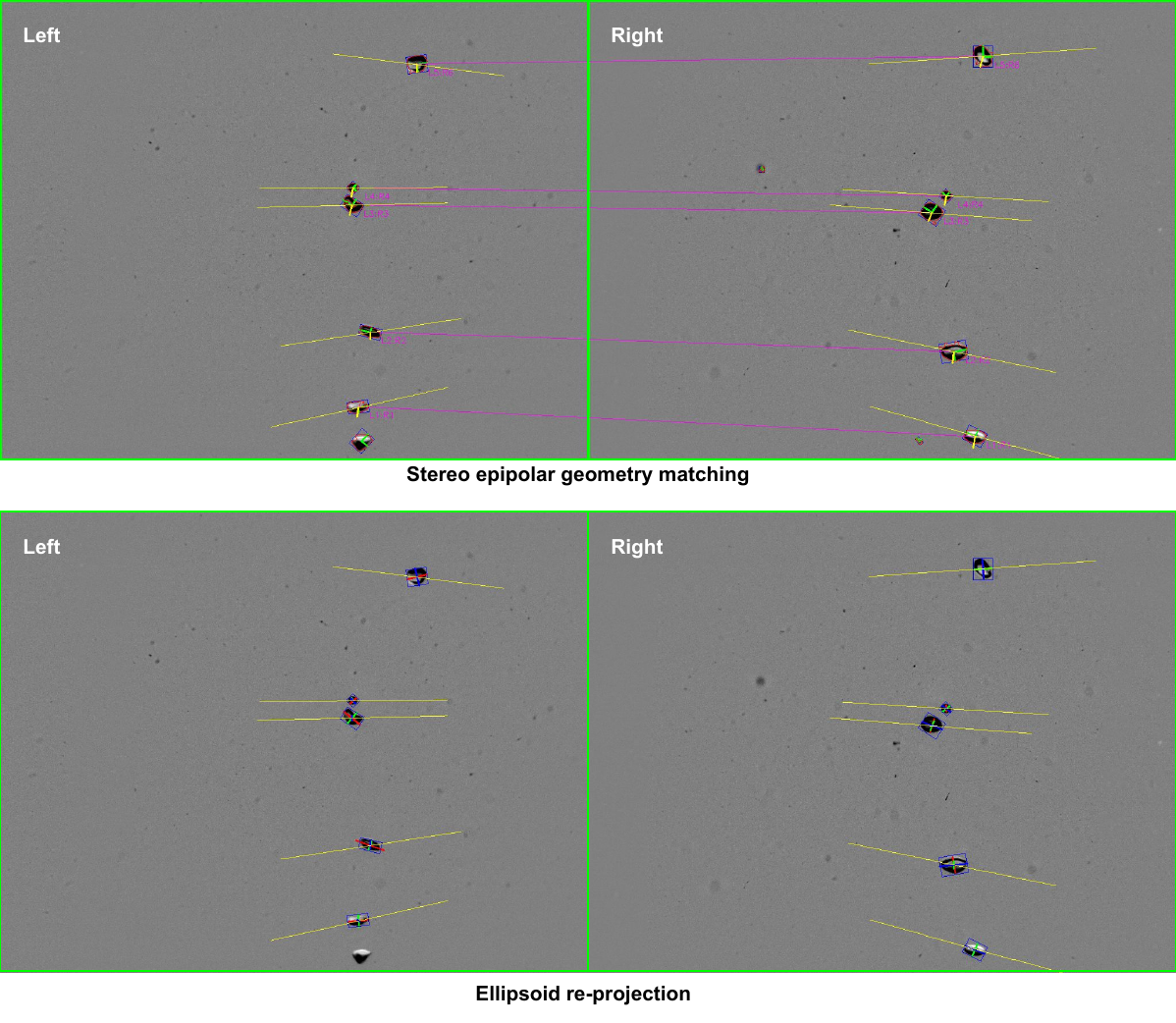}
		\includegraphics[width=0.78\textwidth]{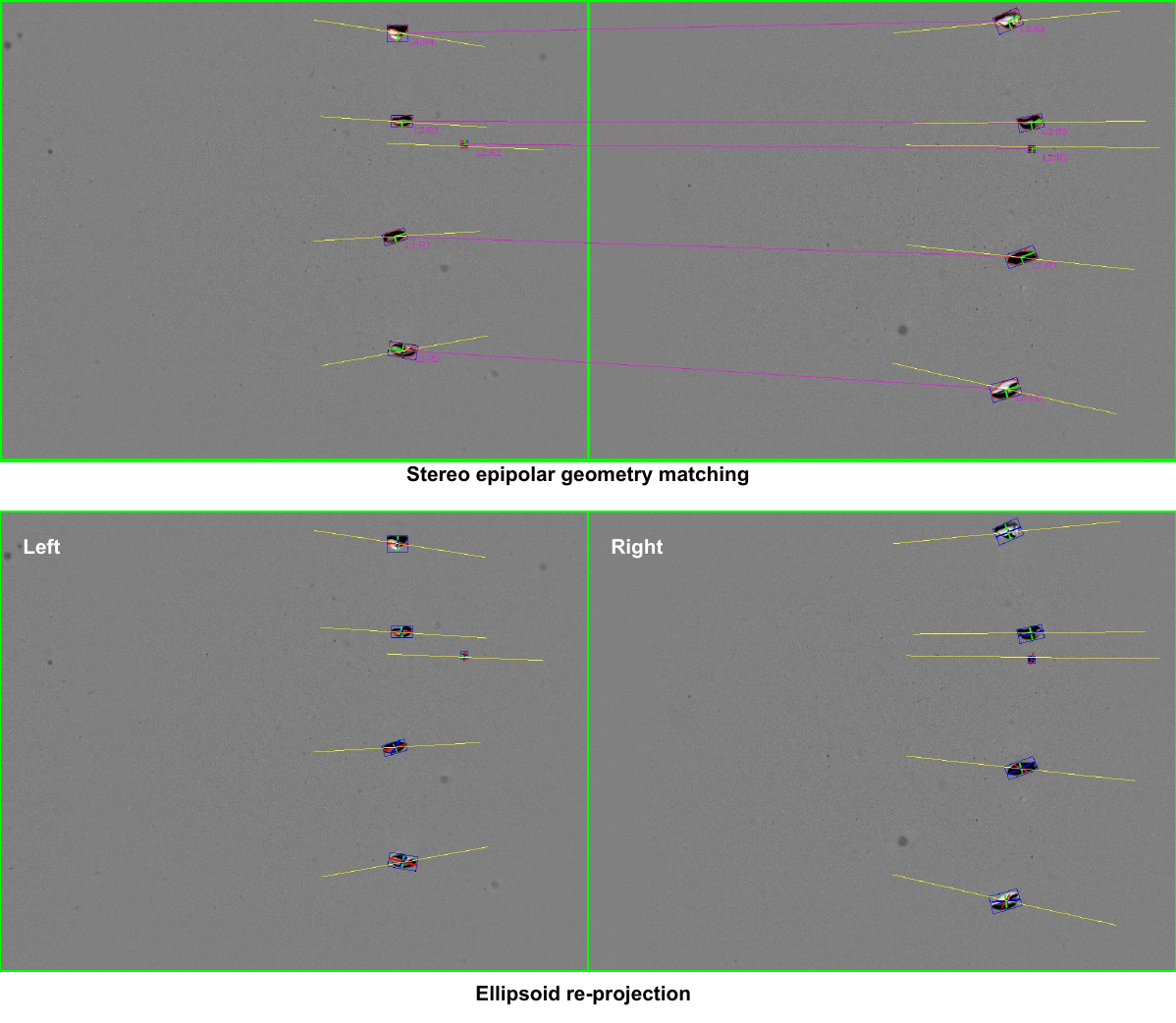}
	\end{center}
\caption{Sample results of stereo epipolar geometry matching and ellipsoid re-projection for sequence 06/19, 16:50.}
\label{fig:falkor_results_matching_ellipsoid}
\end{figure}

\begin{figure}[!h]
	\includegraphics[width=0.98\textwidth]{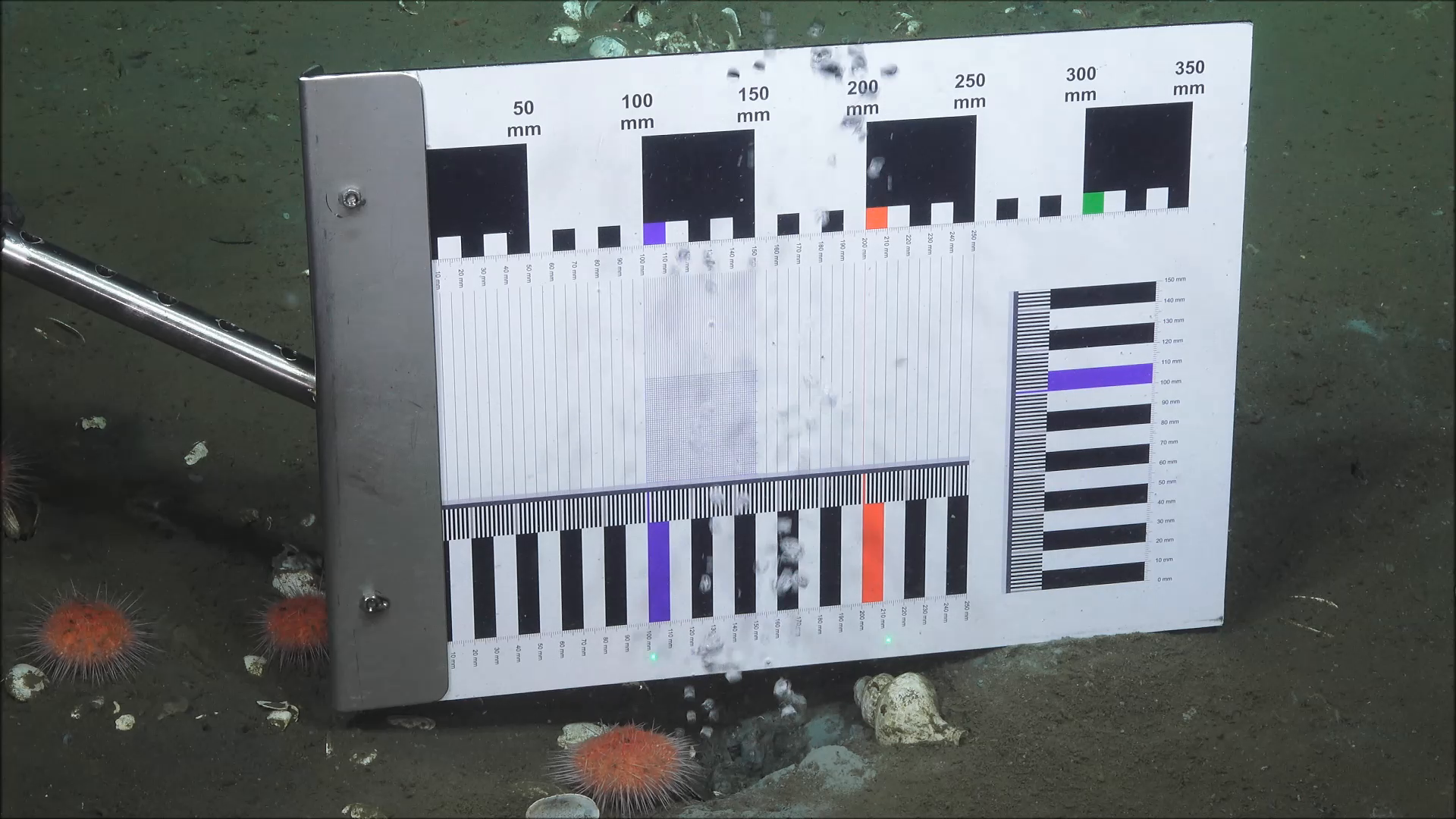}
\caption{A scale-bar board was used to roughly examine the estimated bubble sizes and rising velocity. Picture taken by ROV SuBastian, Schmidt Ocean Institute.}
\label{fig:scale_bar}
\end{figure}

Exemplarily, we show the bubble stream characterization results and some sample  results from intermediate steps for the first sequence in Fig. \ref{fig:bubble_stream_plot} and Fig. \ref{fig:falkor_results_matching_ellipsoid}.
The algorithm correctly finds bubble correspondences, and the re-projected ellipsoids are well located inside the bounding boxes of the identified bubbles.
No absolute ground truth measurements with the trapping cylinder could be performed during the cruise. However, we positioned a scale-bar board (see Fig. \ref{fig:scale_bar}) next to the measured bubble streams prior to the stereo camera measurements and estimated bubble radii and rise velocity from the ROV camera footage.
Using such measurements is common practice when no dedicated instrument can be used to provide important estimates of size distribution and overall bubble flow
\cite{Nikolovska08hydroacoustic,Sahling14svalbard,SCHNEIDERVONDEIMLING2011867}.
The accuracy of such measurement techniques is significantly poorer than the accuracy of the stereo camera measurements. However it allowed to verify that our gross estimates of bubble sizes of mainly 5mm to 7mm diameter are within a reasonable range\cite{veloso2015new}. 

\section{Discussion}
Both, the real data as well as the ground truth experiments indicate that the overall approach works well, even in the deep sea under real conditions, and that the measurement process is robust even in presence of sediment particles in the water and other nuisances.
 The achieved accuracy under good conditions is much better than can be expected from monocular methods. However, there are some limitations and sometimes complete failures in the practical applications:

\subsubsection*{Both Cameras Must See the Bubbles}
\begin{figure}
	\begin{center}
	\subfloat[Failure case 1]{
		\includegraphics[width=0.45\textwidth]{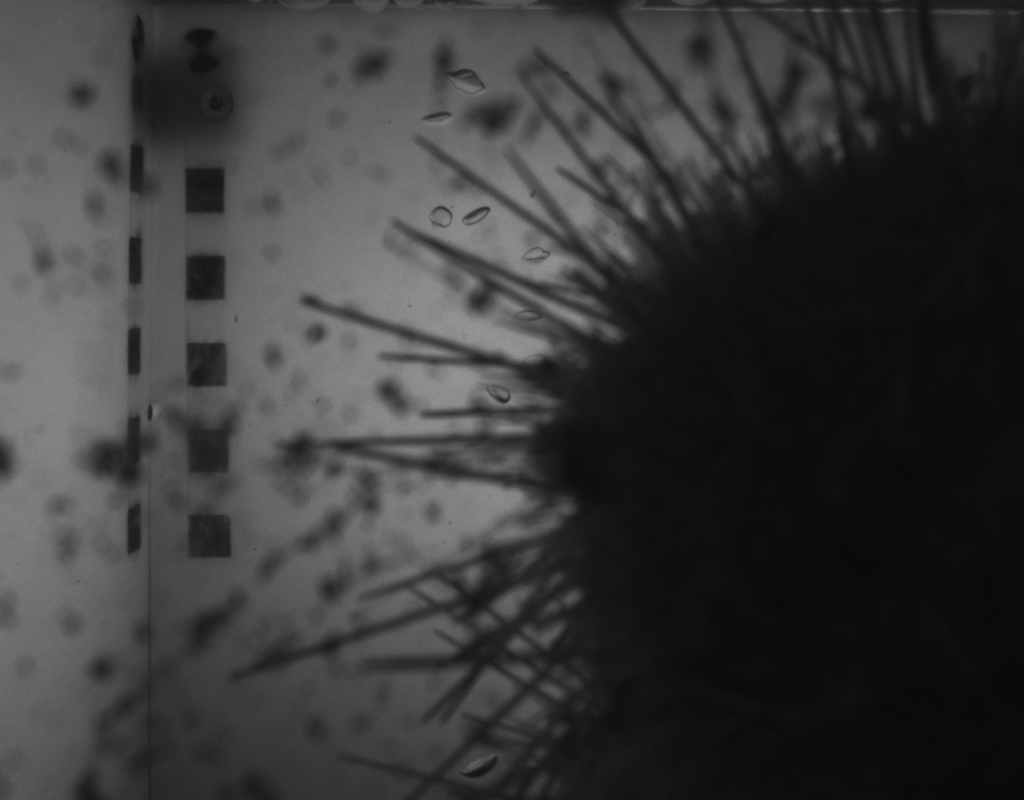}
		\includegraphics[width=0.45\textwidth]{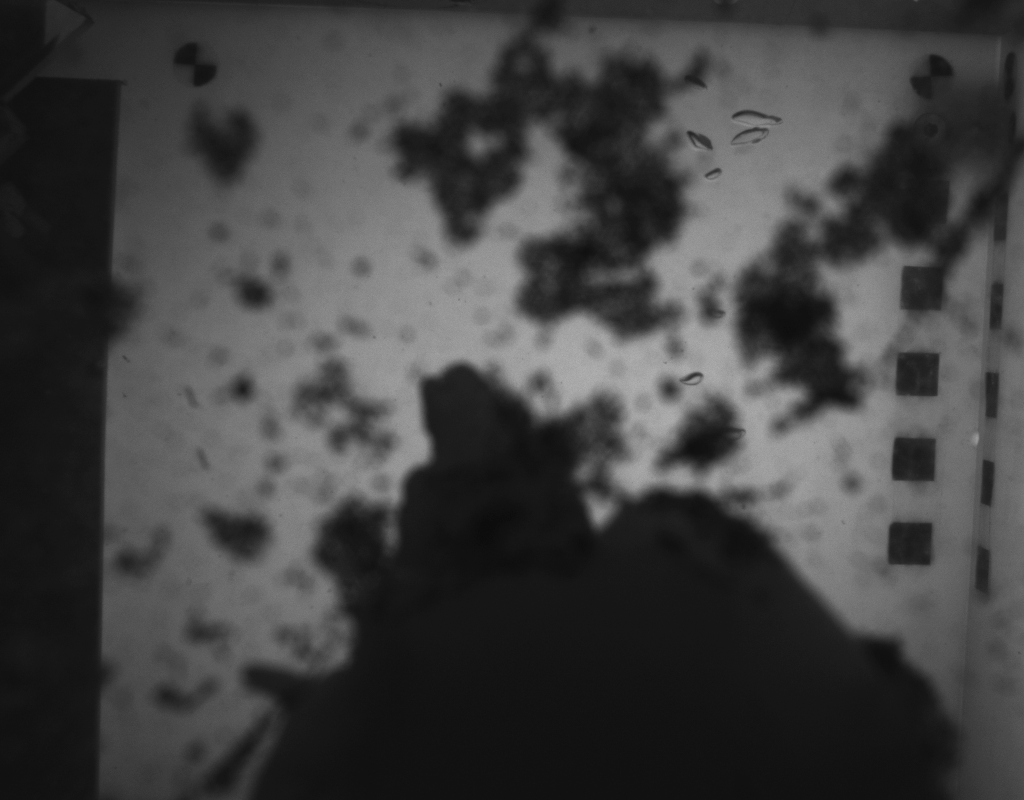}
	\label{fig:failure_case_1}
	}\\
	\subfloat[Failure case 2]{
		\includegraphics[width=0.9\textwidth]{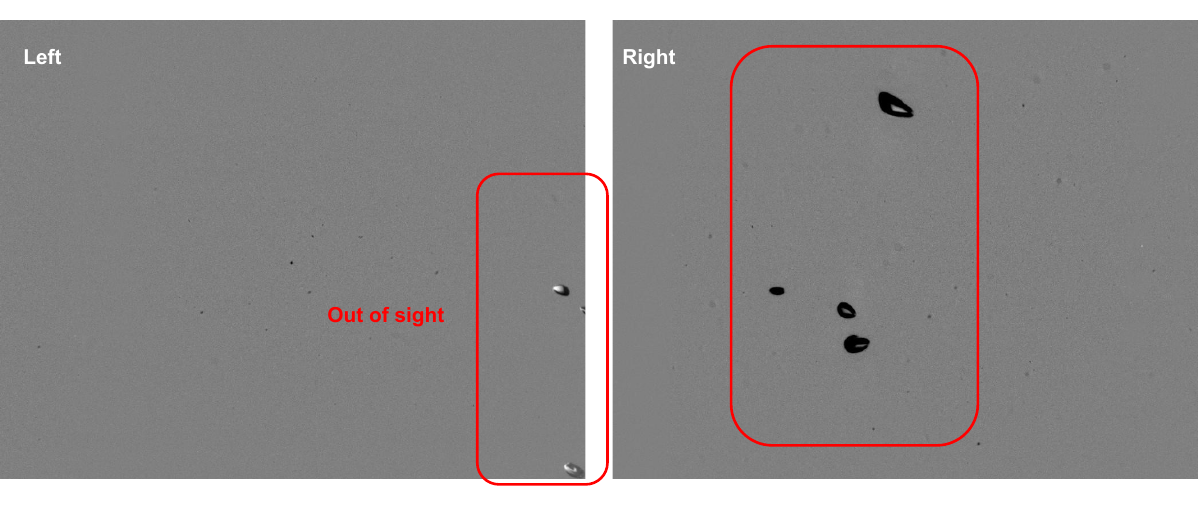}
	\label{fig:failure_case_2}
	}
	\end{center}
\caption{Failures. (a) Left, a sea urchin is sitting on the dome port. Right, a large piece of sediment is blocking the camera view. (b) The bubble stream does not rise vertical through the box, due to the instrument is not standing upright (standing on a slope, sunken into the sediment). When the bubbles get out of sight for one camera the stereo quantification does not work any more, and single view approximations must be applied.
}
\end{figure}
For a stereo evaluation, both cameras have to see every bubble. 
If the view of one camera is blocked by an animal, turbid water, the stereo matching procedure fails (see Fig. \ref{fig:failure_case_1}).
Also, when the BBox is not deployed in an upright orientation, e.g. at a slope or in case it sinks into the sediment (also see Fig. \ref{fig:green_light_in_dark} right), bubbles will not rise exactly through the corridor and might get out of sight for one of the cameras, as shown in Fig. \ref{fig:failure_case_2}.
In this case single view approximations (shape and position assumptions) have to be used.
In future versions we will use a wider image, and not crop it strictly to the 8cm corridor that works well only in laboratory conditions. This should increase robustness.

\subsubsection*{Density of Bubble Stream}
The entire system is designed for a stream of bubbles where bubbles rise more or less one by one. In case multiple or many bubbles are seen at the same height, they tend to occlude each other and they are difficult to distinguish from each other.
In this case small bubbles tend to be overlooked. This may bias the bubble size distribution, though the overall flow will probably only be slightly affected as most of the gas volume is transported by the bigger bubbles.
In future work, techniques to extract bubbles from a bubble cluster can be investigated \cite{honkanen2005recognition,zhang2012method}. 
Another possible direction is to employ deep learning based techniques to identify and extract bubbles\cite{haas2020bubcnn}, since it can work directly with images that contain complex background structures, background learning and removal steps can be omitted.

The opposite problem occurs in case the water contains many small particles (e.g. from suspended sediment). From a geometrical point of view they are difficult to distinguish from small bubbles, so a minimum bubble size threshold was used. This is a general specificity versus sensitivity problem.
A potential future solution could be to learn motion patterns and appearances. However, we currently do not know how well such an approach generalizes to previously unseen bubble sites, particularly if bubbles exhibit very different shapes with size\cite{ostrovsky08gasechosounder} or change shape at the same volume due to a hydrate skin\cite{rehder02bubblelifetime}.

\subsubsection*{Technical Limitations}
A current limitation of the instrument is its power consumption in standby-mode since only the flashes are disabled during standby, but the computers are still running. This limits maximum runtime. Similarly, recording full images rather than saving only the bubbles quickly fills the storage space. We have tried an experimental real-time encoding using live background subtraction. However, processing 0.8 gigapixel per second reliably is at the performance limit of the current hardware and since capturing data at sea is expensive we decided not to risk deleting the raw data.

Another limitation of the instrument is that the rise corridor inside the box largely isolates the bubble stream from effects of the surrounding currents, and so, under certain conditions, bubbles might not rise with the same speed inside the box as they would without the instrument. When entering the box from below, they have to pass a narrow (8cm $\times$ 8cm) opening, such that they rise centrally through the 20cm $\times$ 20cm sized corridor. 
It has been observed in \cite{TemporalVariationsofaNaturalHydrocarbonSeepUsingaDeepSeaCameraSystem} that bubble streams can produce a wake and rise speed is enhanced as compared to isolated bubbles of the same size.
In \cite{wang15bubblecrossflow} it has been shown that heavy crossflow can limit this bubble wake. The proposed instrument would however block such a crossflow and the confined corridor inside the box might contribute as well to a bubble wake. Consequently, in such a situation, the bubble rise velocity inside the box could be increased and provide an overestimate of the rise velocity without the instrument.

\section{Conclusion}
\label{sec:conclusion}
We presented and discussed a robust photogrammetric bubble stream characterization system for deep ocean deployment, the Bubble Box. 
The overall system has been deployed in more than 1000m water depth and was used to quantify methane fluxes offshore Oregon. 
In a test tank we verified the accuracy of the fully automated bubble radius estimation to be correct up to a few percent using hand-measured glass marbles and air bubbles as target objects.
Besides the robust bubble characterization method we have also presented a new in-situ calibration procedure that does not rely on point correspondences, but works with silhouettes in a wide baseline setting. 
This procedure allows for refining the calibration of the system in the environment where the measurements are conducted. This can become important for practical applications in the deep ocean where the environment conditions (e.g. optical properties of water) or also the misalignment of the instruments to each other (cameras and mirrors) might change.
Key limitations of the method such as bubbles out of sight of one camera in case the instrument is deployed on a slope or sunken into the sediment will be addressed in future versions and future work should also reduce power consumption.
The overall system however proved robust and is useful for characterizing bubble streams accurately, in case the number/density of gas bubbles is small to moderate. 

\section*{Acknowledgments}
The authors would like to thank Matthias Wieck, Jan Sticklus, Eduard Fabrizius and Thorsten Schott for co-designing and constructing the many components of the systems. Early ideas about building a bubble measurement device reach back to discussions with Anne Jordt, Peter Linke, Matthias Haeckel and Reinhard Koch, and a preliminary prototype has been built with Jens Schneider von Deimling.
We would particularly like to thank Anne Jordt and Claudius Zelenka for providing the software from \cite{jordt2015bubble} and many fruitful discussions. We are grateful to Lasse Petersen for providing the CUDA tool for background removal and Furkan Elibol for dockerization and the Jupyter notebook interface.
This work has received funding from Deutsche Forschungsgemeinschaft DFG (German Research Foundation) through the Emmy-Noether-Programme, Projektnummer 396311425 "DEEP QUANTICAMS". We would also like to thank Schmidt Ocean Foundation and the crew and team members of the Falkor cruise FK190612 for making the entire expedition and the bubble measurements offshore Oregon a success. Last but not least, the authors are grateful for support from the Chinese Scholarship Council (CSC) for Mengkun She and Yifan Song.

%\section*{References}

\bibliography{reference}

\end{document}